\documentclass[final,1p,times]{elsarticle}

\usepackage{url}
\usepackage{amsmath}
\usepackage{amsfonts}
\usepackage{amssymb}
\usepackage{graphicx}
\usepackage{caption}
\usepackage{subcaption}
\usepackage{hyperref}
\usepackage{color}
\usepackage{mathabx}
\usepackage{enumitem}
\usepackage{soul} 
\usepackage{marginnote}

\usepackage{siunitx}
\usepackage{algorithm}
\usepackage{algpseudocode}

\journal{Robotics and Autonomous Systems}

\begin{document}

\begin{frontmatter}

\title{From exploration to control: learning object manipulation skills through novelty search and local adaptation}

\author{Seungsu Kim}\ead{seungsu.kim@isir.upmc.fr}
\author{Alexandre Coninx}\ead{alexandre.coninx@sorbonne-universite.fr}
\author{Stephane Doncieux}\ead{stephane.doncieux@sorbonne-universite.fr}
\address{Sorbonne Universit\'{e}, CNRS, Institut des Syst\`{e}mes Intelligents et de Robotique, ISIR, F-75005 Paris, France}

\begin{abstract}
Programming a robot to deal with open-ended tasks remains a challenge, in particular if the robot has to manipulate objects. Launching, grasping, pushing or any other object interaction can be simulated but the corresponding models are not reversible and the robot behavior thus cannot be directly deduced. These behaviors are hard to learn without a demonstration as the search space is large and the reward sparse. We propose a method to autonomously generate a diverse repertoire of simple object interaction behaviors in simulation. Our goal is to bootstrap a robot learning and development process with limited information about what the robot has to achieve and how. This repertoire can be exploited to solve different tasks in reality thanks to a proposed adaptation method or could be used as a training set for data-hungry algorithms.

The proposed approach relies on the definition of a goal space and generates a repertoire of trajectories to reach attainable goals, thus allowing the robot to control this goal space. The repertoire is built with an off-the-shelf simulation thanks to a quality diversity algorithm. The result is a set of solutions tested in simulation only. It may result in two different problems: (1) as the repertoire is discrete and finite, it may not contain the trajectory to deal with a given situation or (2) some trajectories may lead to a behavior in reality that differs from simulation because of a reality gap. We propose an approach to deal with both issues by using a local linearization of the mapping between the motion parameters and the observed effects. Furthermore, we present an approach to update the existing solutions repertoire with the tests done on the real robot. The approach has been validated on two different experiments on the Baxter robot: a ball launching and a joystick manipulation tasks.

\end{abstract}

\begin{keyword}
Quality-diversity search, evolutionary algorithm, long-life learning, robotics, developmental robotics, behavior repertoire
\end{keyword}

\end{frontmatter}

\section{Introduction}

Open-ended learning refers to a process where an agent or robot must learn how to solve an unbounded sequence of tasks that are not known in advance. It is a key challenge to efficiently apply robotics to real world tasks~\cite{Doncieux2016driver}: real world environments are diverse and changing, robot users can define new tasks or modify existing tasks, and the robot capabilities themselves may change due to damage, upgrades or wear and tear. As modifying the robot programming to manually adapt it to new goals and situations is costly and unpractical, we strive to provide the robot with a developmental system allowing it to autonomously acquire and adapt its perceptual, cognitive and motor abilities in an iterative way. It has been proposed that this system can be implemented by combining some exploration processes able to build datasets of motor, perceptual and other relevant data about the world and the tasks at hand, and a representational redescription process able to use this data to build and update markovian models of the tasks, which can then be exploited~\cite{Doncieux2018framework}.

End-to-end learning techniques, such as deep reinforcement learning algorithms, combine both processes to directly build policies~\cite{Levine2016handseye}, and can theoretically be used to learn new tasks and adapt to environmental variations in such open-ended scenarios. However, they require very large learning datasets, which is not suited to real robotics tasks where evaluating the outcome of actions or policies is slow and costly~\cite{Mouret2016microlearning}, except in the rare cases when many real robots can be used in parallel~\cite{Levine2016handseye}. They are also notably slow to converge when reward states are sparse, which is often the case in tasks necessitating to reach some very specific goal states using only partial, high-dimensional and noisy obervations of the world.

An approach to deal with this issue is to keep exploration and redescription separated and to explore in a simulated environment. Evaluating actions in simulation is cheaper and faster than with real robots. Sample-inefficient exploration algorithms, such as quality-diversity (QD) algorithms~\cite{Pugh2016,Cully2018qdmodular}, can thus be used. Rather than a single policy, these techniques learn a repertoire of actions covering a given behavioral space, i.e. a large dataset of diverse actions that reach various goals in that space. Such repertoires of diverse actions can be exploited to solve different tasks, and to adapt to new unplanned situations arising in open-ended scenarios, such as obstacles in the environment, or the robot breaking down. They can also be used as a training dataset for other machine learning algorithms, e.g. deep learning, to learn state representations~\cite{LaversanneFinot2018goalspaces} or parameterized policies~\cite{Jegorova2018gan}.

Most of the previously cited works, however, remain in the simulation domain. In order to exploit these repertoires in the real world, we have to cross the reality gap~\cite{Jakobi1995}, i.e. to adapt the behavior to the differences between the simulated and real environments and robots, and transfer it to the physical system. A related issue arises from the discrete nature of the repertoires: although a large number of actions are learnt, they can only imperfectly cover a continuous behavior space, and depending on the accuracy required for the task, some points may remain unreachable. It is therefore necessary to find an adequate adaptation process able to deal with those issues and solve real world tasks. This adaptation process needs to rely on robot's experience, but since evaluating actions in the real world is costly and slow, it is important to use sample-efficient techniques that minimize the number of trials and maximize the amount of information drawn from each trial.


In this work, we propose such a process based on a modeling process of the mapping between the action parameter space and the behavioral space in the real world. First, we use QD algorithms to learn a large but discrete action repertoire in simulation, covering a given behavioral space. Then, we define a goal in this behavioral space and try to reach it using the repertoire in a real robotics setup. After an action has been tried on the real robot, in case of failure to reach the goal, the nearest neighbors in the repertoire are used to build a local linear estimation of the Jacobian matrix mapping the motion parameters to the observed effect. The Jacobian is then used to determine the modifications to apply to the action parameters to reach the goal despite the discretization error or the reality gap. We apply this approach two object manipulation tasks, a ball throwing task and a joystick manipulation task, on a Baxter robot arm with seven degrees of freedom.

The contribution of this paper is therefore twofold: we show that QD algorithms can efficiently be used to learn action repertoires in simulation for object-oriented tasks, and we propose a method to exploit the repertoires in a real setup, dealing with the reality gap and the discretization error. In section~\ref{sec:related}, we discuss related works and in particular approaches to deal with action repertoire acquisition and with the reality gap problem. We then expose our methods, first the QD search and then the adaptation process, in section~\ref{sec:method}. We present our experimental validations in section~\ref{sec:experiment}, describing the experimental setup and the results for the ball throwing and the joystick manipulation tasks. Finally, we discuss the outcomes in section~\ref{sec:discussion} and conclude with future work.

\section{Related works}
\label{sec:related}














Three notions are central in the literature about robot behavior acquisition and adaptation: actions, policies and skills. Before entering in the details of related works, we will define them to avoid any ambiguities. The definitions used here rely on the formalism of Markov Decision Processes (MDP). An MDP is a tuple $<S,A,p,R>$, where $S$ is a state space, $A$ an action space, $p$ a (stochastic) transition function and $R$ a reward. A policy $\pi: S \to A$ is a function that associates an action to a state. In a reinforcement learning setup, it is built by the learning process so that the sequence of actions and the resulting states, when the system follows this policy, maximizes the cumulated reward. The MDP formalism does not further specify what an action is: it is part of the MDP definition and left to the system designer. As our focus here is on how these actions could actually be built, we need to give more detail. Following the formal framework for open-ended learning introduced in \cite{Doncieux2018framework}, we define an action as a policy defined in a lower level action space $A'$, and a goal i.e. a target point in the state space. This is a recursive definition, that suggests to start from low level actions (motor orders) and progressively build more abstract actions. A skill (motor skill) is an action generator that associates an action to an initial state and a goal state. Building a system able to deal with unforeseen object interactions can be dealt with by using algorithms able to generate (1) a set of actions to dig into or (2) skills to use on-the-fly. It is also possible to learn policies relying on known high-level actions, but we will not focus on these approaches as these high level actions are difficult to design in practice and limit reachable robot behaviors to combinations of these actions. See \cite{kober_reinforcement_2012}, for instance, for a review. Section \ref{sec:method} introduces the formalization in more detail.

\subsection{Goal exploration and skill learning} 

Several methods have been put forward to make robots autonomously explore their environment and learn new skills to reach various goals. An important part of this literature treats this as a control problem, where the system learns an inverse model (or multiple inverse models) to generate the motor commands in order to reach an arbitrary position in a given goal space. Following some works showing that human babies learn how to control their body through ``body babbling''~\cite{Meltzoff1997}, several methods use a similar motor babbling approach in robots with random motor orders~\cite{DSouza2005,Demiris2005motor2hammer}. The two main issues of these approaches are that (1) learning the inverse model from collected samples is a challenging supervised machine learning problem for non-toy problems and that (2) sampling the motor space - which may be large - in a sample-efficient way is required and also difficult to achieve. Several approaches have been proposed to tackle these issues.

A proposal to make the sampling process more efficient is to sample in the goal space rather than in the motor space, choosing points in the goal space and using the robot's current inverse model to \textit{attempt} to reach them, thus generating new samples that allow to further train the inverse model online. This ``goal babbling'' approach has been found to be more sample-efficient than motor babbling when using a small set of simple predefined goal space targets~\cite{Rolf2009}, or random goals that can be generated without any prior knowledge of the expected robot behavior~\cite{Rolf2010inversekinematics}. Since an existing inverse model is required to generate motions at any time, this model has first to be bootstrapped, for example with random initialization. Further developments of this goal babbling approach improve the sampling efficiency through the use of intrinsic motivations that choose goals maximizing the learning progress~\cite{Oudeyer2007intrinsic}.

In order to simplify inverse model learning, it has been proposed to divide the goal space in several regions according to a spatial segmentation~\cite{Baranes2009riac,Baranes2013goalexploration}, or in several independant subspaces corresponding to the state of different objects~\cite{Forestier2016icirs}, and learn a different inverse model for each subspace. Another alternative is to use unsupervised learning to learn a goal state representation with which the goal babbling process is improved~\cite{Pere2018,LaversanneFinot2018goalspaces}. 

These approaches have proved to be very efficient to learn to control a goal space. However, as they focus on inverse model building, they create a unique mapping from the goal space to the motor space. This implies that they can only learn a single way to reach a given goal, or at least one per model. This may be a limitation in real world scenarios, as the solution found may not be applicable, because of obstacles, for instance. This is especially relevant for highly redundant robotic systems, such as multi-joint robot arms, for which different arm movements are possible to reach a particular goal. Having several different trajectories instead of a single one opens the way to more adaptive controllers that could select on-the-fly a trajectory that prevents the arm from bumping into obstacles instead of requiring to start again exploring from scratch to adapt to the current situation.

\subsection{Learning a repertoire of actions}

The previously mentioned methods follow a top-down appraoch where high-level goals are set and the learning system strives to achieve them, learning increasingly better inverse models. A different approach to building skills is to proceed in in a bottom-up way, by starting with low level sensorimotor exploration and then restructuring the acquired knowledge to build higher level actions and skills. This can be done by using direct policy search algorithms (such as fitness-based evolutionary algorithms) to solve some specific low-level sensorimotor tasks, and then to extract from those low-level policies some \emph{learning traces}, i.e. sequences of sensorimotor behavior that are correlated to goal-directed behavior~\cite{Zimmer2018}. Unsupervised learning algorithms can then extract a small set of higher level actions (e.g. obstacle avoidance, goal approach behavior, etc.) from those traces, which can then be used with a reinforcement learning algorithms to generalize and reach arbitrary new goals. Although this approach works well for problems like navigation where a policy can naturally be described as a sequence of lower level decisions and actions, it is not straightforward to apply to problems such as object manipulation primitives, which lack this structure.


Other approaches directly learn a repertoire of high-level actions through divergent evolutionary algorithms such as novelty search~\cite{Lehman2011ns}. Evolutionary algorithms (EAs;~\cite{back1996evolutionary,eiben2003introduction,vikhar2016ea}) are a class of gradient-free optimization algorithms~\cite{Sigaud2018policysearch} inspired from the principles of variation and selection that structure the evolution of life. To solve an optimization problem, an EA will start by generating a random set of candidate solutions (the initial population), and evaluate them using a fitness function assessing their performance for the stated problem. The worst performing solutions are then discarded whereas the best performing ones (the parent population) are duplicated and subjected to small random perturbations. This new set of candidates (the offspring population) is then evaluated, and used to generate the parent population of the next iteration. The process is iterated until a good enough solution to the problem is found.

Divergent evolutionary algorithms such as novelty search~\cite{Lehman2011ns} are a variant of EAs where the selection is not (or not only) driven by an explicit optimization objective: instead of a scalar fitness function that must be maximized, the algorithm uses a \emph{behavior characterization function} which maps actions to points in a small- or medium-dimensional \textit{behavior space} - such as a goal space. For example, a behavior space for a mobile robotics scenario can be the final position of the robot, or a discretized trajectory. The goal of a divergent EA such as novelty search is not to reach a given point in this space but to learn a repertoire of actions covering the whole space. Since those approaches do not explicitly build an inverse model, they explore in the motor space and because they do not rely on a unique (or locally unique) inverse model, they can find multiple actions to reach a goal in different ways.

Further works introduced quality-diversity algorithms~\cite{Lehman2011nslc_virtual,Pugh2015QD,Cully2018qdmodular}, which combine the behavior space exploration with a global or local quality metric, allowing to learn repertoire of diverse, high-quality actions (under some given quality measure). They rely on the principle of local competition: new actions are added to the repertoire if they are novel, but existing actions in the repertoire can also be replaced by newer ones with similar behavior but higher quality as they are found. Some of those algorithms, such as MAP-Elites~\cite{mouret_illuminating_2015}, rely on discretizing the behavior space into a set of bins (a behavioral map), and then filling the bins with increasingly higher quality actions. This makes the algorithm simpler, allows for efficient data visualization, and easily allows to exploit the discrete behavioral map through techniques such as bayesian optimiation~\cite{cully_robots_2015}. For high-dimensional behavior spaces, the number of bins for the MAP-Elites grows exponentially with the number of additional dimensions. In order to handle high-dimensional behavior space,~\citet{vassiliades_scaling_2016} propose an extension to replace the equal grid of the MAP-Elites algorithm by an adaptive one using centroidal voronoi tessellations. It partitions the behavior space into $k$ homogeneous geometric regions, and prevents the number of bins from growing exponentially for high-dimensional datasets.


Exploiting this repertoire raises challenges that differ from those of learning an inverse model. Reaching a known goal for which an action is present in the repertoire~\cite{Lehman2011ns,Pugh2015QD}, is very simple and inexpensive, as the correct action just needs to be selected and executed. Adapting to changes in the problem domain (for example a slightly different environment, or a damaged robot, which modifies the effect of actions) and finding the right action to reach an unknown goal can be done in a sample-efficient way by discretizing the behavior space and using bayesian optimization to quickly discover the best action~\cite{mouret_illuminating_2015,cully_robots_2015,vassiliades_scaling_2016}. Some recent works propose to use the repertoire as a training set for a conditional generative-adversarial network~\cite{mirza2014conditional} in order to build a goal-conditioned action generator~\cite{Jegorova2018gan}. Although the learnt model still has limited accuracy, this approach is promising as it both allows generalization to arbitrary goal spaces positions (as for inverse models) and can generate multiple, diverse actions for a given target position.

\subsection{Crossing the reality gap}

The previously described skill learning techniques are generally performed in a simulator, which acts as a direct model of the robotic system and its environment. Simulation can be much more practical than real robotics, being cheaper, safe from damaging the robot and its environment during experiments, and much faster than real experiments as simulations can be performed faster than real time and massively parallellized. However, the actions learnt in a simulator often do not transfer well to a real robot and environment, because the exact physical properties of the robot body and its environment can never be perfectly modeled. The behavioral differences between simulated and real experiments have been termed the reality gap~\cite{mouret_crossing_2013}.

Crossing the reality gap is therefore a key issue to a wider use of simulation-based methods. Some existing methods focus on improving the simulation to make the policies more transferable, for example by adding noise to the simulation to generate more robust behaviors~\cite{Jakobi1995}, by co-evolving the policy and the simulator~\cite{Bongard2004,Zagal2004}, or by updating the simulator's forward model thanks to real observations~\cite{kupcsik_model-based_2017}. Those methods, however, are not sufficient yet to tackle high-dimensional problems in open environments~\cite{Golemo2018phd}. 

Other approaches rely on combining learning in the simulated domain with more limited experimentations in the real domain. For methods learning inverse models or parameterized policies, the reality gap is then a domain transfer problem: the learnt simulation controller must be transferred to the real world. Using state of the art deep transfer learning methods, such as learning domain-invariant features~\cite{Ganin2015} is challenging due to an unbalanced dataset between simulated and real data, and the lack of an exploratory process to efficiently sample real world data. Other techniques have been proposed to learn a deep model of the reality gap itself and apply it to the controller, either in action space~\cite{Hanna2017} or in goal space~\cite{Golemo2018}. Those approaches are successful at crossing the reality gap for complex, high-dimensional problems; however they require a relatively large number of real world experiments (tens to hundreds).

When relying on an acquired repertoire, the reality gap can be handled by a different approach: the simulation-generated repertoire contains a large set of diverse actions that can be exploited to cross the gap. Using actions learned in simulation on a real robot is both a domain transfer problem (from simulation to real world) and an intra-domain generalization problem, since there is no guarantee that the repertoire contains an action that can reach a given specific goal in the real world, because of the discretization error. The proposed approach addresses those two problems simultaneously with a local adaptation of action and, in case of a failure of this adaptation process, the diversity of the repertoire allows to find another candidate action to try. In contrast, approaches such as promoting controllers that transfer well to the real world~\cite{Koos2010,Koos2013} only addresses the first problem and reduces sample efficiency as poorly transferable controllers are not exploited at all. Other more general domain transfer methods have been proposed, allowing to very efficiently explore the repertoire and find the adequate behavior with very few real world trials~\cite{cully_robots_2015,Chatzilygeroudis2016}, but they still make the assumption that the repertoire does contain an adequate behavior and do not address the intra-domain problem.

\section{Method}

\subsection{Approach overview}

This article presents an approach to control a given goal space with a real robot and with few real world tests, by first creating a large repertoire of diverse, efficient behaviors in a simulated environment, and then using this repertoire to reach an arbitrary goal in the real world with only a few trials.

Experiments involve a robotic arm with simple, parameterized, open-loop control policies. Actions are joint space trajectories defined by joint space position and velocity at the end of the motion (which constitutes the search space of the QD algorithm). They are represented by a third order polynomial. We first use a QD algorithm to build, in simulation, an actions repertoire, i.e. a set of policy parameters associated to the goal they have reached. The action repertoire is structured by a behavior space suited to the task at hand (e.g. for a targeted ball throwing task, the position of the ball when it crosses the target plane, see section~\ref{sec:method_qd}). After building the repertoire, we apply this repertoire to real world experiments, trying to reach a target in the behavior space. If a suitable action is present in the repertoire generated in simulation, the behavior is attempted; if the target is reached, the task is solved. If the target is not reached, the behavior space error between the expected (simulated) position and the reached position is measured and a local linear direct model is computed from the attempted action and its neighbors in the motor space from the repertoire. This local linear model is inverted, and the resulting local inverse model is used to adapt the action to remove the error, locally crossing the reality gap. This process is iterated until the target is reached, which typically only takes a few trials (section~\ref{sec:crossing_reality_gap}). If no suitable action is present in the simulation repertoire, a similar approach can be used to generalize from the existing actions, starting from the closest known action (in the behavior space), computing a local inverse model, and using it to reach the behavior space goal by gradient descent (section~\ref{sec:method_jacobian}).



We validate the proposed approach in two different applications where the task is non-trivial, the behavior space quite large and a highly accurate control is needed: throwing an object into a basket, and controlling a joystick to reach specific angles. We assume that the robot does not have any prior information about the skills beside its body kinematics. Furthermore, no feedback nor demonstration is provided by humans at any stage, besides defining the environment and the behavior spaces for exploration, and the target goal for exploitation.


\subsection{Problem formalization and notations}

\label{sec:method}

\begin{figure}[h!]
\center
	\begin{subfigure}{.44\textwidth}
	\centering
	\includegraphics[width=7Cm]{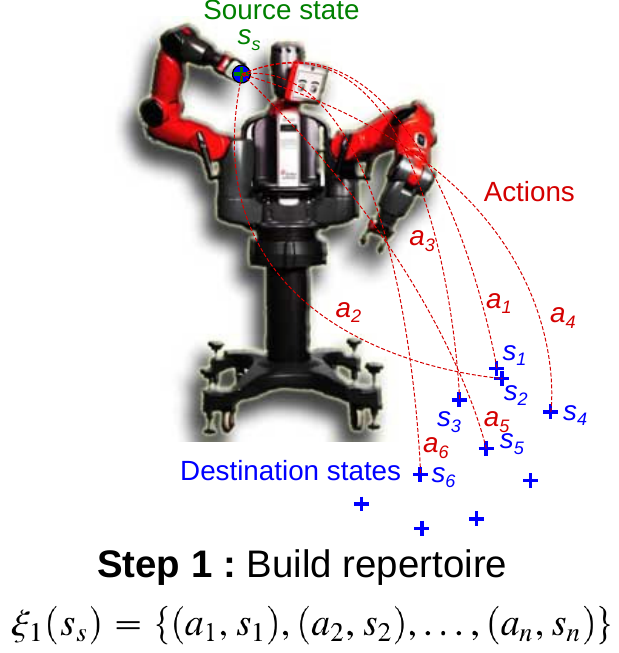}
	\caption{}
	\label{fig:general-description-step1}
	\end{subfigure}
	\hspace{.1\textwidth}
	\begin{subfigure}{.44\textwidth}
	\centering
	\includegraphics[width=7Cm]{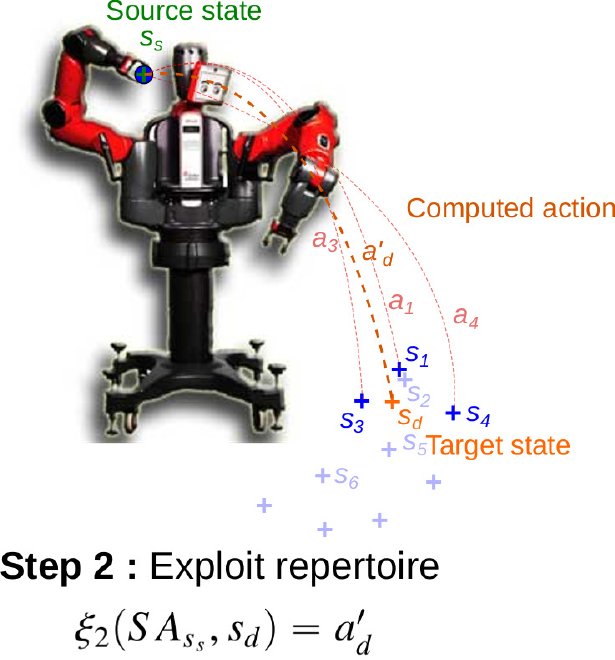}
	\caption{}
	\label{fig:general-description-step2}
	\end{subfigure} 
\caption{General description of the proposed approach with a robot ball throwing scenario. Here, the state space $S$ is the ball position and the action space $A$ is the parameter space of a parameterized motion primitive. The robot initially holds the ball in a state $s_s$. In a first step (Fig.~\ref{fig:general-description-step1}), an exploration method is used to build a repertoire $\xi_1(s_s)=\left\{ (a_1, s_1), (a_2, s_2), \dots, (a_n,s_n)\right\}$ of state-action pairs allowing to reach many points in $S$, i.e. to throw the ball at various positions. Note that very similar states can be reached by different actions, as illustrated by state-action pairs $(a_1, s_1)$ and $(a_2, s_2)$. In the second step (Fig.~\ref{fig:general-description-step2}), the repertoire is used to learn how to reach arbitrary target states $s_d$ from $s_s$, using an action $a'_d$. $a'_d$ can directly be drawn from the repertoire if $s_d$ is close enough to a destination state in the repertoire, or can be computed from a set of neighbors in $A$ using the jacobian linearization method described in section~\ref{sec:method_jacobian}. The whole process constitutes a \emph{motor skill}, allowing to efficiently control the given state space $S$.}
\label{fig:general-description}
\end{figure}

The proposed method is summarized with an example in Figure~\ref{fig:general-description}. It aims at learning a skill to generate a large range of effects from an initial state. The skill is a function $\xi$:

$$\xi: S \times S \to A $$


where  $A$ is the action space and $S$ is a state space, or behavior space, defined by \textit{behavior descriptors}, which are features of the action effects we want to focus on. The proposed approach aims at controlling this state space, i.e. defining a process $\xi(s_s, s_d)$ that generates the action to reach a destination state $s_d$ from a source state $s_s$. To this end, it relies on two steps: (1) generation of a large and diverse repertoire of actions and (2) exploitation and adaptation of the repertoire: $\xi = \xi_1 \circ \xi_2$.

The first step aims at finding, for a given $s_s$, the set of all reachable states with the corresponding action. Let us call $SA_{s_s}\in \mathcal{SA}$ a set of $(s_i, a_i)$ pairs where $a_i$ is the action to reach $s_i$ when starting from state $s_s$:

$$\xi_1: S \to \mathcal{SA}  $$

$\xi_1(s_s)=\left\{ (a_1, s_1), (a_2, s_2), \dots, (a_n,s_n)\right\}$ 

The method aims at covering, as well as possible, the space of reachable states when starting from $s_s$. $n$, the number of action-state pairs, is not fixed. It depends on a particular execution and computational budget: the longer the algorithm is run, the larger $n$ is. It should be noted that this approach does not need any demonstration to bootstrap. It relies on an exploration method based on a QD algorithm. As this algorithm requires a large number of action evaluations, they are performed in simulation. 

The second step $\xi_2$ extracts from a $SA_{s_s}$ an action to reach a desired destination state $s_d$:

$$\xi_2: \mathcal{SA} \times S \to A $$

$\xi_2(SA_{s_s}, s_d)=a'_d$

$s_d$ may not be in  $SA_{s_s}$ or the execution of $a_d$, the action associated to $s_d$ in $SA_{s_s}$, may lead to an error, because of the reality gap. For both reasons, $\xi_2$ includes an adaptation part and generates an action $a'_d$, that may be different from $a_d$. Details of these two steps are provided in the following sections.

\subsection{Step 1: Offline QD search}
\label{sec:method_qd}

The goal of this step is to generate $SA_{{s_s}}$, a repertoire of action-state pairs that is as diverse as possible, so that it contains any relevant action for the robot to deal with the situations it will be faced with. This repertoire acquisition process relies on a QD algorithm. QD algorithms are evolutionary algorithms that use the Darwinian principles of variation and selection to fill a repertoire of actions~\cite{Pugh2015QD}. Instead of being driven by performance, they are mainly driven by a novelty score that rewards actions that have led to a new and original effect, as measured by the distance to other effects in the behavior space. A task-specific quality score is also used at a local scale, to keep the best performing actions among those generating a similar behavior.

A robot action is described by a vector of real parameters $\mathbf{g} \in \mathbb{R}^n$, which constitutes our action space ($A = \mathbb{R}^n$). For the purpose of the QD algorithm, the parameter vectors $\mathbf{g}$ are also called \textit{genotypes}, as those are the elements on which the genetic algorithm operates. The QD algorithm explores in parallel a set of genotypes that are initially randomly generated. Each action is evaluated using a robot simulator, represented by an action evaluation function $\mathbf{f}$, to determine to which behavior it leads: $\mathbf{f}(\mathbf{g}) = \mathbf{b}$, where  $\mathbf{b} \in \mathbb{R}^m$ is the \textit{behavior descriptor} associated to $\mathbf{g}$, and $S = \mathbb{R}^m$ is the behavior space. The descriptor can include any observable states of a robot and/or its environment. It is predefined and structures the repertoire of actions. It should be noted that the mapping $\mathbf{f}$ between the action and the descriptor is highly non-linear and that no closed-form expression is known for it.

Several different QD algorithms have historically been proposed~\cite{Lehman2011nslc_virtual,Pugh2015QD,cully_robots_2015}, and many other variants can be defined. Recent work by Cully \& Demiris \cite{Cully2018qdmodular} proposed a modular framework which defines the main features of a QD algorithm (the type of container used, the procedure to create a parent population, and the procedure to add individuals to the container) and explores the possible variants based on the existing literature and new ideas. We used this framework to explore different QD algorithms in preliminary experiments and select the one to use for exploration in this work.

The behavior spaces used for our scenarios tend to contain discretized robot arm trajectories: they tend to be quite high-dimensional, and furthermore they are constrained by robot's kinematics and dynamics: only a subset of high-dimensional space corresponds to trajectories that can actually be achieved. For this reason, we eschewed the methods that rely on discretizing the behavior space to fully cover it (as in MAP-Elites~\cite{mouret_illuminating_2015} and its developments~\cite{vassiliades_scaling_2016}) and we chose an algorithm using an unstructured archive (as in NSLC~\cite{Lehman2011nslc_virtual}) rather than a structured grid or tessellation.

The algorithm we selected is the QD variant ``arch\_novelty'' proposed in \cite{Cully2018qdmodular}. This algorithm performed well in preliminary testing experiments, it is shown in \cite{Cully2018qdmodular} to consistently be among the top performing variants, and it is easily implementable as it only uses building blocks well known to the QD community (novelty, local quality, novelty archive). It selects parent population from the archive according to the novelty score, and insert new individuals in the archive if they are novel or higher quality than their nearest neighbor (in which case the lower quality neighbor is discarded). Algorithm~\ref{algo:qdsearch} shows the corresponding pseudo-code.

\begin{figure}
\center
\includegraphics[width=10Cm]{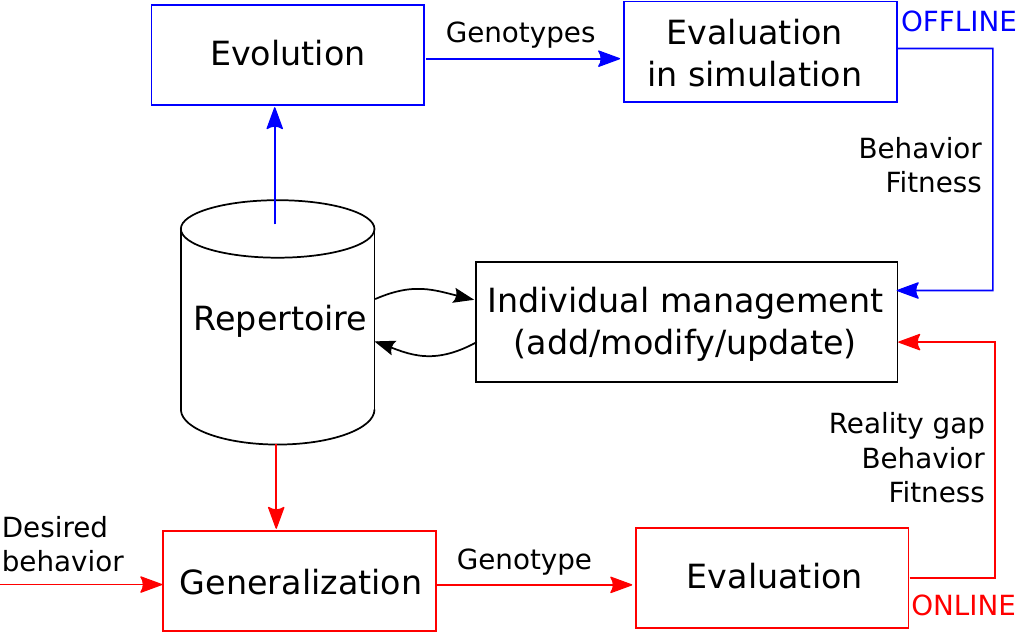}
\label{fig:overview}
\caption{Overview of the proposed method. The offline repertoire generation technique (correpsonding to the blue part) is described in section~\ref{sec:method_qd}, the online adaptation and reality gap crossing technique (correpsonding to the red part) is exposed in section~\ref{sec:method_genotype_computation}.}
\end{figure}

\begin{algorithm}
\caption{``arch\_novelty'' QD algorithm}
\begin{algorithmic}
\State \emph{repertoire} $\leftarrow$ {$\emptyset$}
\Repeat  \Comment{initialization}
	\State \emph{pop} $\leftarrow$ \Call{populate}{\emph{random}}
	\ForAll{action $g$ $\in$ \emph{pop}} 
		\State $g$.\emph{quality} $\leftarrow$ \Call{evaluate}{$g$} 
		\If{ \Call{valid}{$g$} == True} \Comment{if $g$ is valid}
			\State \Call{add}{$g$, \emph{repertoire}}
		\EndIf
	\EndFor
\Until{repertoire size $>=$ 1}

\Repeat \Comment{main loop}
	\State \emph{parents} $\leftarrow$ \Call{select}{\emph{repertoire}} \Comment{proportional to novelty score}
	\State \emph{pop} $\leftarrow$ \Call{populate}{\emph{parents}} \Comment{mutate and crossover}
	\ForAll{action $g$ $\in$ \emph{pop}} 
		\State $g$.\emph{quality} $\leftarrow$ \Call{evaluate}{$g$} 
		\If{ \Call{valid}{$g$} == False} \Comment{if $g$ is not valid}
			\State \Call{continue}{}
		\EndIf
		\State $g$.\emph{novelty} $\leftarrow$ \Call{novelty}{$g$, \emph{repertoire} $\cup$ \emph{pop}} 
		\State \emph{nn} $\leftarrow$ \Call{closest}{$g$, \emph{repertoire}} \Comment{the closest action}
		\If{ \Call{dist}{$g$, \emph{nn}} $>$ $l_{repertoire}$ } \Comment{distance between $g$ and \emph{nn}}
			\State \Call{add}{$g$, \emph{repertoire}}
		\ElsIf{$nn$.\emph{quality} is significantly better than $g$.\emph{quality} }
			\State replace \emph{nn} with $g$ in the \emph{repertoire}
		\EndIf
	\EndFor
	\State update the novelties of all the actions of the \emph{repertoire}
	\Until{total generation $>=$ \emph{max-gen}}
\end{algorithmic}
\label{algo:qdsearch}
\end{algorithm}


Initially the repertoire is empty. The algorithm first evaluates a set of randomly generated genotypes (TThe number of individuals is the same as the population size which is a user defined parameter. See Appendix \ref{appendix:qd_pram}). It repeats the random genotype evaluations until it finds any valid actions. The valid actions are added into the initial repertoire. Here, valid actions mean any actions where the quality can be computed without any failure. For example, in the ball throwing experiment, if the robot reaches its joints' position/velocity limits or is self collided for performing its actions, the actions are considered invalid actions. 

The novelty score of an action $g_c$ is computed by kernel density estimation in the behavior space:
\begin{equation}
nov(g_c) = 1.0 - \frac{1}{N h} \sum_{i=1}^{N} \Phi \left( \frac{dist(\mathbf{f}(g_i), \mathbf{f}(g_c))}{h} \right) 
\label{eqn:novelty}
\end{equation}
where $dist(.)$ is a Euclidean distance; $N$ is the number of actions in the archive; $\mathbf{f}$ is the evaluation function that generates the behavior descriptor resulting from the evaluation of a genotype in a given domain; $\Phi$ represents a Gaussian kernel function $\Phi(x) = \exp( -0.5 x^2 ) / \sqrt{2\pi}$; $h$ is a kernel width which is computed by Silverman's rule of thumb approach~\cite{Silverman1986}.

The parent population is the set of actions selected to generate new actions. They are selected from the population and the archive on the basis of their novelty score. Following the principles of novelty search~\cite{Lehman2011ns}, the actions that breed the new generation are then the most original ones. 

The new population is created from the parent set of actions. Each parent is part of a pair that generates two new actions through an SBX crossover operator~\cite{agrawal_simulated_1995}. A polynomial mutation is then applied to the newly created actions~\cite{deb_multi-objective_2001}.

Each action is evaluated in simulation. After evaluation, the behavior descriptor of a genotype is compared to its closest neighbor in the repertoire. If this distance is above a threshold ($l_{repertoire}$), then it is added to the repertoire, otherwise, it replaces the closest neighbor if it is better performing according to a predefined quality score function. This sequence is repeated for a given number of generations.



The outcome of the process is a repertoire $SA_{s_s}=\left\{(a_1, s_1), \dots, (a_n, s_n) \right\}$ containing actions that allow the robot to reach, from state $s_s$, a set of states that is as diverse as possible. This step is the longest one and may require more than 10 hours of computation. It is done only once before robot use.


\subsection{Step 2: Online action choice and adaptation}
\label{sec:method_genotype_computation}

The QD algorithm builds a repertoire of solutions in the continuous action and state spaces. The repertoire contains a sampling of possible actions. When using it to perform an action to reach a given goal in the behavior space, two different issues may arise: (1) the behavior in simulation may differ from the behavior observed on the real robot (reality gap, i.e. inter-domain generalization problem) or (2) the desired state may not be among the samples in the repertoire (intra-domain generalization problem). 

The reality gap may result from two different situations. The simulated behavior may be completely unrealistic. In this case, the solution found is useless. A possibility to deal with this issue would be to make some experiments on the real robot during the QD search to identify these particular solutions and avoid them~\cite{Koos2013}, but to keep the repertoire generation process as simple as possible, we have not used this possibility. We have thus made the assumption that most behaviors are realistic. The second situation occurs when the behavior in simulation is realistic, but just slightly different from what occurs in reality. In this case, the action found is informative and can fit the needs with only a limited and local adaptation. We have focused on this second case and describe in the following a method to perform this local adaptation. Furthermore, the same approach also solves the sampling problem as soon as a solution in the repertoire is close enough to the desired state. As the QD search generates, by construction, a repertoire that is expected to cover the space of possible behaviors as completely as possible, we have considered that this hypothesis holds.

The proposed method performs a local linearization of the mapping between the action and the behavior state spaces around a particular action and modifies the action parameters accordingly to reach the target behavior.

For a given desired target behavior, the most similar action, in terms of Euclidean distance in the state space, is selected in the repertoire. The corresponding local Jacobian between the action and state spaces is computed. Then, using a gradient descent approach based on this Jacobian, a new action is built for a given desired target state. More details are provided in the following subsections. 


\subsubsection{Local Linearization}
\label{sec:method_jacobian}
As discussed in section \ref{sec:method_qd}, 
a state is described by the behavior effect $\mathbf{b} \in \mathbb{R}^m$ observed when the robot controller is parameterized with a vector of real values $\mathbf{g} \in \mathbb{R}^n$ in a simulation or in the real environment; 
$\mathbf{f}: \mathbb{R}^n \rightarrow \mathbb{R}^m$, $\mathbf{b}=\mathbf{f}(\mathbf{g})$. 
The mapping  $\mathbf{f} = \left[\begin{matrix} f_1& f_2& \cdots & f_m \end{matrix} \right]^T $ is highly non-linear and its closed-form expression is unknown. 
In order to estimate how the behavior will be affected by local changes of the action parameters, $\mathbf{f}$ is locally linearized at a point of interest $\mathbf{g}_c$.

\begin{equation}
\Delta \mathbf{b}_c = 
\left .
\begin{bmatrix}  
	\frac{\partial f_1}{\partial g_1} & \frac{\partial f_1}{\partial g_2}  & ... & \frac{\partial f_1}{\partial g_n}\\ 
	 &   & \ddots  &  \\
	\frac{\partial f_m}{\partial g_1} & \frac{\partial f_m}{\partial g_2}  & ... & \frac{\partial f_m}{\partial g_n}
\end{bmatrix}
\right\rvert_{\mathbf{g}_c} 
\Delta \mathbf{g}_c 
= 
J({\mathbf{g}_c}) \Delta \mathbf{g}_c 
\end{equation}
where $J({\mathbf{g}_c})$ is the Jacobian matrix at $\mathbf{g} = \mathbf{g}_c$.

As we do not have an analytical form of $\mathbf{f}$, we estimate the Jacobian using the samples in the repertoire. 
A set of $\mathbf{g}_c$ neighbors in the repertoire is collected: $\{g_k, b_k\}^K_{k=1}$, where the distance of each of these actions to $\mathbf{g}_c$ is smaller than $\epsilon$, $||g_k-g_c|| < \epsilon$. To estimate $J({\mathbf{g}_c})$, the Jacobian near $\mathbf{g}_c$, a least squares method is used : 
\begin{equation}
\tilde{J}({\mathbf{g}_c})  = BG^T(GG^T)^{-1}
\end{equation}
where $G=[\mathbf{g}_1, \mathbf{g}_2, ... \mathbf{g}_K]-\mathbf{g}_c$ is a derivative of action parameters matrix near $\mathbf{g}_c$; $B=[\mathbf{b}_1, \mathbf{b}_2, ... \mathbf{b}_K]-\mathbf{b}_c$ is the corresponding derivative matrix in the behavior space.

The quality of the linearization depends on the local linearity of the mapping or on the repertoire sparsity. 
The confidence of the Jacobian, $\lambda$, is defined as the quality of the linearization: 
\begin{eqnarray}
\eta &=& | \tilde{J}({\mathbf{g}_c}) G - B | \\
\lambda &=&  \begin{cases}
    1- \eta/\eta_{\text{threshold}},&  \eta < \eta_{\text{threshold}}\\
    0                       ,& \text{otherwise}
\end{cases}
\label{eqn:local_linearity}
\end{eqnarray}
where $\eta_{\text{threshold}}$ is a user defined positive constant that is related to the maximum allowed error in the behavior space. 

If the repertoire is locally linearizable near $g_c$, the behavioral estimation error $\eta$ is near $0$, then $\lambda = 1.0$.  Otherwise, $\lambda$  decreases proportionally to the behavioral estimation error $\eta$.  This value will be used later as a weight to adapt the action parameters.

\subsubsection{Action parameters computation for an arbitrary behavior}
Starting from a given desired target state $\mathbf{b}_d$, the corresponding action parameters vector is computed with a gradient descent. The gradient descent is initialized with the nearest action $\{\mathbf{b}_c, \mathbf{g}_c\}$ from $\mathbf{b}_d$ in the repertoire, $\{\mathbf{b}_j, \mathbf{g}_j \}|_{j=0} = \{\mathbf{b}_c, \mathbf{g}_c \}$, where $j$ is an iteration index for gradient descent method. 

\begin{equation}
\mathbf{g}_{j+1} =  \mathbf{g}_{j} + \lambda \tilde{J}({\mathbf{g}_j})^{+} (\mathbf{b}_d - \mathbf{b}_j)
\label{eqn:genotype_computation}
\end{equation}
where $\tilde{J}({\mathbf{g}_j})^{+}$ is the pseudo-inverse of $\tilde{J}({\mathbf{g}_j})$.  $\lambda$ is the Jacobian confidence coefficient. 

We iteratively solve Equation \ref{eqn:genotype_computation} and evaluate ${g}_{j+1}$ to get the corresponding behavior descriptor $\mathbf{b}_{j+1}$. 
These computations are repeated until the behavioral difference is smaller than the threshold or the number of iterations is greater than a predefined maximum value.

Note that the behavior descriptor used in QD may actually be larger than the state space to control. In the following experiments, $\mathbf{b}$ is actually composed of two parts, one that corresponds to the state to be controlled and another part aimed at obtaining a diversity in the actions to reach it. For example in the ball throwing task, the behavior descriptors is made up with the ball contact position on the ground and some intermediate positions of the robot end-effector. As the goal is to reach a particular state (no matter how), this second part is not taken into account in the proposed action computation process.


\subsection{Crossing the reality gap}
\label{sec:crossing_reality_gap}

\subsubsection{Single action adaptation}

QD algorithms are easy to parallelize as evaluations can be performed simultaneously on different cores or CPUs. Consequently, it is interesting to run QD algorithms in simulation. The same experiment run on a real robot would require days of tests. As discussed, it is difficult, if not impossible, to have a simulation that perfectly replicates all physical phenomena, in particular for frictions or collisions. The behavior observed on the real robot may be different from the behavior in simulation, creating a reality gap, which must be crossed to efficiently control a real robot.

Our approach to cross the reality gap for our repertoire-based control system consists in locally updating the genotypes in the repertoire generated in simulation, using tests on the real robot to locally reduce the reality gap and finally cross it.

Each time an action (with action $\mathbf{g}_a$) is executed on the real robot and produces a real world behavior ($\mathbf{b}_a$) that is far from the expected behavior ($\mathbf{b}_d$), a new action ($\mathbf{g}_d$) is computed to reduce the behavior difference $(\mathbf{b}_d - \mathbf{b}_a)$. 
$\mathbf{g}_d$ is computed with a local linearization of the Jacobian, as previously presented in section~\ref{sec:method_genotype_computation}: 
\begin{equation}
\label{eqn:crossing_reality_gap}
\mathbf{g}_d =  \mathbf{g}_a + \lambda \tilde{J}({\mathbf{g}_a})^{+} (\mathbf{b}_d - \mathbf{b}_a)
\end{equation}

\subsubsection{Repertoire update}
\label{sec:method_update}
As long as the action $\mathbf{g}_a$ is not in the repertoire, the trial is first added to the repertoire with the behavior descriptor corresponding to the observation $\mathbf{b}_a$. 



At the same time, with the assumption that actions with similar parameter vectors would be affected similarly by the observed reality gap, all other actions in the repertoire are updated according to the observed reality gap. Let $\mathbf{b}_{i}$ be the behavior associated to $\mathbf{g}_{i}$. $\tilde{\mathbf{b}}_{i,r}$ is the expected behavior after the updates associated to $r$ observed reality gaps. It is computed from $\mathbf{b}_{i}$ thanks to a compensation term $\eta_{i,r}$ evaluated iteratively after each observed reality gap:

\begin{equation}
\tilde{\mathbf{b}}_{i,r} = \mathbf{b}_{i} + \eta_{i,r}
\end{equation}

Initially, the compensation term for all actions is equal to zero: $\eta_{i,0}=\left[0, ..., 0 \right]$. $\eta_{i,r} $ is computed iteratively, taking into account the new reality gap observation weighted by a term taking into account the distance between the considered action's parameter vector $\mathbf{g}_i$ and that of the action that leads to the reality gap observation $\mathbf{g}_a$: 

\begin{eqnarray}
\eta_{i,r}  &=& \eta_{i,r-1} + w_{i,r} \left( \tilde{\mathbf{b}}_{i,r-1} - \mathbf{b}_{a} \right)\\
w_{i,r}     &=& \exp(-0.5 \left(\mathbf{g}_i - \mathbf{g}_a \right)^{T} \mathbb{D}_r \left(\mathbf{g}_i - \mathbf{g}_a \right))
\end{eqnarray}
where 
$\mathbb{D}_r$ represents a distance metric that determines the shape of local linear validity near the action $\mathbf{g}_a$.
In this study, we set the kernel matrix as $\mathbb{D}_r = \lambda I$, where $\lambda$ is a local linearity confidence. 
The amplitude of the correction is higher for actions that are close to $\mathbf{g}_a$, the action that lead to the reality gap observation, in the action parameter space.



\section{Experimental validation}
\label{sec:experiment}

Two different robot arm applications have been considered to test the proposed method:
(1) learning to throw a ball into a basket and 
(2) learning to control a joystick to a desired direction with a desired angle. 

The only differences between the setups associated to these two experiments are the objects in the environment (ball or joystick on a table), the behavior descriptors and the quality functions. The following sections describe each implementation in detail.
%

\subsection{Setup}
\label{sec:general-setup}

The following experiments were performed on a Baxter research robot with two arms, each with 7 degrees of freedom (DOF) plus a 1-DOF gripper. For simplicity, only the right arm is used. The simulator used during the QD search is DART\footnote{\url{http://dartsim.github.io/}} with the FCL collision detector\footnote{\url{https://github.com/flexible-collision-library/fcl}}. The simulation time step is set to \SI{0.002}{\second}. 

In both experiments, the motion planning used a simple open-loop controller: the robot always started from a fixed initial pose with zero velocity, each action is defined by the joint-space position and velocity of the arm at the end of the motion, plus the duration of the motion, and a smooth arm trajectory is generated by fitting a third order polynomial with those final and initial states. Our action space is therefore continuous and 15-dimensional (final position and velocity of each of the 7 DOF, plus motion duration). This 15-dimensional space also constitutes our genotype space for the QD search. 
Note that the planned motion ends at the ball releasing instant for the ball throwing experiment. Hence, in order to stop its motion smoothly, an additional motion is followed right after the action. As the additional motion works after the ball is released, it does not affect the QD result.
This simple controller is used as it provides a very simple and compact representation allowing a smooth motion generation and yet is sufficient to solve our chosen tasks. But neither the QD-search nor the reality gap crossing technique depend on this specific action representation; other more advanced action representations such as, for instance, DMP~\cite{ijspeert_trajectory_2001}, DS-GMR~\cite{khansari-zadeh_learning_2012} or splines could be used.

%
%


\subsection{Ball throwing experiment}
The purpose of this experiment is to build a repertoire that contains diverse and energy-efficient robot motions for throwing a ball into a basket placed at an arbitrary position on the ground. In order to use the proposed QD algorithm, three main components have to be defined: the parameter vector (or genotype) that defines the action, the behavior descriptors that describe the behavior space to be explored, and the quality function. 

The actions are defined by a 15-dimensional vector parameterizing the controller described in section~\ref{sec:general-setup}. The space we wish to explore is the space of ball contact points with the basket plane, i.e. the possible intersection points of the ball trajectory with the plane containing the top of the basket, which will be referred in the following as the \textit{target position}. This is the space that the proposed QD search will learn to control. To add more diversity in the generated actions and generate several trajectories to reach the same target position, the behavior descriptor used in QD search contains the target position, but also a rough sampling of the gripper trajectory during the robot motion: three intermediate positions of the robot gripper are recorded and added to the descriptor. The behavior descriptor is therefore 11-dimensional (the two-dimensional target position of the ball in the basket plane, plus three three-dimensional samples of the gripper position.)
 
The quality function is independent of the task and simply penalizes energy consumption (sum of the torques applied to the robot joints during motion).
\begin{equation}
l(\mathbf{g}_i) = - \frac{1}{T_i} \sum_{t=1}^{T_i} ||\tau(\mathbf{g}_i)_t|| 
\label{eqn:lq}
\end{equation}
where $t$ represents the time step index starting at $1$ and ending at the motion duration $T_i$ of an action $\mathbf{g}_i$; $||\tau(\mathbf{g}_i)_t||$ represents a norm of the torques applied to the robot at time index $t$.

The QD search algorithm is run for \num{2000} generations. Indeed, larger number of generations obviously provide better results in terms of quality and diversity both. In this experiment, however, we have limited the number of generations so as to complete the QD search within a day using a machine in the lab.  
In each generation, the \num{240} actions with the highest novelty score are selected in the repertoire to be the parents of the next generation. If the repertoire has less than \num{240} actions, the missing actions are randomly generated. Crossover (crossover rate: \num{0.10}) and mutation (mutation rate: \num{0.20}) operators are applied to generate \num{240} offspring. Actions exhibiting self-collisions and joint limits violations (in position, velocity and acceleration) are discarded. 

A random motion generation method has been considered as a control experiment to show the efficiency of the QD search. It follows the same process as QD search (as described in Algorithm~\ref{algo:qdsearch}), but the evolutionary process of QD, including selection, mutation and crossover, is replaced by a random genotype generation.

\subsubsection{Step 1 results: QD search performance}
\label{sec:qd_search_result}

\begin{figure}[h!]
\center
	\begin{subfigure}{.44\textwidth}
	\centering
	\includegraphics[width=7Cm]{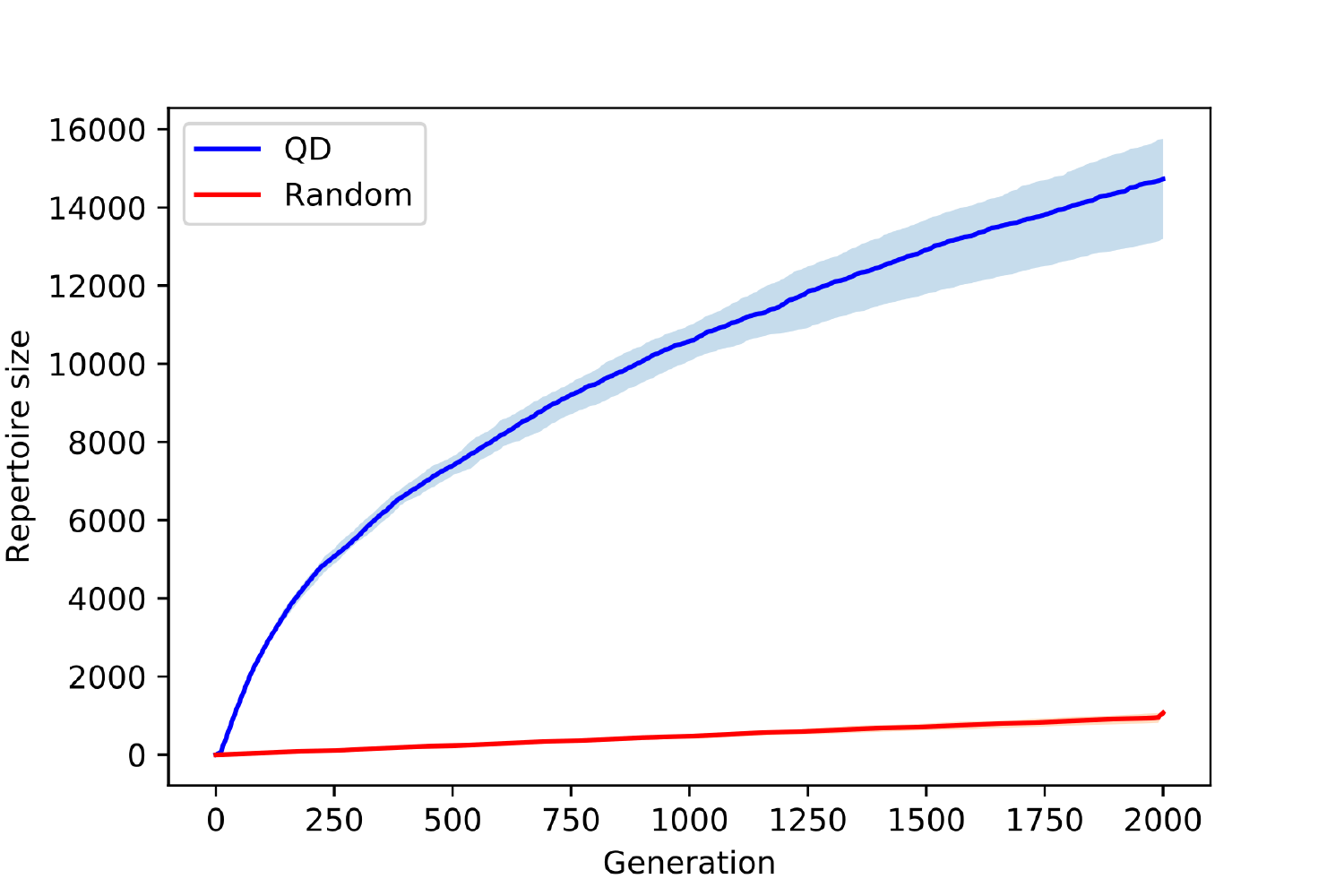}
	\caption{Repertoire size}
	\label{fig:nbsolution_vs_gen}
	\end{subfigure}
	\hspace{.1\textwidth}
	\begin{subfigure}{.44\textwidth}
	\centering
	\includegraphics[width=7Cm]{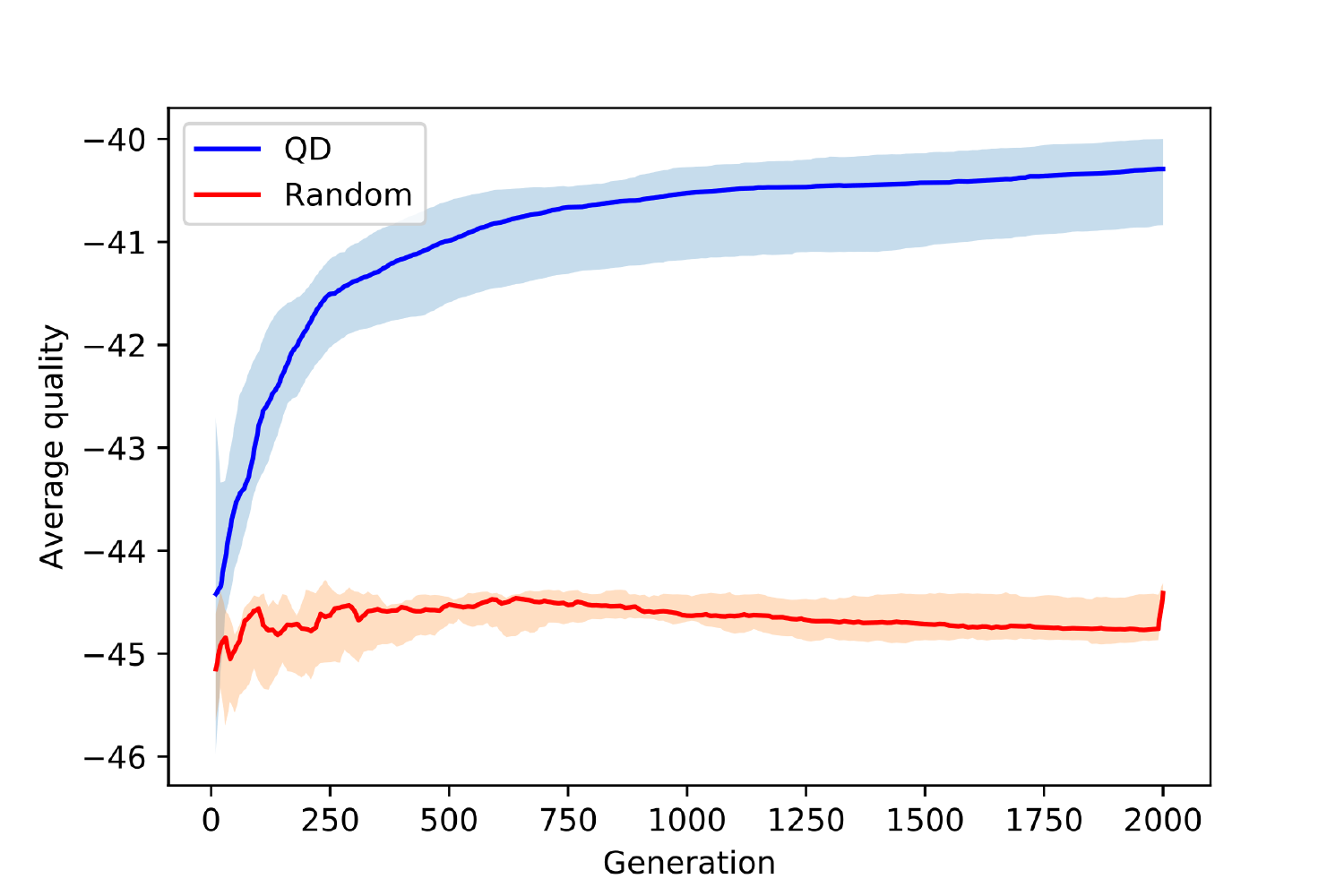}
	\caption{Quality score}
	\label{fig:fitness_vs_gen}
	\end{subfigure} 
\caption{Number of solutions (a) and average quality of those solutions (b) found by QD search and uniform random search on the ball throwing scenario. The median value and interquartile interval on the 26 runs are shown. The QD search finds solutions much more efficiently, in terms of not only the amount of solutions (a) but also the average quality (b).}
\label{fig:nbsolutions_vs_gen}
\end{figure}

\begin{figure}
\center
\begin{subfigure}{.40\textwidth}
\includegraphics[width=5Cm]{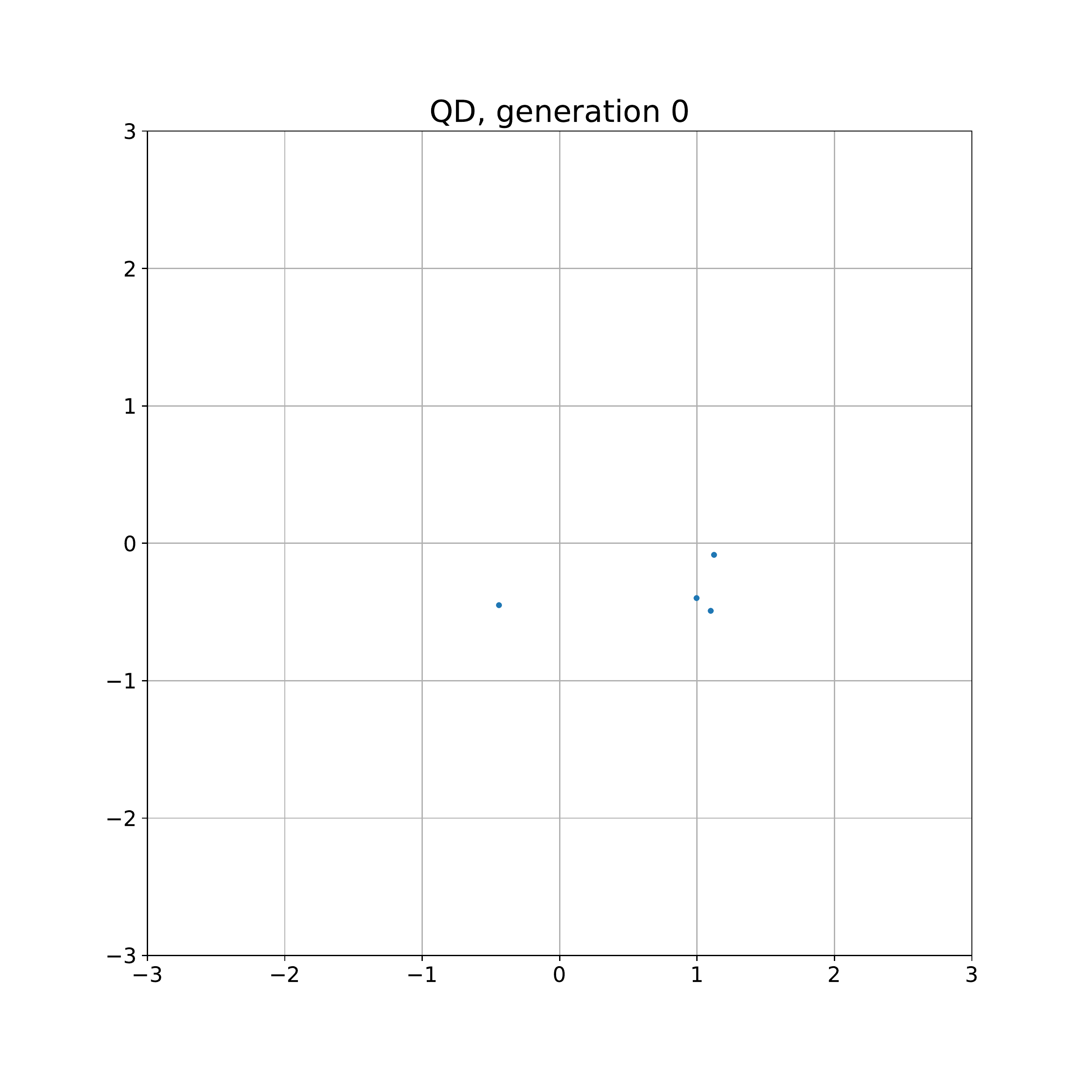}
\end{subfigure}
\hspace{.05\textwidth}
\begin{subfigure}{.40\textwidth}
\includegraphics[width=5Cm]{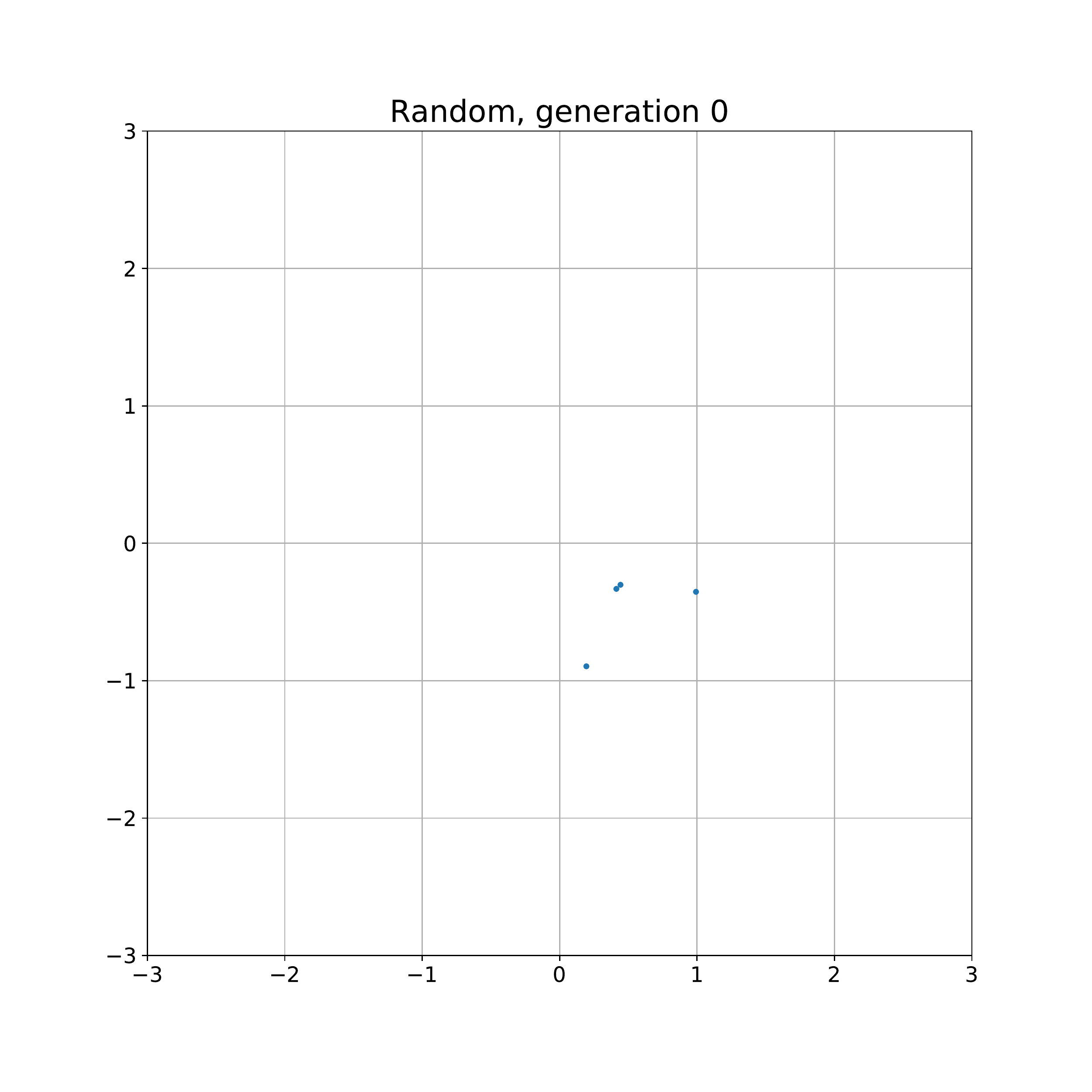}
\end{subfigure}
\\
\begin{subfigure}{.4\textwidth}
\includegraphics[width=5Cm]{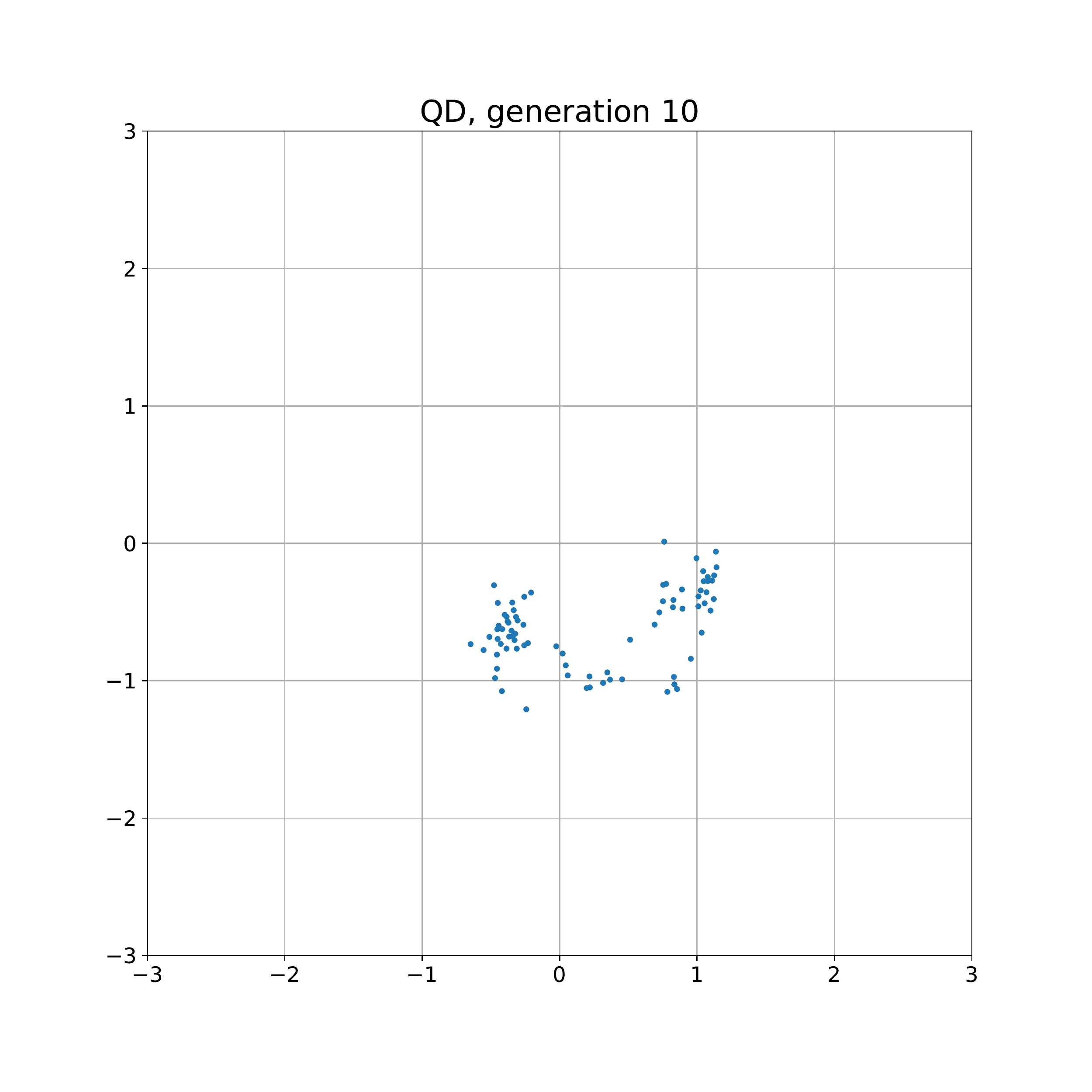}
\end{subfigure}
\hspace{.05\textwidth}
\begin{subfigure}{.4\textwidth}
\includegraphics[width=5Cm]{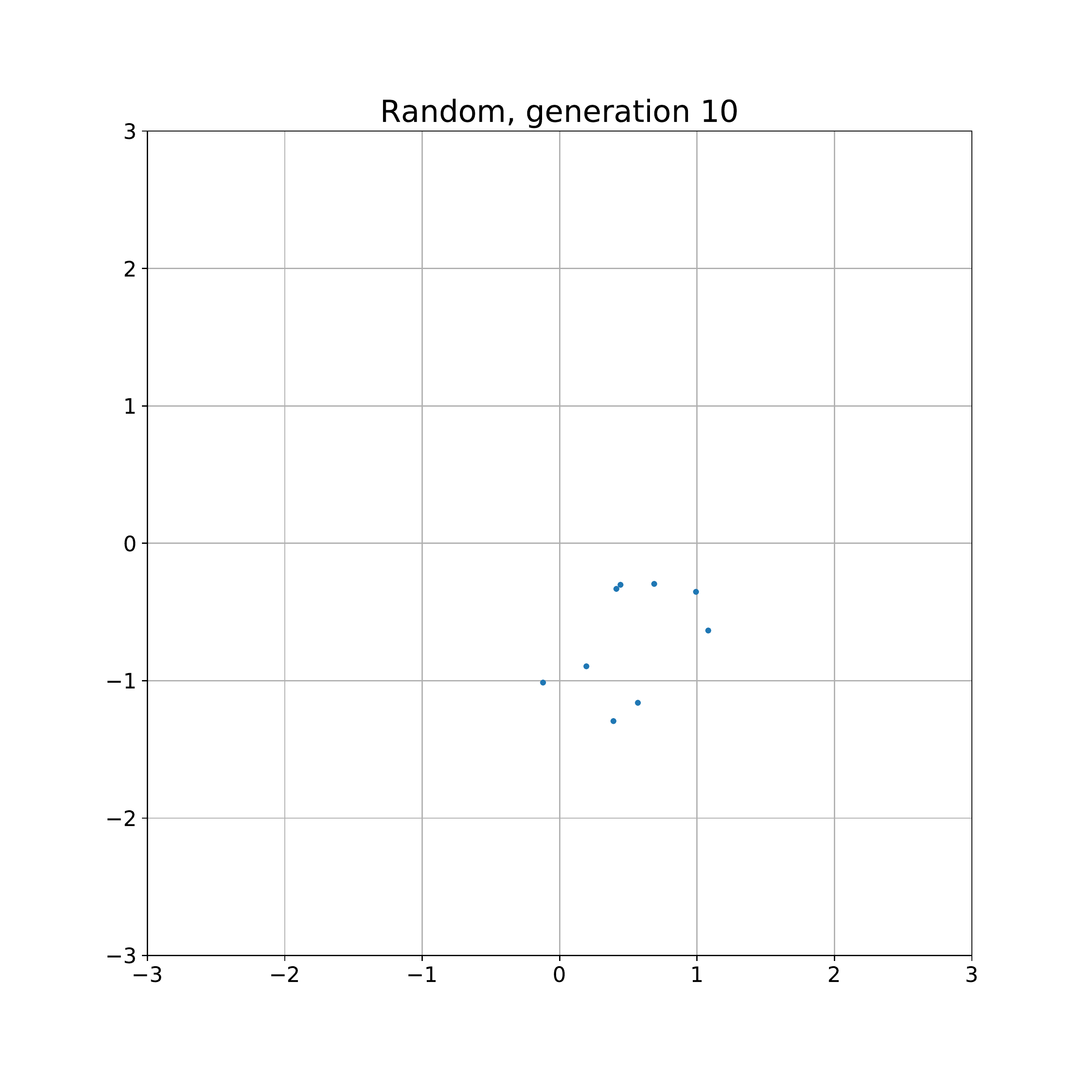}
\end{subfigure}
\\
\begin{subfigure}{.4\textwidth}
\includegraphics[width=5Cm]{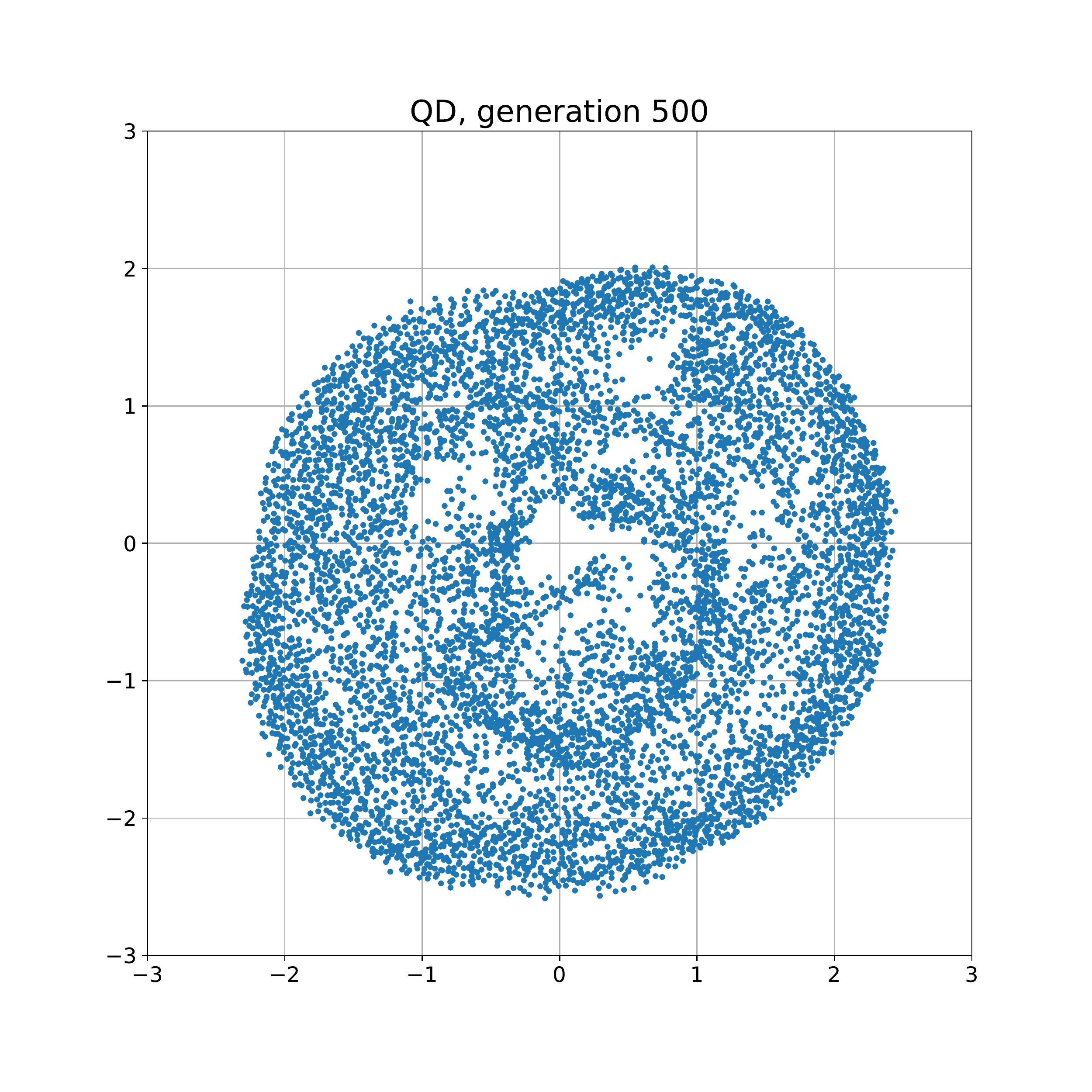}
\end{subfigure}
\hspace{.05\textwidth}
\begin{subfigure}{.4\textwidth}
\includegraphics[width=5Cm]{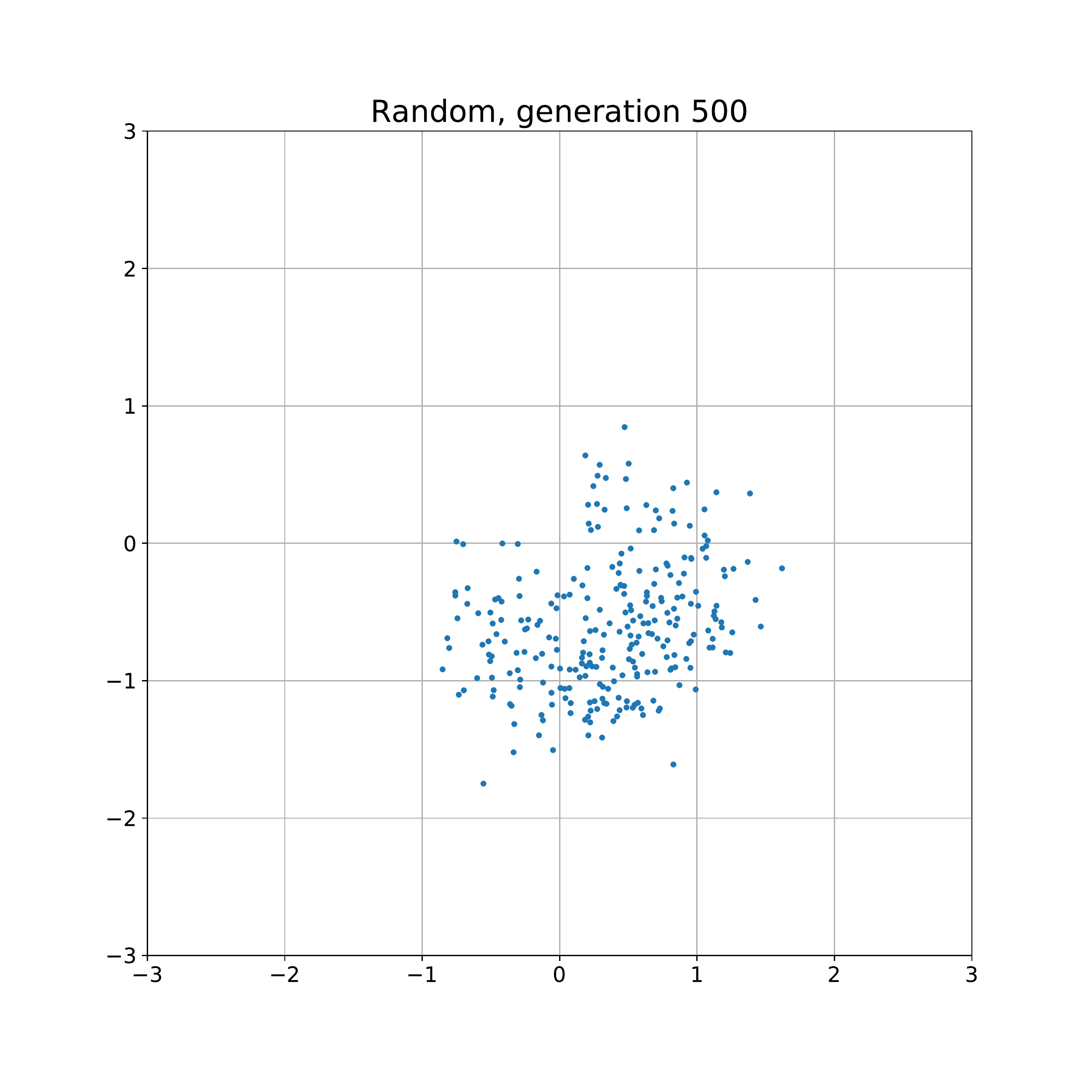}
\end{subfigure}
\\
\begin{subfigure}{.4\textwidth}
\includegraphics[width=5Cm]{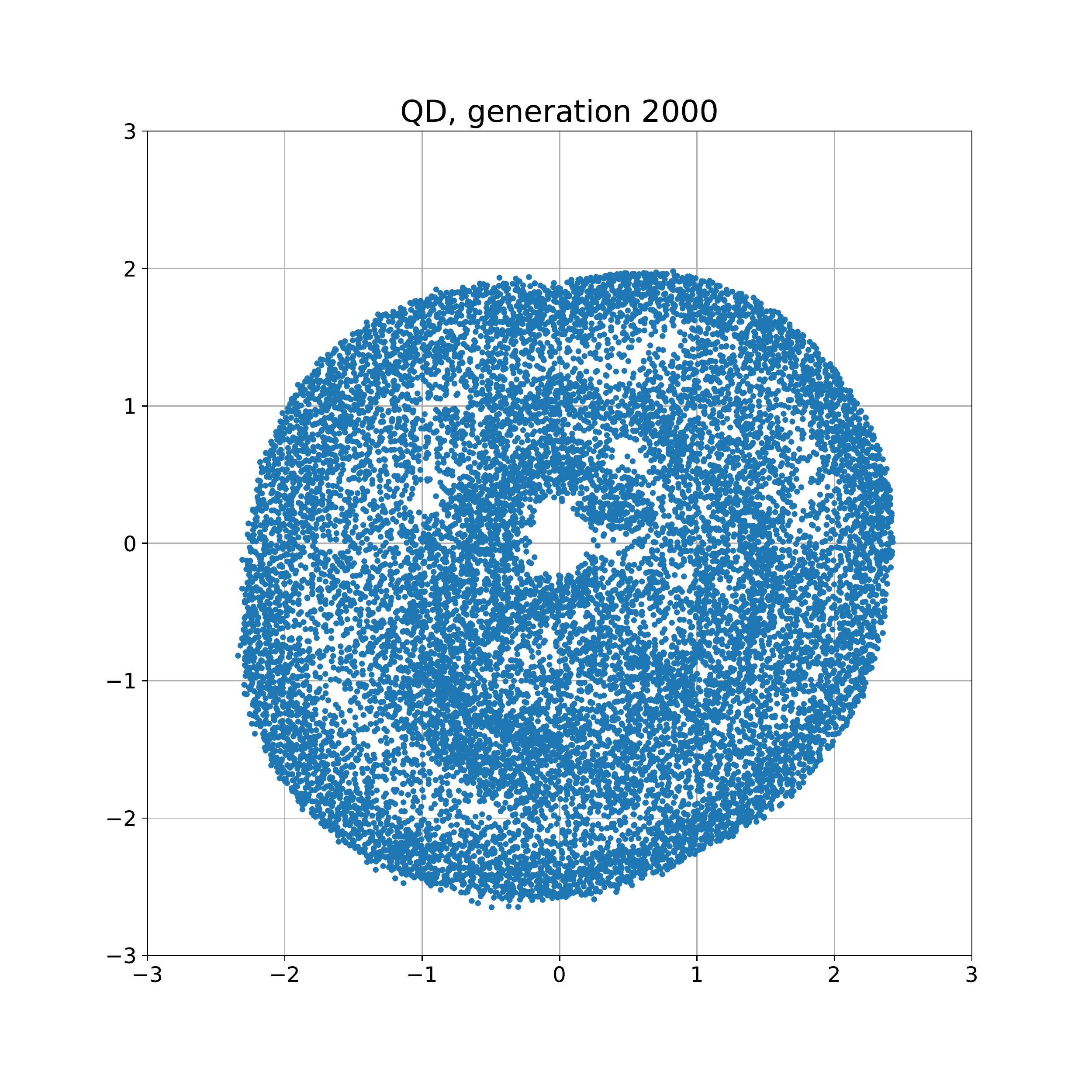}
\end{subfigure}
\hspace{.05\textwidth}
\begin{subfigure}{.4\textwidth}
\includegraphics[width=5Cm]{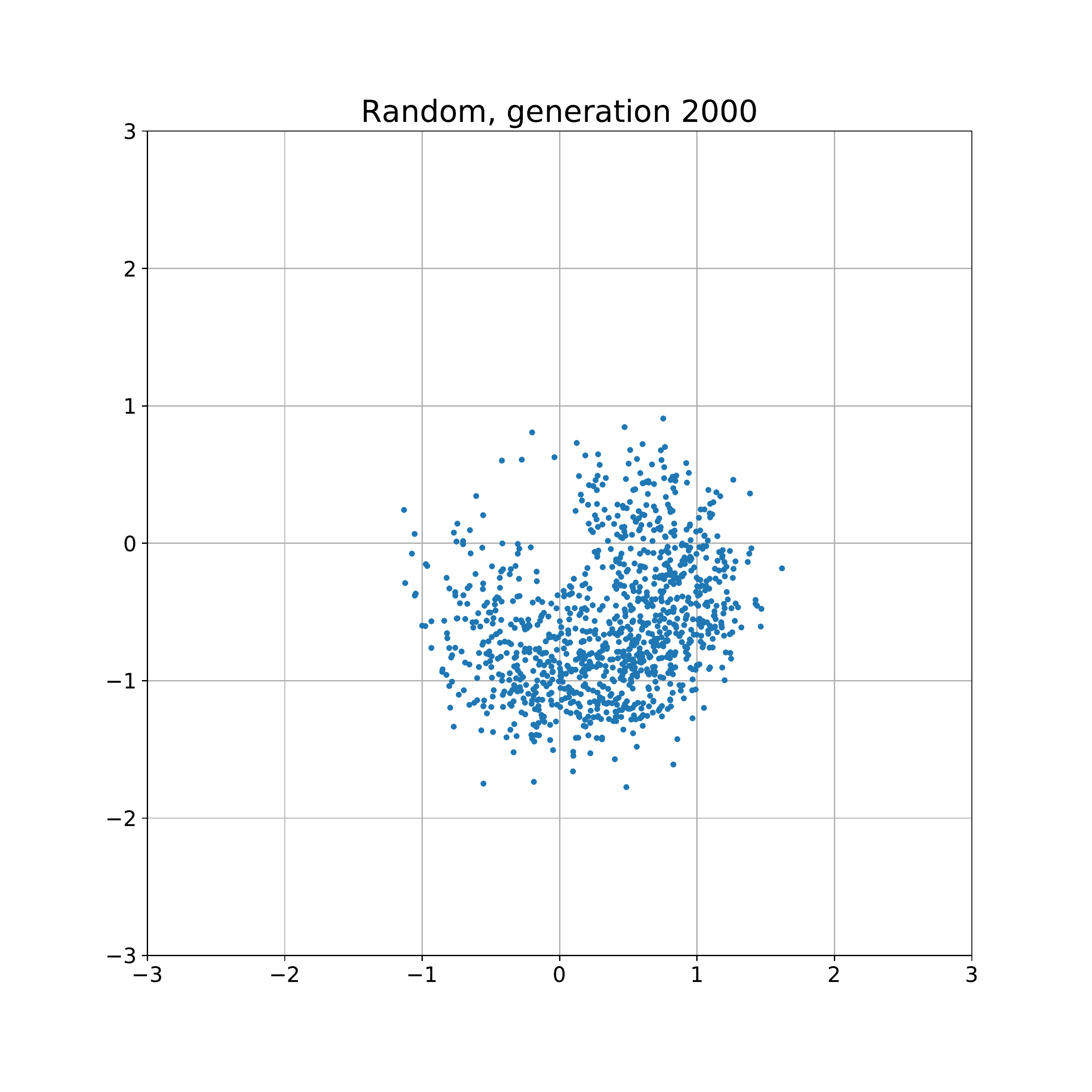}
\end{subfigure}

\caption{Solutions found during QD search (left) and with random parameter generation (right). Target positions in the 2D plane are shown. Dense area means that there are multiple diverse solutions to throw the ball at a given target basket position. A video showing this process is available on \url{https://youtu.be/NssH9ytU4Bs}.}
\label{fig:qd_result}
\end{figure}

\begin{figure}[t]
\center
	\includegraphics[width=13Cm]{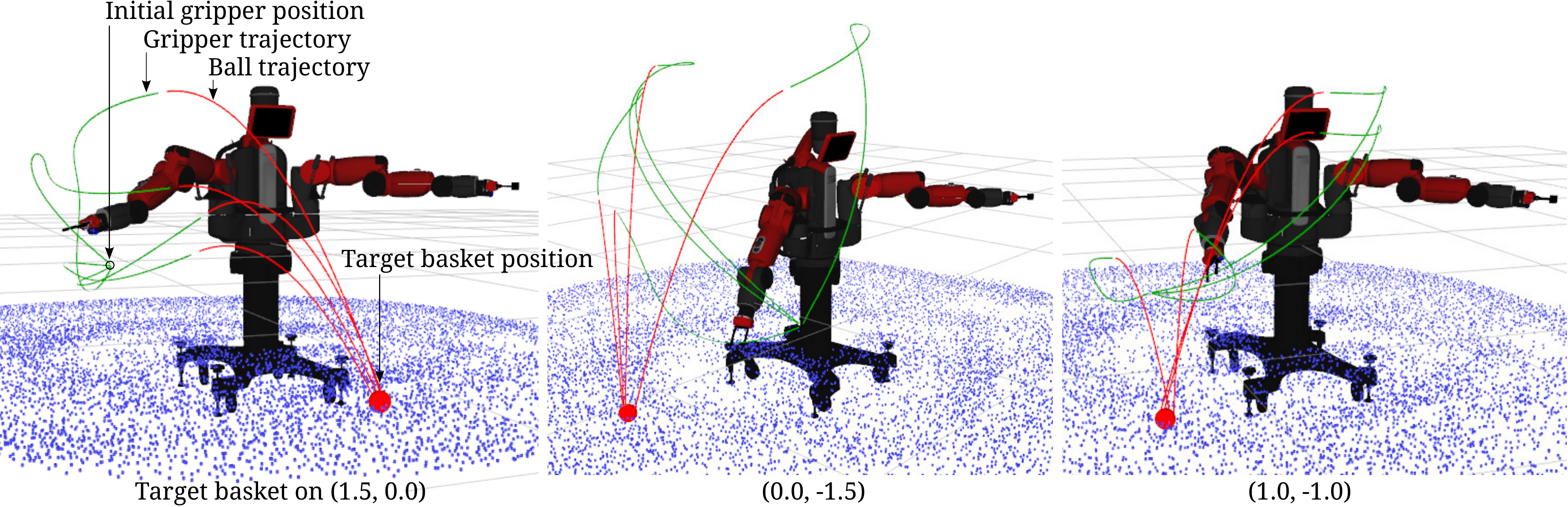}
\caption{Example of diverse trajectories found by QD search for different basket positions.}
\label{fig:qd_3d_result}
\end{figure}

The QD search process was run 26 times. The median and interquartile range of the number of solutions are shown in Figure~\ref{fig:nbsolutions_vs_gen}. The QD algorithm gradually finds actions to reach a more diverse set of states and with a higher quality score than a random genotype generation. After \num{2000} generations, the QD search finds \num{14473 \pm 1619} actions, whereas the random generation only finds \num{1085 \pm 26}. Furthermore, as shown in Figure~\ref{fig:qd_result}, although randomly generated genotypes are uniformly distributed in the feasible genotype space, they do not generate equally distributed behaviors in the feasible behavior space. 
Figure~\ref{fig:qd_3d_result} shows examples of diverse trajectories to throw the ball at different locations.

\begin{figure}[h!]
\center
	\includegraphics[width=.18\textwidth]{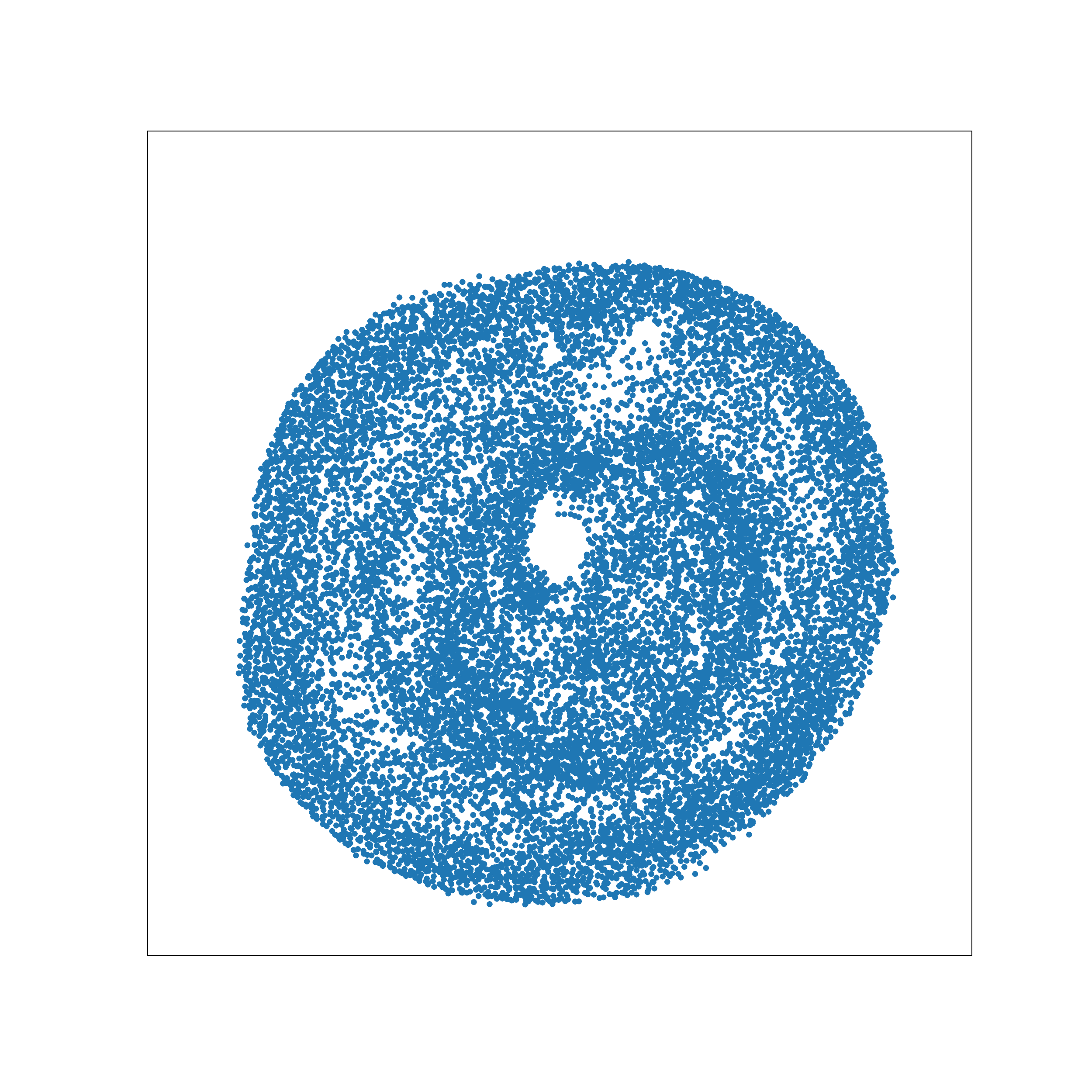}
	\hspace{.01\textwidth}
	\includegraphics[width=.18\textwidth]{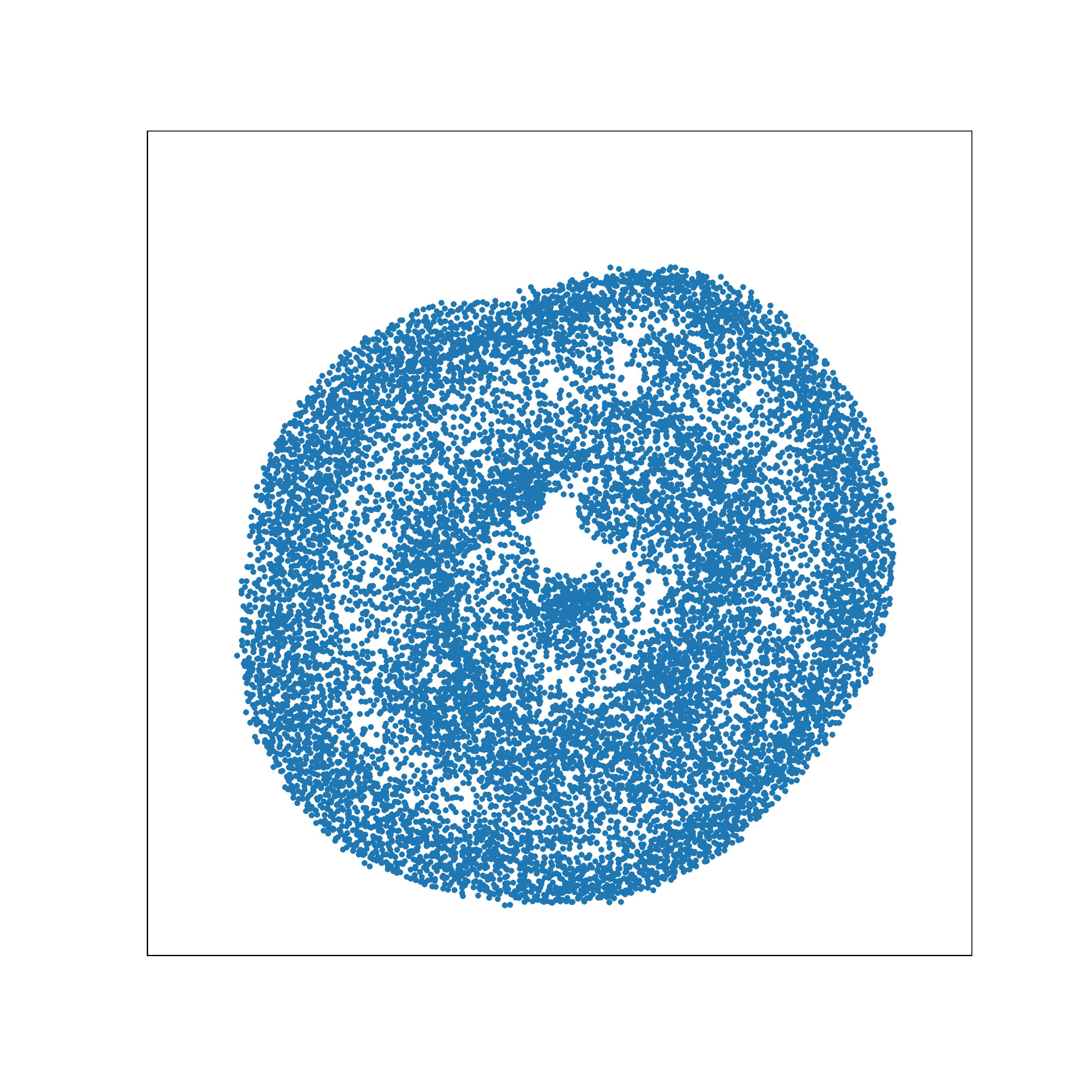}
	\hspace{.01\textwidth}
	\includegraphics[width=.18\textwidth]{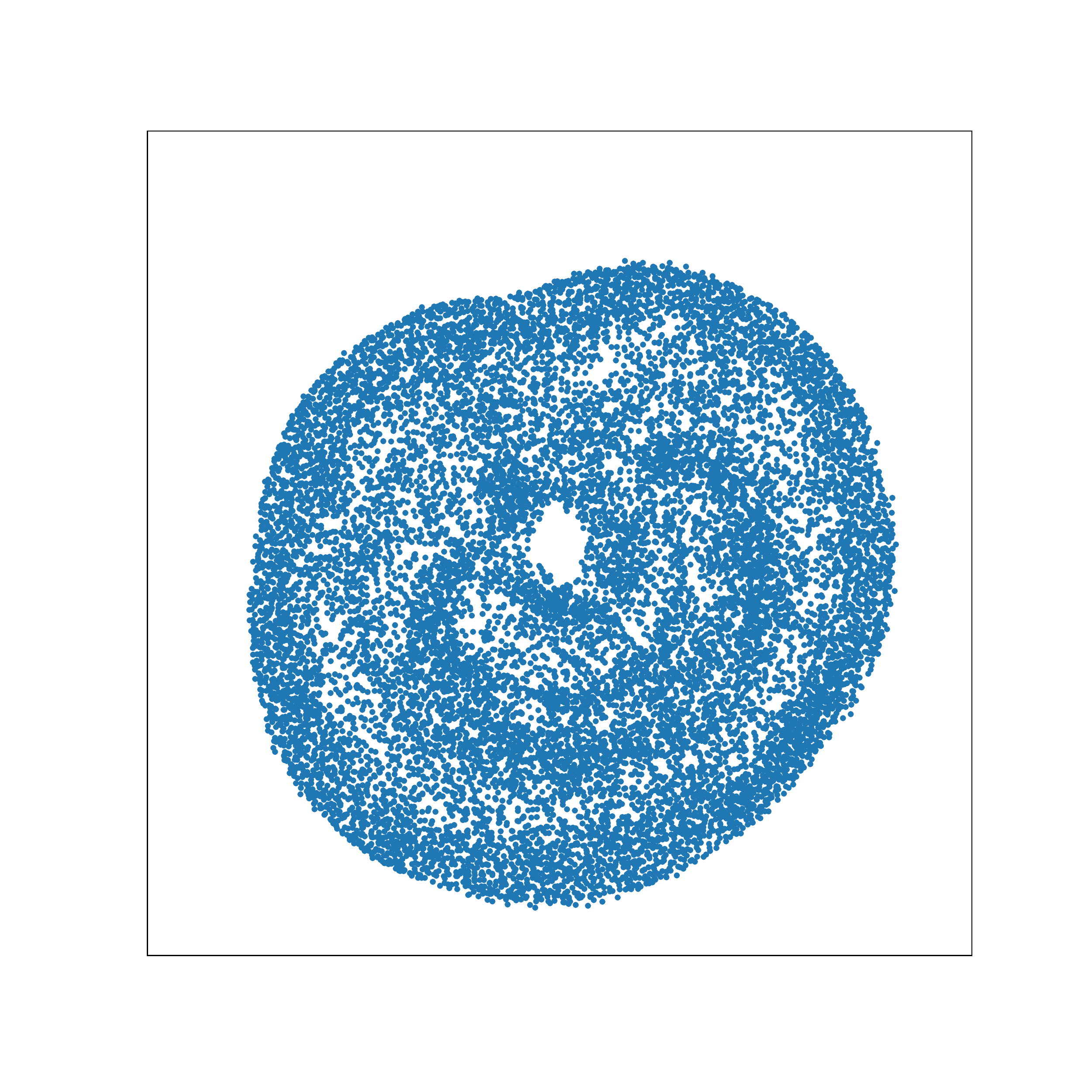}
	\hspace{.01\textwidth}
	\includegraphics[width=.18\textwidth]{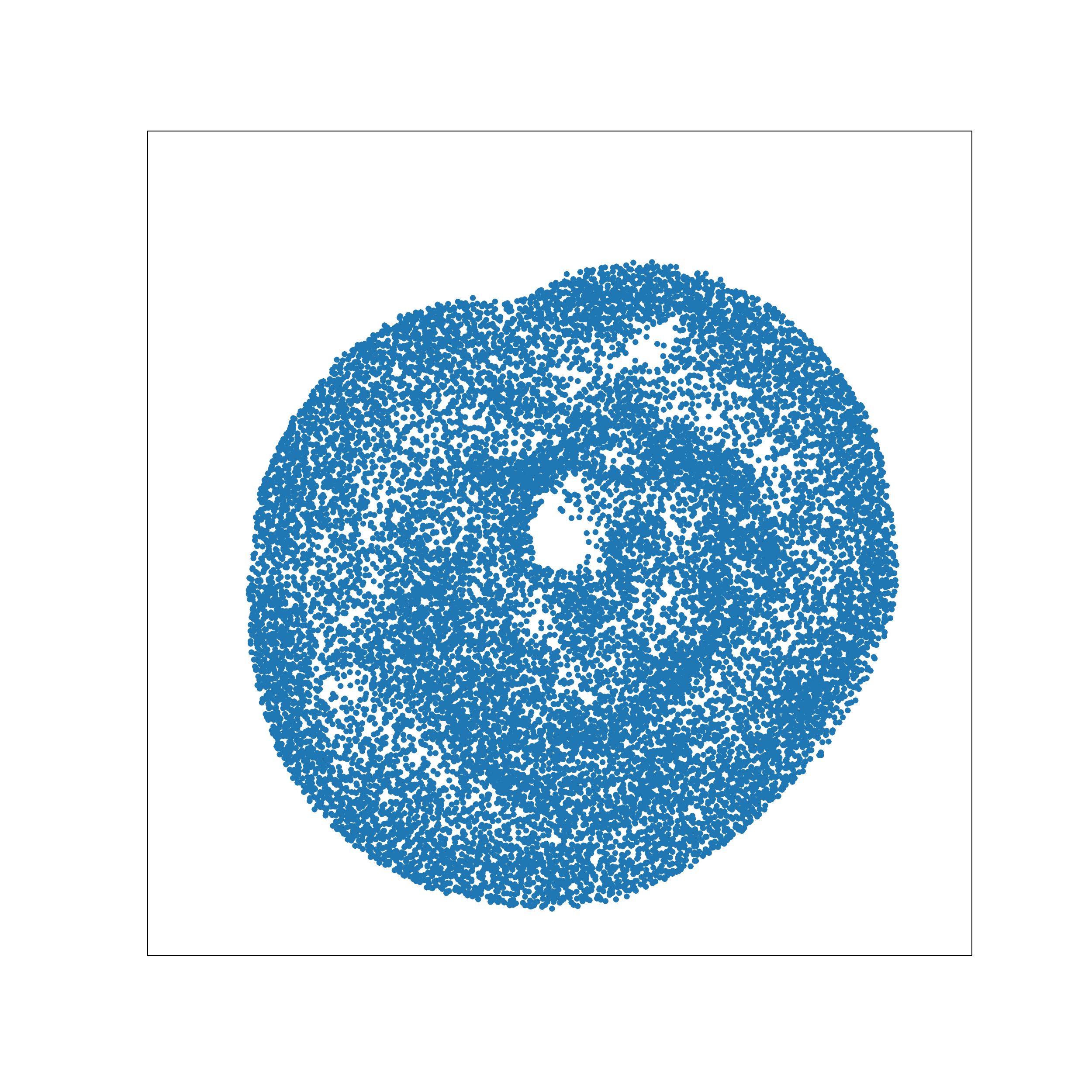}
	\hspace{.01\textwidth}
	\includegraphics[width=.18\textwidth]{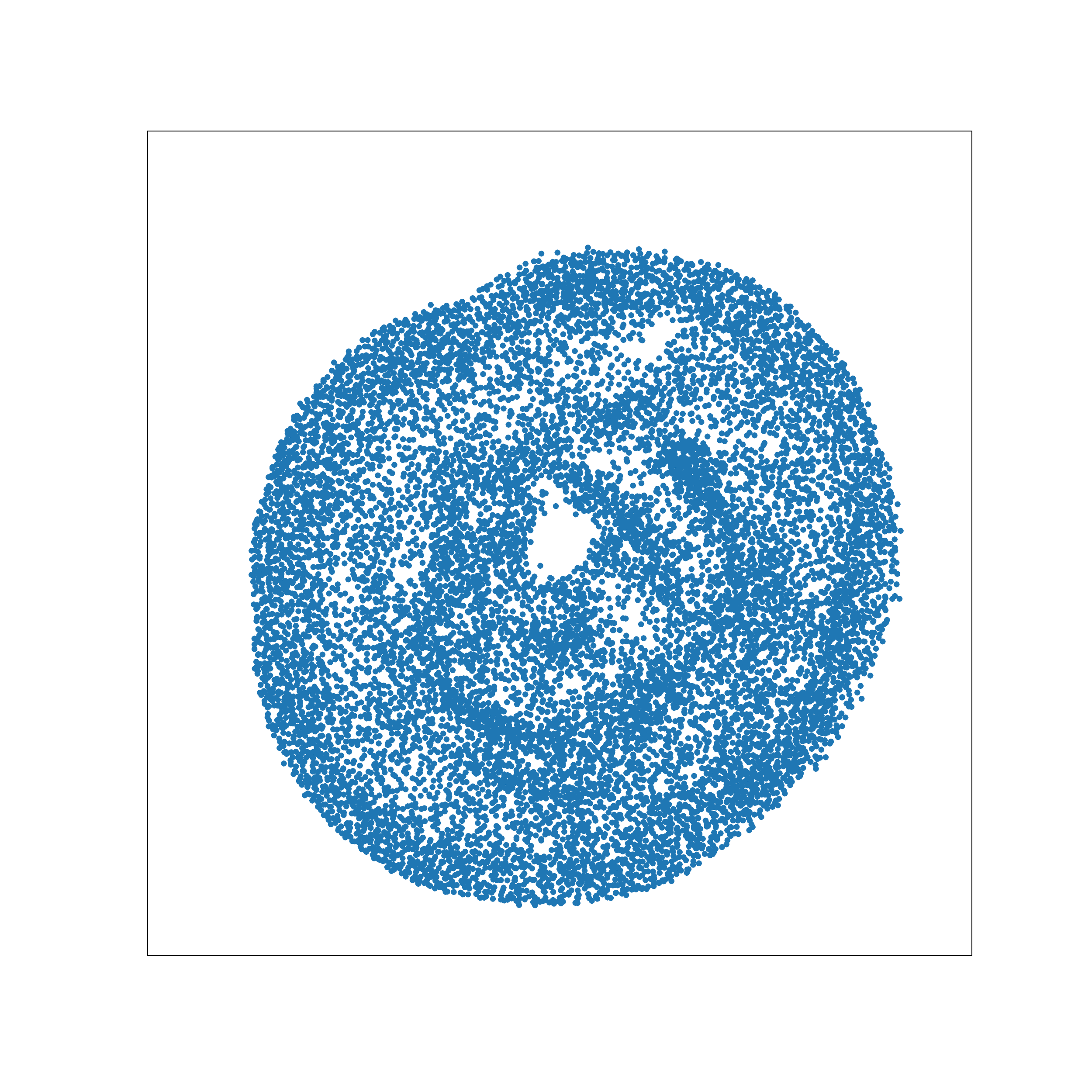}
\caption{Illustration of 5 of the 26 generated repertoires after 2000 generations of the QD algorithm. All generated repertoires cover the whole goal space reachable by robot motions.}
\label{fig:consistency-throw}
\end{figure}

The results in the next sections show the performance of the second step of the proposed approach: a generated repertoire is exploited with a local adaptation of the actions it contains to deal with the sampling and the reality gap problems. In the following, the repertoire used is the one containing the median number of individuals among the runs. However, as illustrated by Figure~\ref{fig:consistency-throw}, all the generated repertoires satisfyingly cover the whole reachable goal space: the choice of the individual repertoire used for control is therefore not sensitive, and any other repertoire could hve be used.
 
\subsubsection{Adaptation to arbitrary basket positions}


To test the ability of the proposed approach to deal with the sampling problem, 1000 basket positions are randomly generated and the ability of the robot to throw to these different target positions is evaluated. The results presented in this section have been obtained in simulation (experiments on the real robot are described in section~\ref{sec:real_robot}).

\begin{figure}
\center
\includegraphics[width=5Cm]{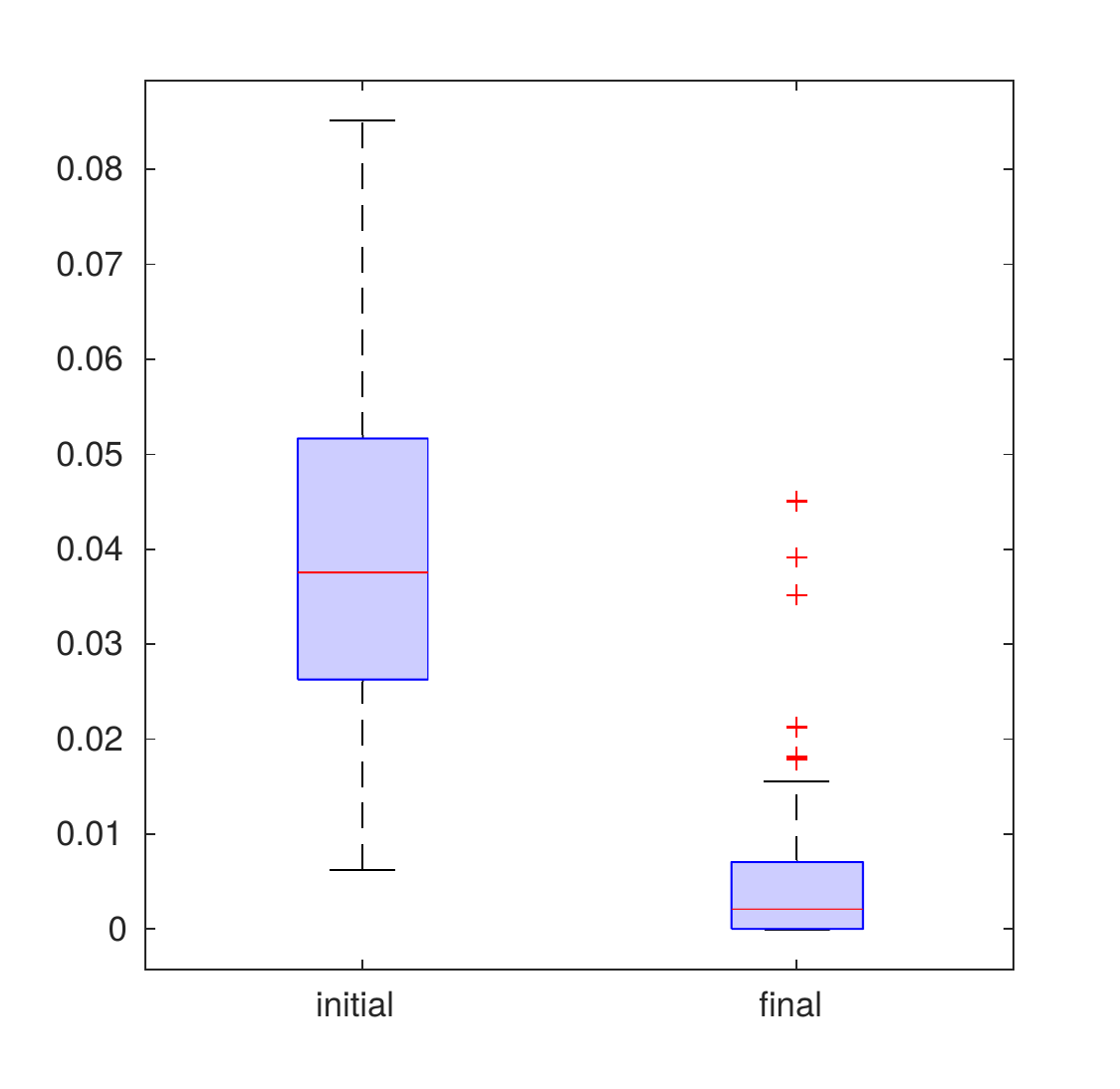}
\caption{Average distance of the ball to the target position (in meters) when trying to throw the ball at an arbitrary position. An average error of \SI{3.84 \pm  1.63}{\centi\meter} is observed if the robot simply uses the nearest actions in the repertoire (left). If the proposed approach is used to compute a new action based on a local linear model (right), the error is reduced to \SI{0.52 \pm  0.82}{\centi\meter}.}
\label{fig:result_generalization_error}
\end{figure}

\begin{figure}
\center
\includegraphics[width=\linewidth]{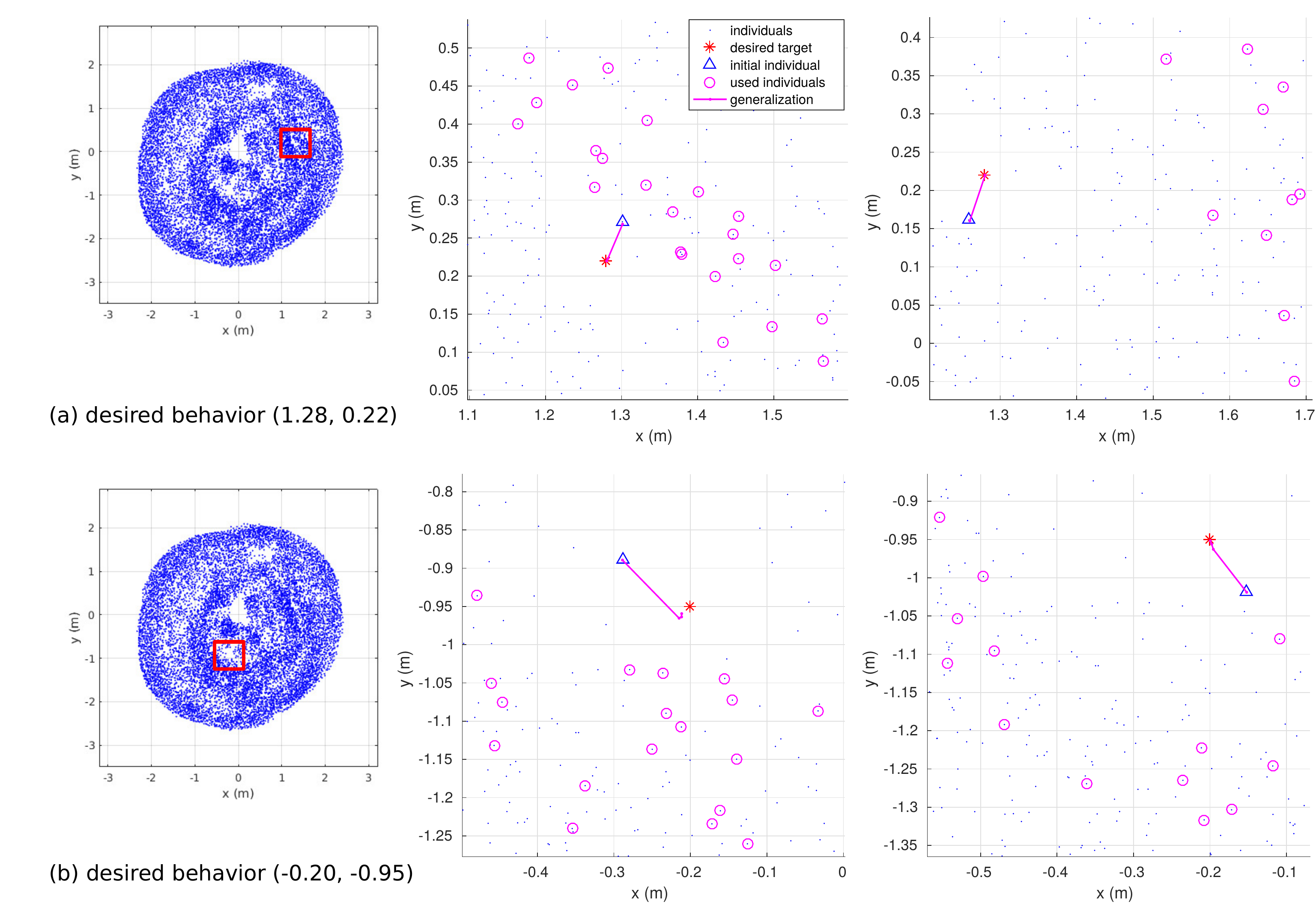}
\caption{Examples of action adaptations for two different desired behaviors. The blue dots represent the behaviors of actions in the repertoire. For a given desired behavior (red star), the presented method iteratively computes new actions to approach the target. The approach starts from an initial action (blue triange) and computes the local Jacobian on the basis on the nearest actions in the \textit{action parameters space}, which are the actions surrounded by pink circles. Uncircled actions are close in the behavior space but not in the action parameters space (they reach a close position, but with a very different trajectory) and are ignored. The local Jacobian is then use to compute a new action (pink path) closer to the target position. This process is iterated until the target position is reached. The approach is tested on two different target behavior space positions, and with two different starting actions for each target position, to show its robustness.}
\label{fig:result_generalization_2d}
\end{figure}

As described in section~\ref{sec:method_genotype_computation}, the actions throwing the ball to the closest positions to the target are selected in the repertoire and they are used as initial starting points of the local linearization approach. 
Figure~\ref{fig:result_generalization_error} shows the average ball throwing error of the 5 closest actions in the repertoire before and after 4 iterations of the proposed gradient descent approach. After 4 iterations of gradient descent adaptation, the new actions have reduced the error from \SI{3.84 \pm  1.63}{\centi\meter} to \SI{0.52 \pm  0.82}{\centi\meter}. The accuracy of the behavior can be further increased by adding iterations of the gradient descent method.
Two examples are shown in Figure~\ref{fig:result_generalization_2d}.

\subsubsection{Crossing the reality gap}
\label{sec:result:crossing_reality_gap}

To test the method proposed to cross the reality gap, mis-configurations of the physics simulator have been introduced between step 1 and step 2. Two mis-configurations have been considered. The first one adds an offset to the robot joints: \SI{0.05}{\radian} (\SI{2.87}{\degree}) were added on the two joints of the robot shoulder. For the second mis-configuration, a \SI{50}{\milli\second} delay has been added to the gripper control (the robot opens its gripper \SI{50}{\milli\second} after the desired releasing time).

The selected repertoire contained \num{11600} actions that behave correctly in the original robot configuration. After applying the artificial mis-configuration in the simulator, an average error of \SI{11.3}{\centi\meter} on the ball contact position is observed, and \num{9331} (around \SI{80}{\percent}) actions of the repertoire launch the ball at a position that is at more than \SI{5}{\centi\meter} from the expected contact point. 

\begin{figure}[h]
\center
\includegraphics[height=5Cm]{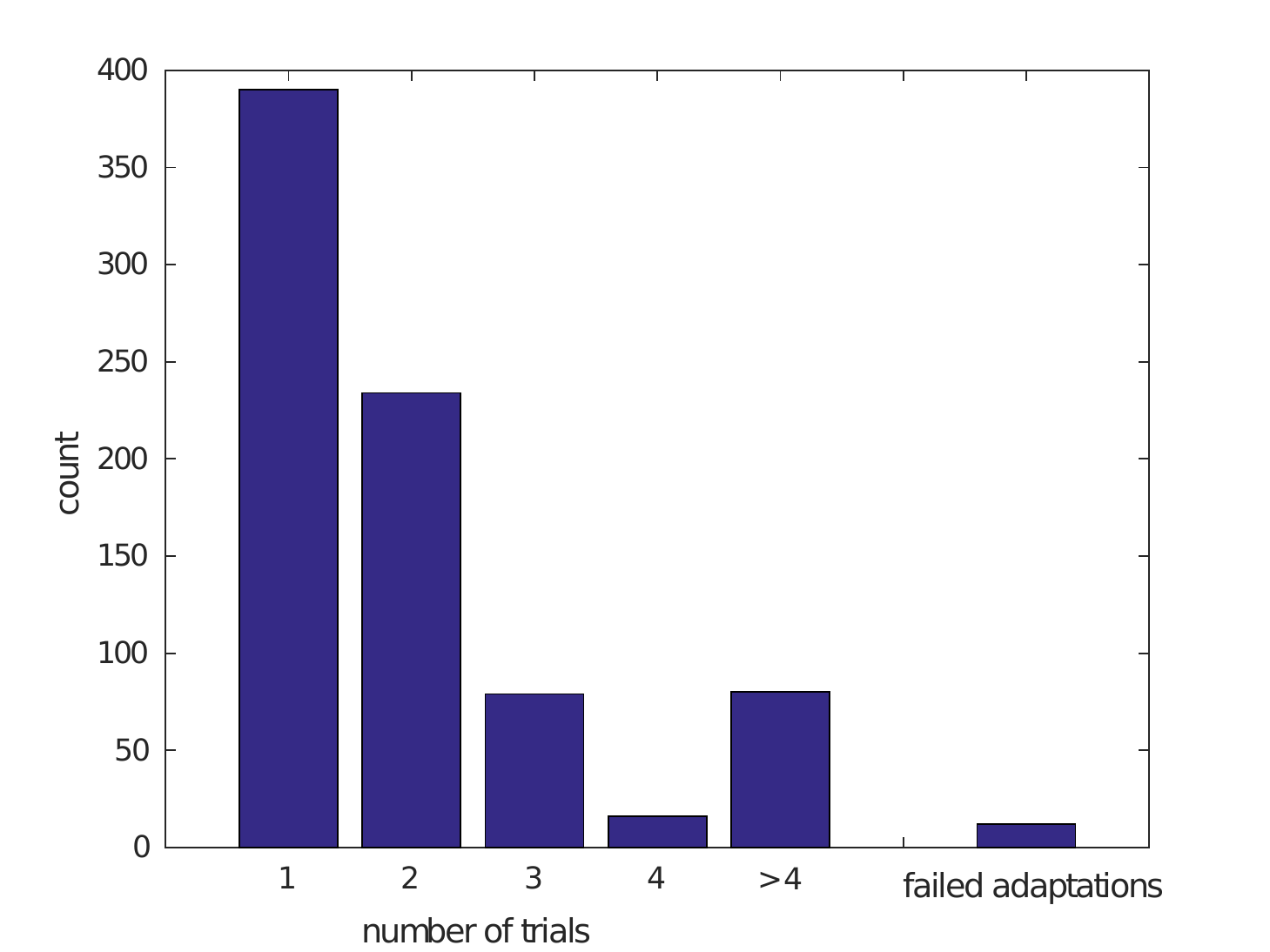}
\includegraphics[height=5Cm]{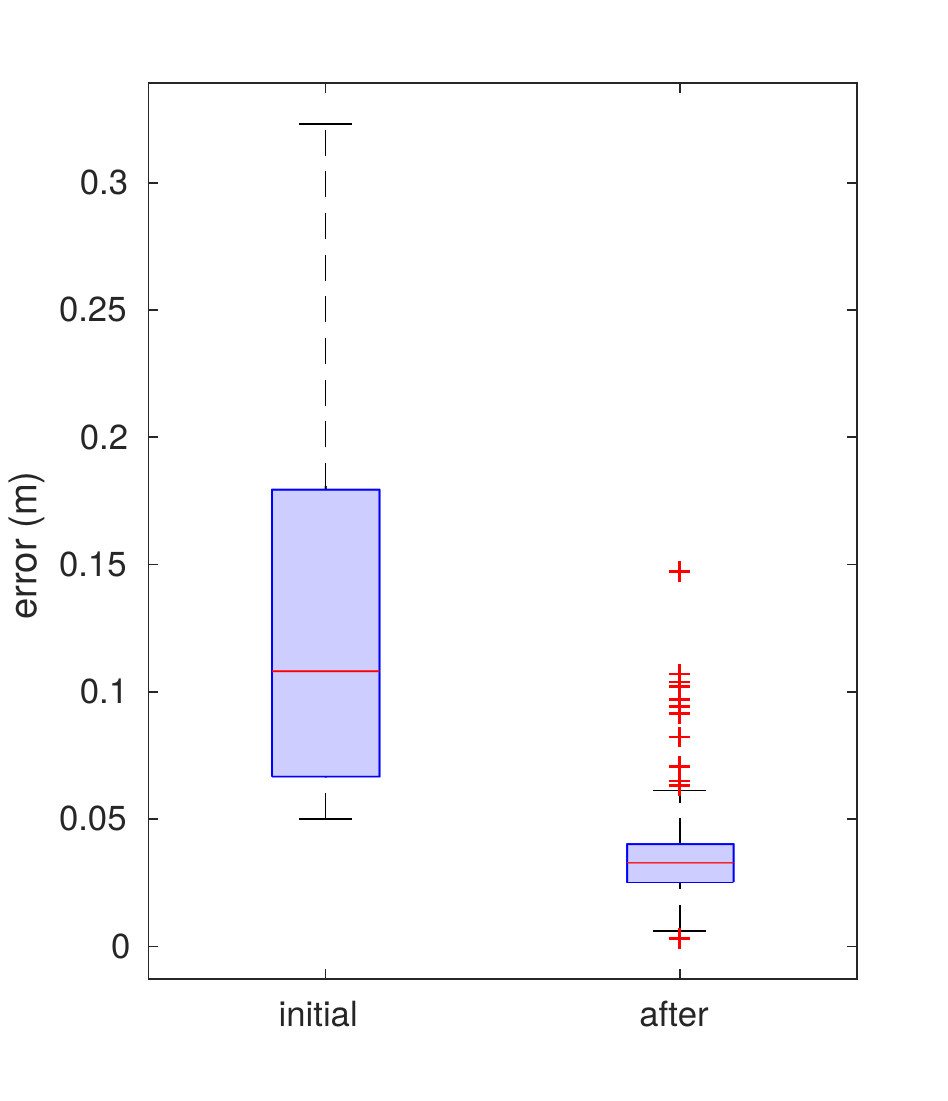}
\caption{Left: histogram of the number of required iterations to cross the reality gap in the ball throwing experiment. Right: reality gap in the 1000 selected actions before and after applying the proposed adaptation approach.}
\label{fig:crossing_reality_gap_no_update}
\end{figure}

To test the method, actions are randomly selected in the repertoire and adapted to the mis-configured simulator with the proposed process (without repertoire update). For each trial, the method is iterated until the reality gap goes below a predefined threshold or exceeds the maximum iteration number; in this experiment, they were set, respectively, to \SI{5}{\centi\meter} and \num{5} iterations. These two values are task specific parameters that users should select. We selected \SI{5}{\centi\meter} so that the thrown ball would fall correctly into the basket with some margin. We limited maximum number of iteration to \num{5} because most of the reality gaps are crossed with \num{4} iterations as we see in Figure \ref{fig:crossing_reality_gap_no_update}.
 
\num{1000} actions were randomly selected in the repertoire and evaluated in the mis-configured simulator. Among them, \num{189} actions did not require any adaptation as their behavior difference was already below the threshold. \num{811} generated an error above the threshold. Most of them (\num{719} actions) were adapted to the new configuration within 4 iterations of the proposed adaptation process (Figure \ref{fig:crossing_reality_gap_no_update}), and for \num{80} others, the behavior did approach the target basket position, but did not go below the error threshold within 4 iterations.

\num{12} actions did fail to approach the target behavior, for three different causes. For \num{9} actions, collisions occured between the robot and the ball, creating complex non-linearities that the method was not able to deal with. For \num{2} others the local jacobian was not locally linear enough, and in \num{1} case the robot motions for the new actions exceeded the joint position limits.

\subsubsection{Repertoire update}

Whenever a reality gap is observed, the corresponding action and its surrounding neighbors are updated (as described in section~\ref{sec:method_update}). This update process is evaluated in this section with the mis-configured simulator used in the previous section. 

To test the method, a set of actions are selected in the repertoire. They are evaluated, one after the other, on the mis-configured simulator and their behavior is observed. If the behavior error (i.e. ``reality'' gap) is greater than the threshold, a new action is computed to achieve the desired behavior. At the same time, the tested action and its neighbors' parameters are updated using the proposed method.

\begin{figure}[h]
\center
\includegraphics[width=6Cm]{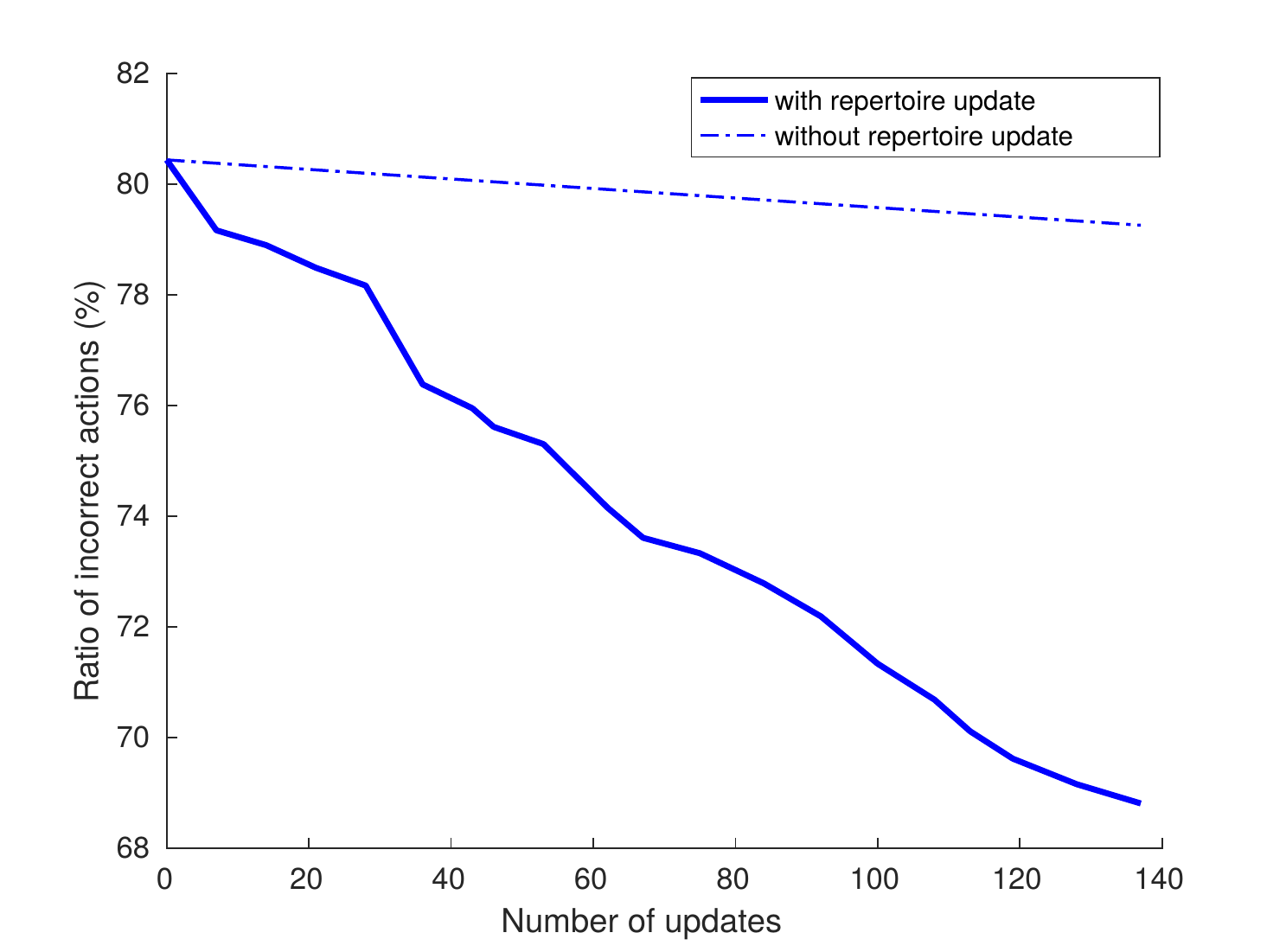}
\includegraphics[width=6Cm]{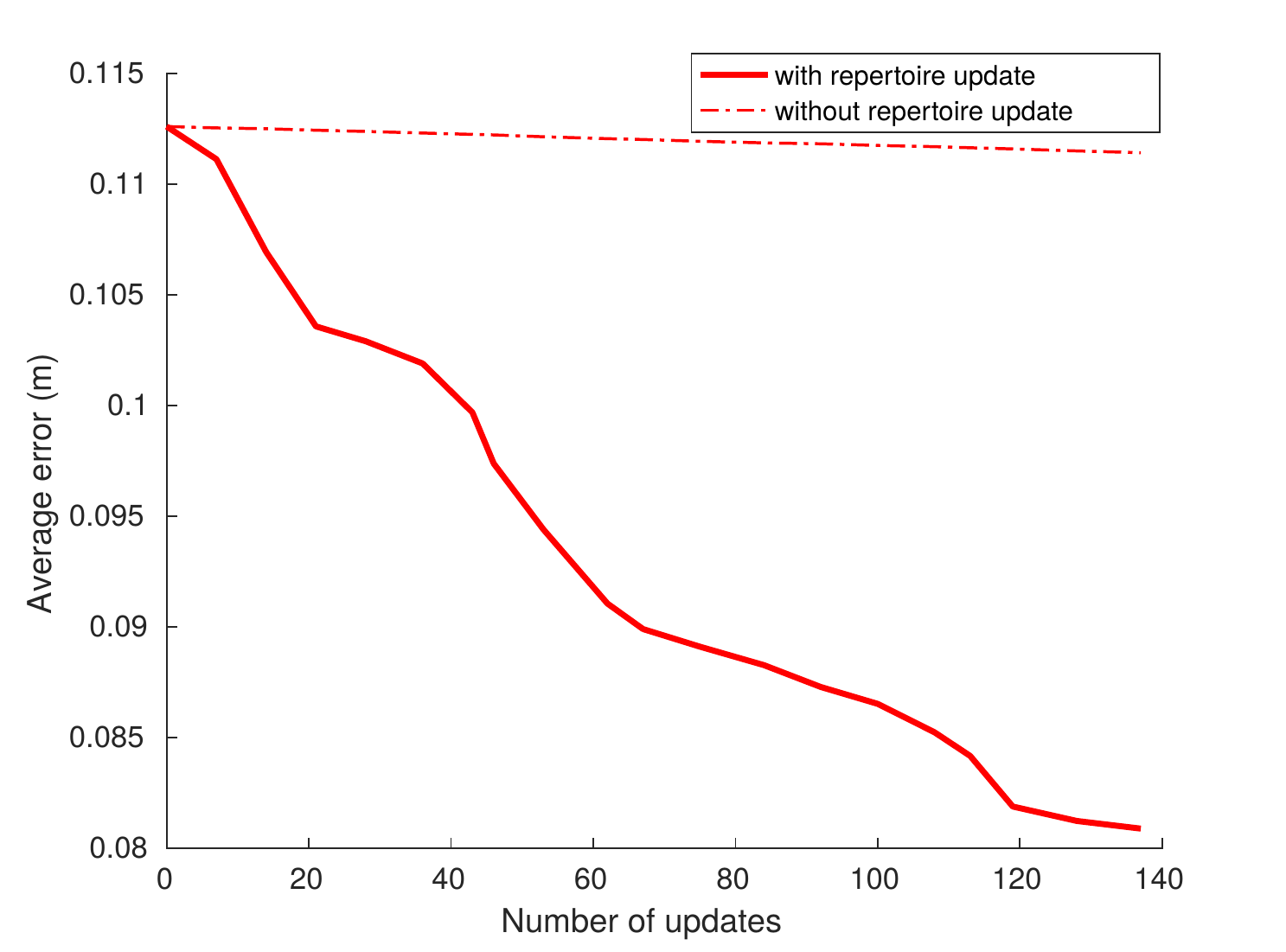}
\caption{Evolution of the reality gap error in the whole repertoire with action adaptations: (left) ratio of actions for which the error is above the threshold ($5cm$)  and (right) average behavioral error. The solid lines show that only running the reality gap crossing process for a few actions and using the computed jacobians to update all the neighboring actions in the action parameters space allows to quickly improve the accuracy of the whole repertoire of \num{11600} actions. By contrast, the dashed lines show the evolution of the same metrics if only the actions on which the reality gap crossing process is run is updated.}
\label{fig:result_reality_gap_by_observations}
\end{figure}

Figure~\ref{fig:result_reality_gap_by_observations} illustrates the performance of the method. In order to measure the impact of repertoire updates, all actions are tested after each update. As can be seen in the figure, the ratio of failing actions (left) and the average behavioral error (right) decreases monotonically with the number of updates. Initially, \num{9331} of the \num{11600} actions (\SI{80}{\percent}) generated an error greater than \SI{5}{\centi\meter} in the artificially mis-configured simulator. After \num{140} trials with repertoire updates, the total number of failing actions is reduced to \num{7995} (\SI{69}{\percent}). It means \num{1336} actions were corrected to an acceptable precision by the \num{140} trials, a speedup factor of \num{9.5} over individual action adaptation. Moreover, the average reality gap on the whole repertoire is also reduced \SI{11}{\centi\meter} to \SI{8}{\centi\meter} an improvement of around \SI{25}{\percent}, whereas individual action adaptation only marginally improves this metric.

\subsubsection{Real robot experiment}
\label{sec:real_robot}

The ball throwing experiment has also been tested on a real Baxter robot. For the target position, a plastic basket with a diameter of $28cm$ is used. The ball and the basket positions are detected by a marker-based motion tracking system (\emph{OptiTrack} by \emph{NaturalPoint}). Four markers were attached to the basket in order to track its position. The ball has been covered with a reflective tape so that the motion capture system can track it. These objects are tracked at $240$ $Hz$.

The ball throwing experiment is repeated \num{50} times with different basket positions that have been randomly selected by the experimenter. The robot finds an action to throw the ball into each selected basket position. Initially, the closest action in the repertoire is used; if it fails and the ball falls outside of the basket, the method to cross the reality gap is applied. 

\SI{92}{\percent} of the launched balls did fall into the basket without adaptation. All of the failed attempts (\SI{8}{\percent}) did reach the basket after a single iteration of the adaptation algorithm. \SI{100}{\percent} of the tests were successful, either before or after adaptation. 

\begin{figure}[h]
\center
\includegraphics[width=12Cm]{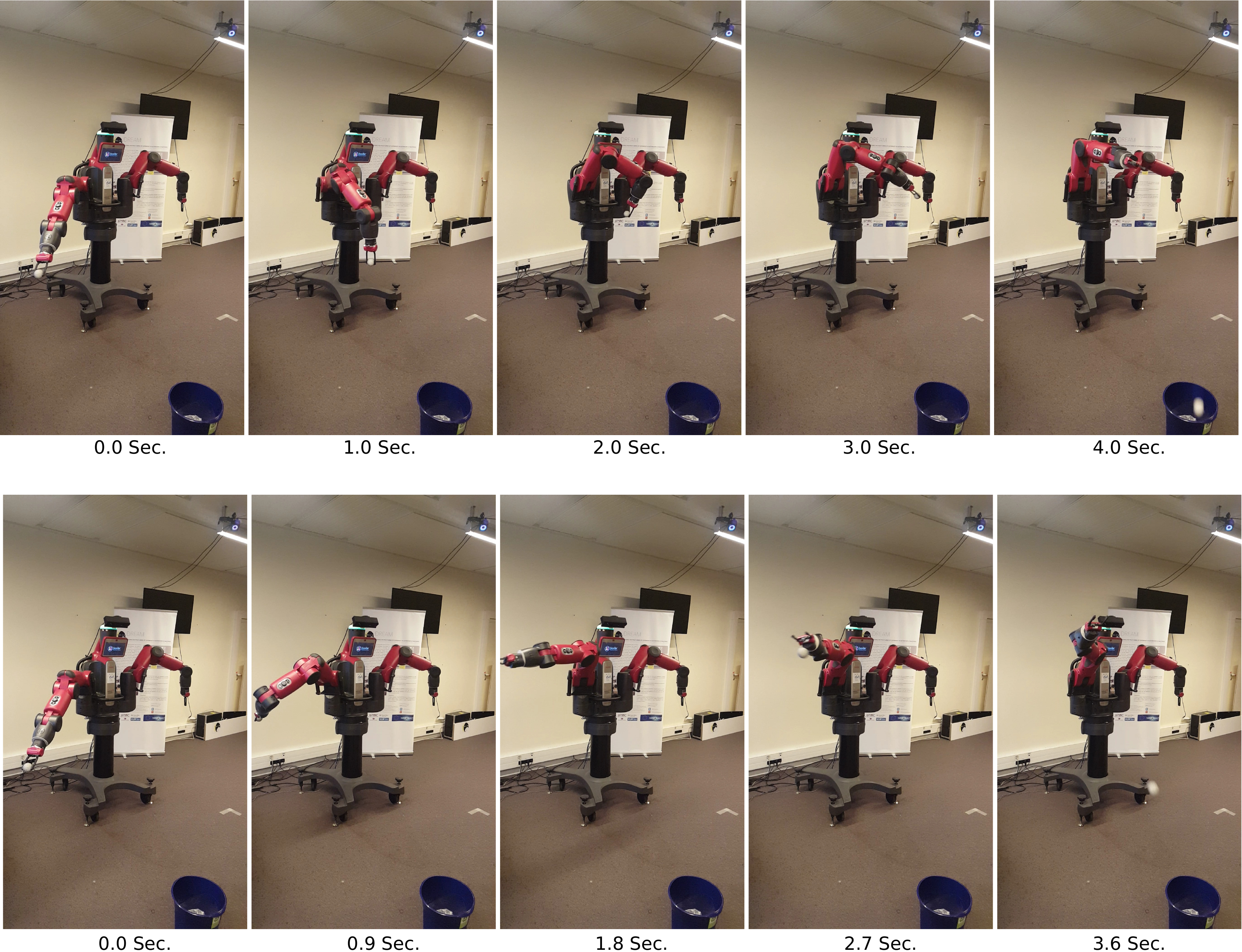}
\caption{Two different throwing actions selected from the repertoire to throw a ball into a single, randomly selected basket position. The full videos are available on \url{https://youtu.be/OVOYIZWZ2R4} }
\label{fig:result_real_throwing}
\end{figure}

Two sets of snapshots are presented in Figure \ref{fig:result_real_throwing}. Videos of the experiments are available on: \url{https://youtu.be/OVOYIZWZ2R4}.

\subsection{Joystick manipulation}
In order to illustrate the general nature of the proposed method, we also apply our method to another task, to control a joystick with the baxter robot. 

\begin{figure}
\center
	\includegraphics[height=3.5Cm]{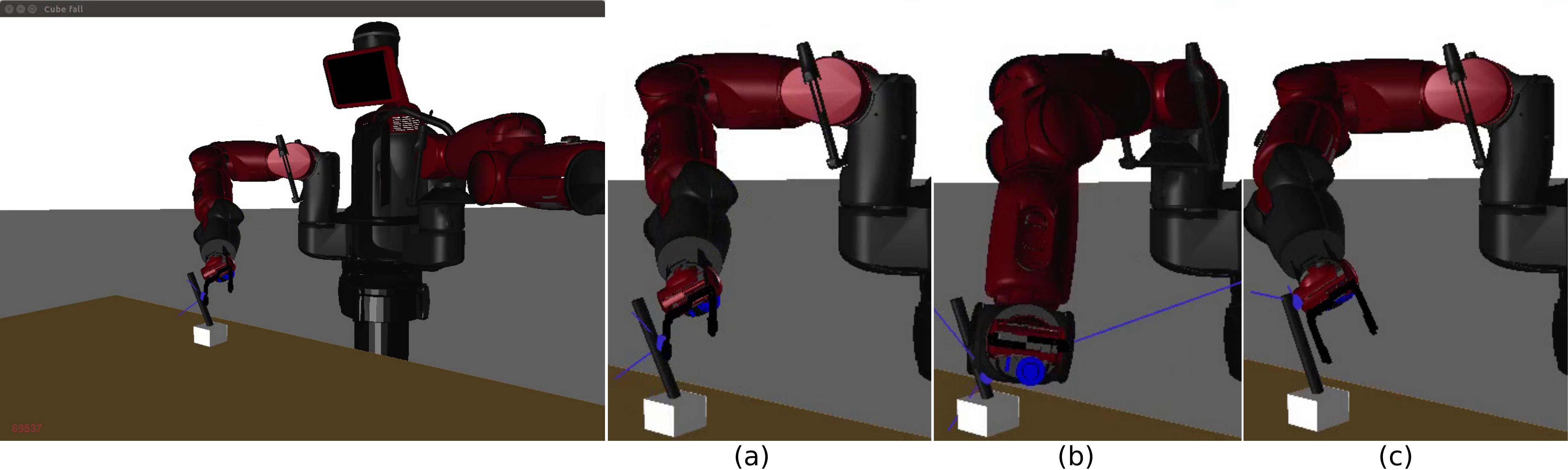} \\
    \includegraphics[height=3.5Cm]{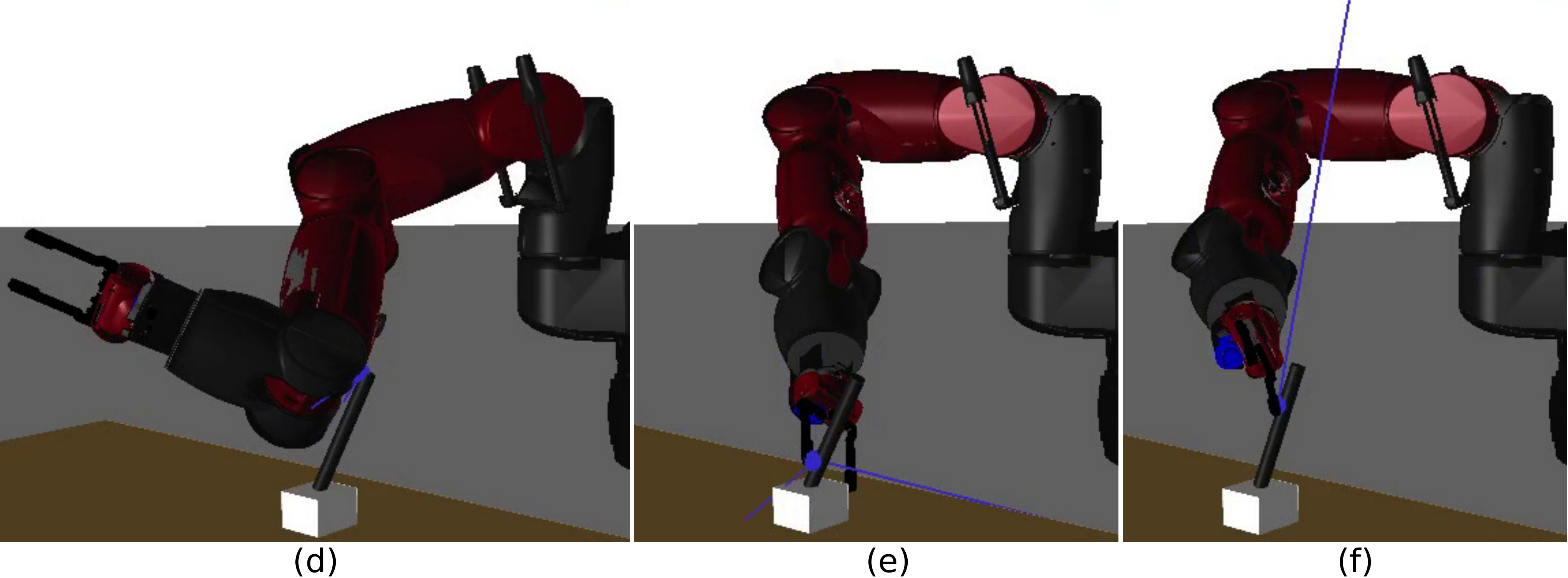}
\caption{The robot controls the joystick using different body parts. Upper and lower figures illustrate various ways of controlling the roll and pitch angles of the joystick for $(0.15, 0.15)$\si{\radian} and $(-0.15, -0.15)$\si{\radian} respectively.
}
\label{fig:qd_3d_result_joystick}
\end{figure}

In this experiment, the robot is not restricted to control the joystick with its gripper. The QD search actually discovers diverse ways of controlling the joystick with other robot body parts. A table is placed in front of the robot, and a joystick is mounted on the table as seen in Figure \ref{fig:qd_3d_result_joystick}. The motion controller is the same as the one used in the previous ball throwing experiment and described in section~\ref{sec:general-setup}. However, the motion duration (the time instance to release the ball for the ball throwing experiment) is set as constant (\SI{3}{\second}) for this experiment. 

The behavior space to control is the joystick position. As in the previous experiment, positions sampled from the gripper trajectories are added to the behavior descriptor used by the QD search in order to generate a more diverse set of actions to reach each particular state. The behavior descriptor is therefore composed of:
(1) two three-dimensional intermediate gripper positions during the motion, and 
(2) the joystick positions (roll and pitch angles) at the end of the robot motion. This behavior space is therefore 8-dimensional.  

To avoid actions that generate too fragile behaviors, i.e. actions that need to be applied with an unrealistic accuracy to get the corresponding state, the quality function represents the robustness of the robot joystick control. It is the negative value of the average variance in the behavior when the action parameters are affected by a small random uniform random noise (minimum \num{-0.01} and maximum \num{0.01} for each dimension). Ten noisy actions are generated and tested on the simulator to evaluate a particular action, and the corresponding state variance (standard deviation of the joystick positions for the noisy actions) is computed.

\subsubsection{Step 1 results: QD search performance}
\begin{figure}[h!]
\center
	\begin{subfigure}{.44\textwidth}
	\centering
	\includegraphics[width=7Cm]{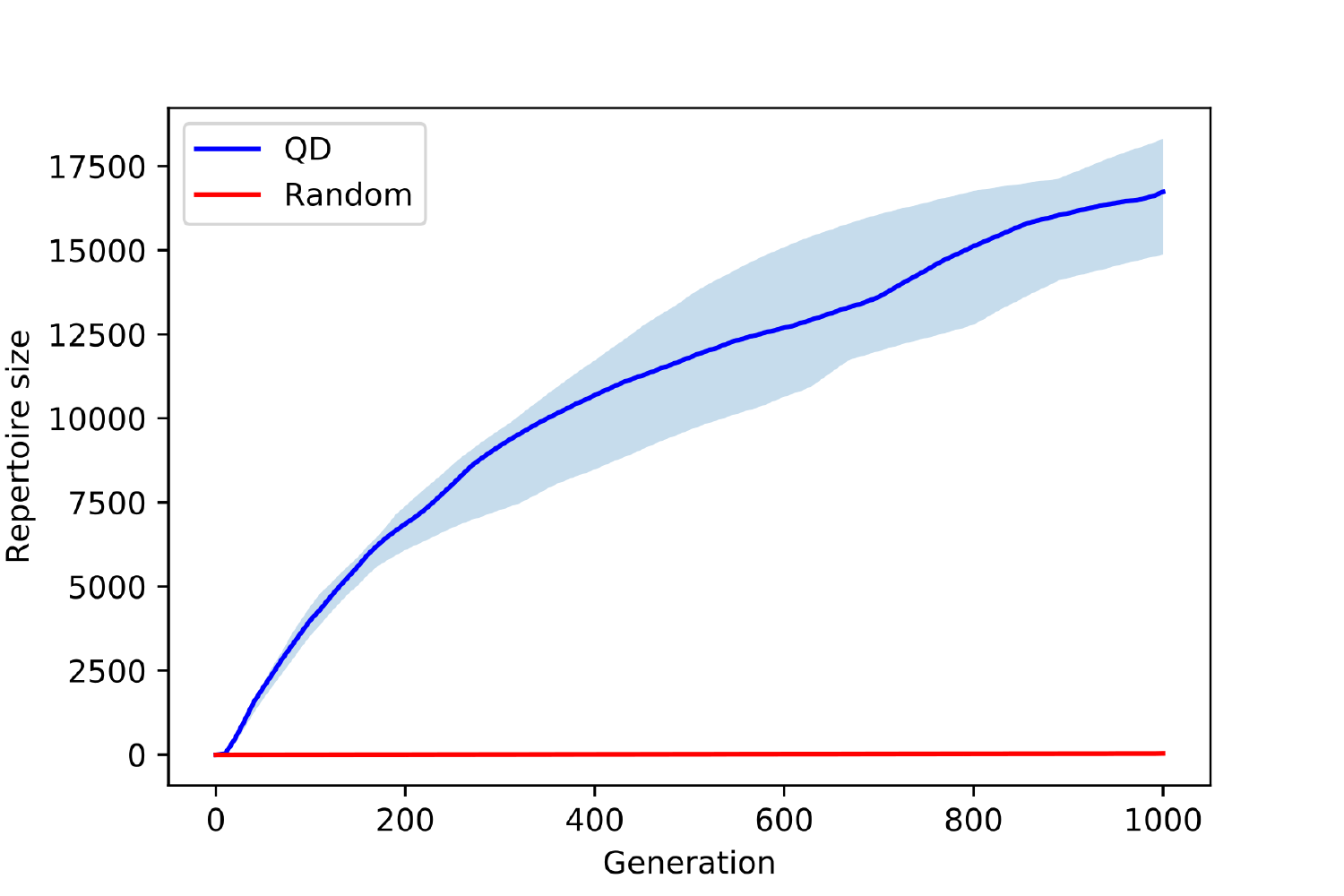}
	\caption{Repertoire size}
	\label{fig:nbsolution_vs_gen_joystick}
	\end{subfigure}
	\hspace{.1\textwidth}
	\begin{subfigure}{.44\textwidth}
	\centering
	\includegraphics[width=7Cm]{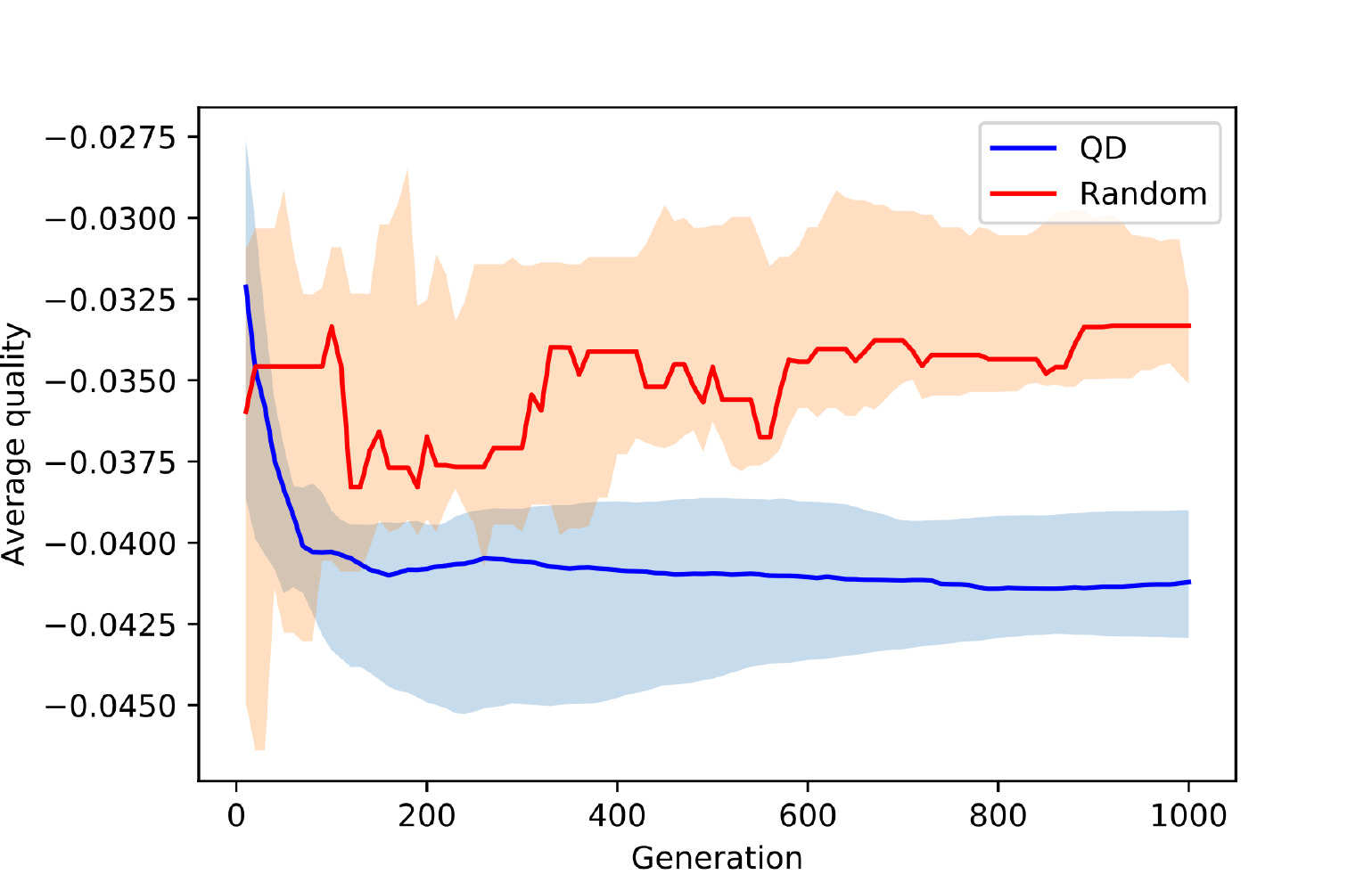}
	\caption{Quality score}
	\label{fig:fitness_vs_gen_joystick}
	\end{subfigure} 
\caption{Number of solutions (a) and average quality of those solutions (b) found by QD search and uniform random search with the joystick scenario.  The median value and interquartile interval on the 12 runs are shown. The uniform random search in action parameter space fails to discover many solutions since only a small subset of the possible actions engage the joystick.}
\label{fig:nbsolutions_vs_gen_joystick}
\end{figure}

The QD search is performed with the same parameters as the ones used for the ball throwing experiment. It was run 12 times and again compared with actions generated by uniform sampling in the action parameter space.The median and interquartile range of the number of actions are shown in Figure~\ref{fig:nbsolution_vs_gen_joystick}. The QD search discovers significantly more solutions than the random sampling approach (\num{16360 \pm 3688} solutions vs \num{40 \pm 4}). The variability of the solutions found by the random sampling is better, but this difference is not significant as the variability of the random sampling setup is evaluated on less than 50 points whereas the variability of the QD repertoire is evaluated on more than 10000 points. The solutions found by the random sampling also have slightly higher quality (Figure~\ref{fig:fitness_vs_gen_joystick}), but again this is a direct consequence of the very small number of those solutions: their quality is simply similar to that of the the first solutions discovered by the QD search (during the first generations).

\begin{figure}
\center
\begin{subfigure}{.40\textwidth}
\includegraphics[width=5Cm]{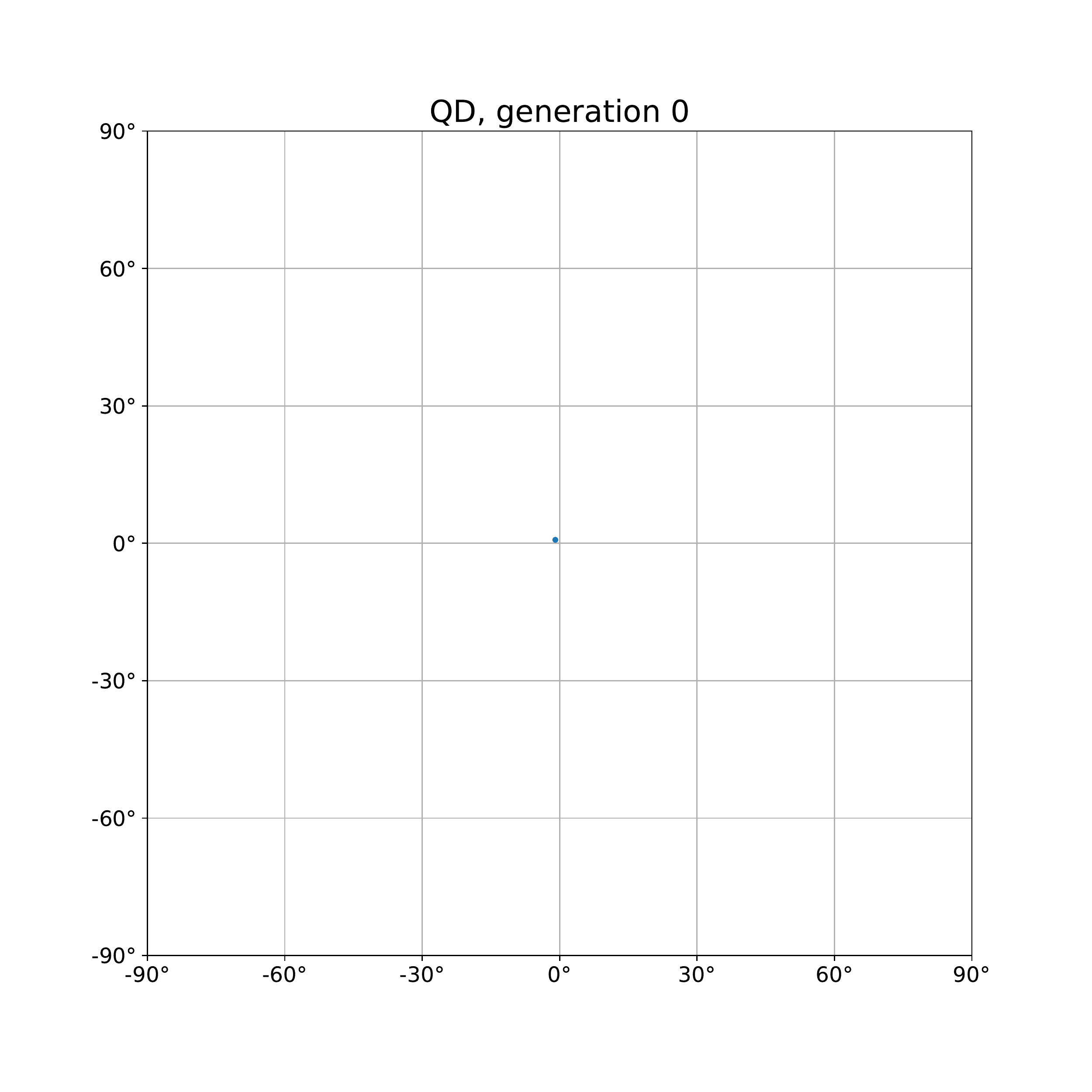}
\end{subfigure}
\hspace{.05\textwidth}
\begin{subfigure}{.40\textwidth}
\includegraphics[width=5Cm]{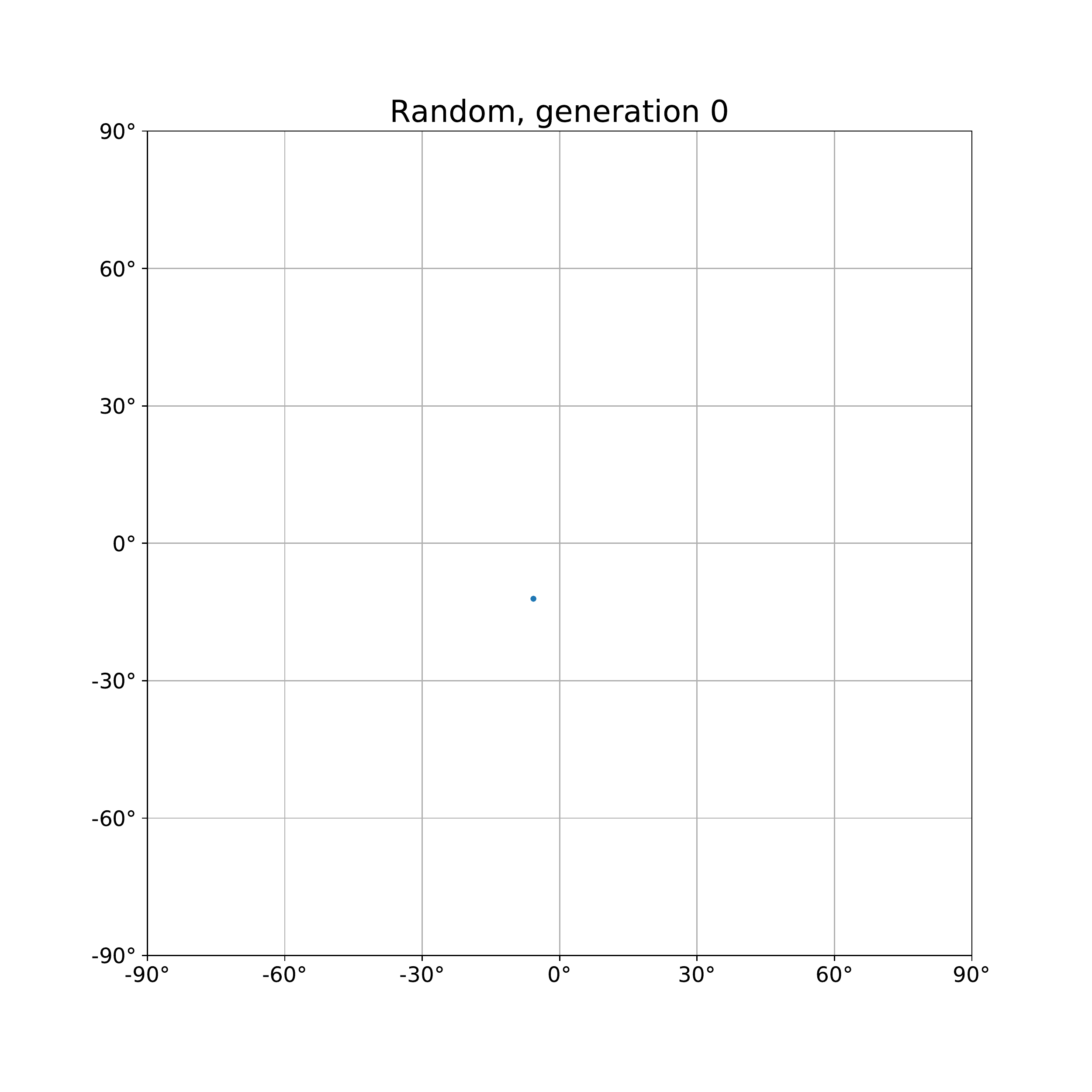}
\end{subfigure}
\\
\begin{subfigure}{.4\textwidth}
\includegraphics[width=5Cm]{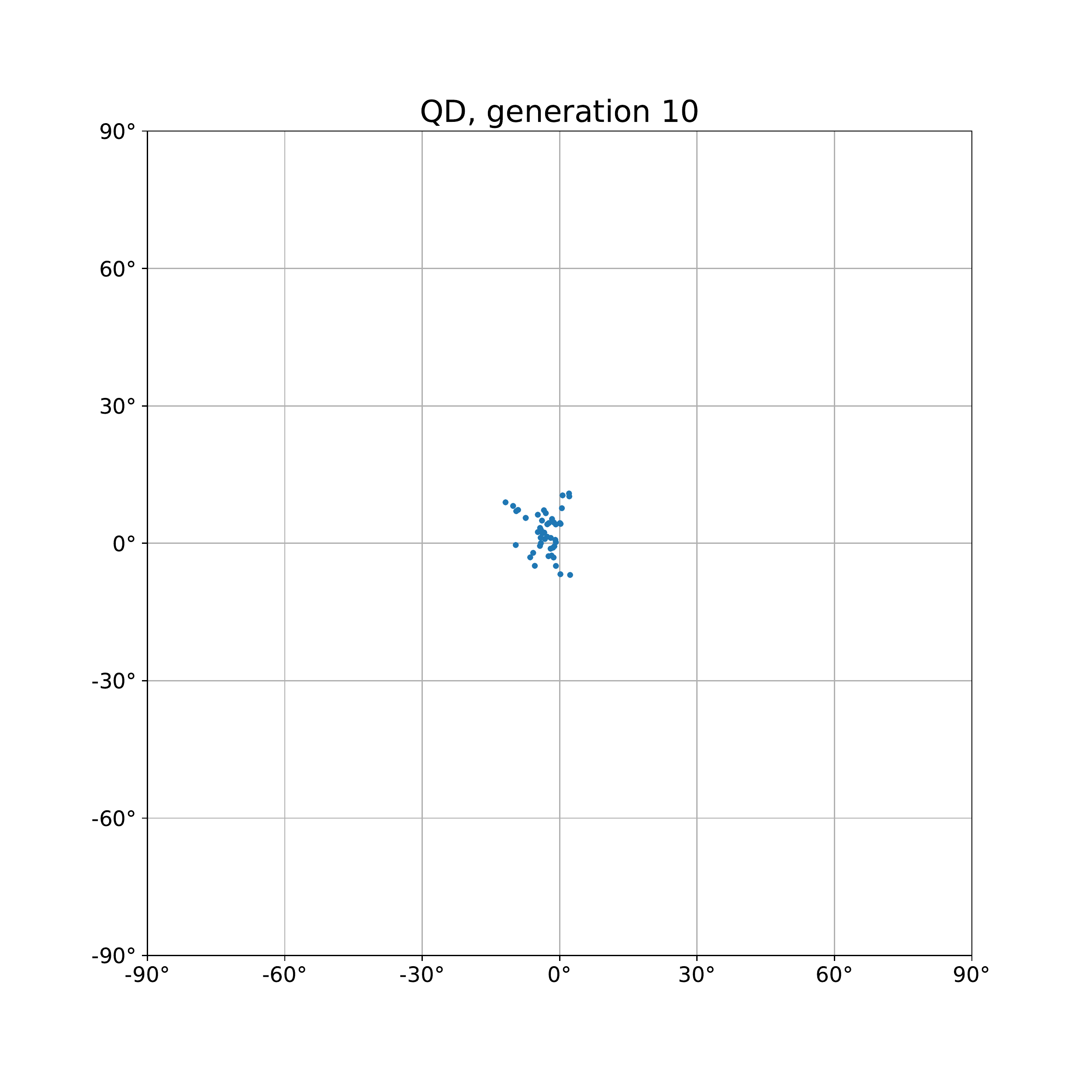}
\end{subfigure}
\hspace{.05\textwidth}
\begin{subfigure}{.4\textwidth}
\includegraphics[width=5Cm]{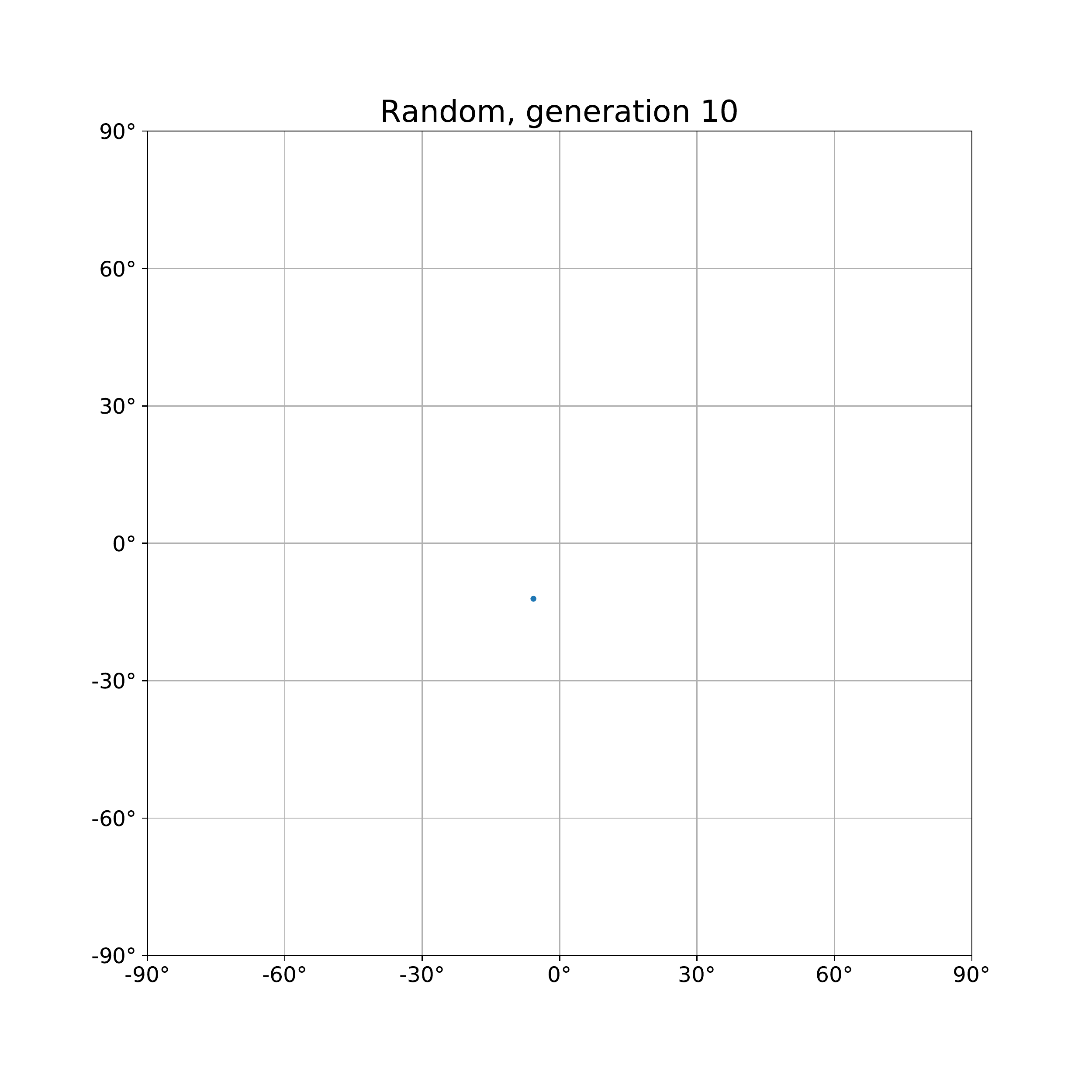}
\end{subfigure}
\\
\begin{subfigure}{.4\textwidth}
\includegraphics[width=5Cm]{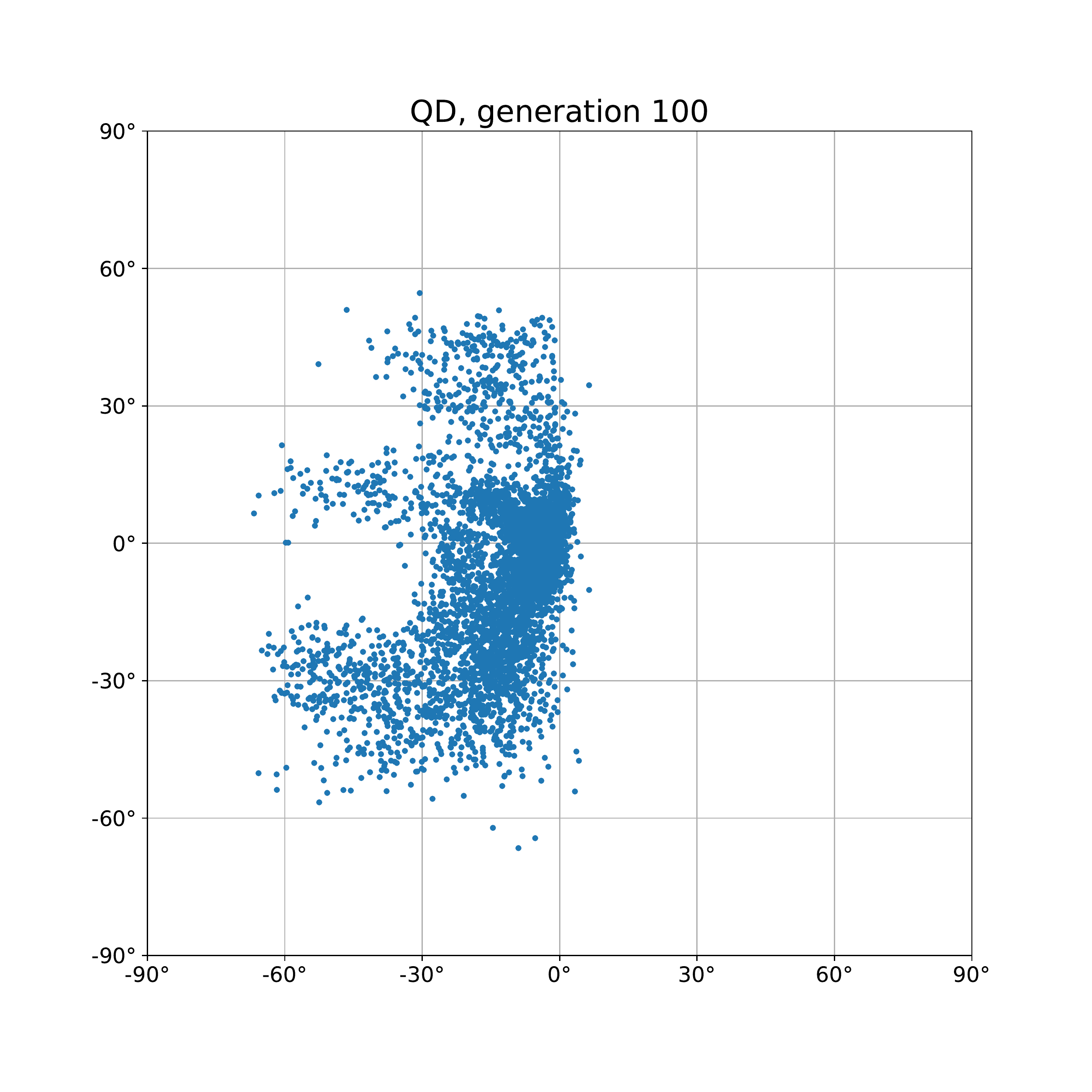}
\end{subfigure}
\hspace{.05\textwidth}
\begin{subfigure}{.4\textwidth}
\includegraphics[width=5Cm]{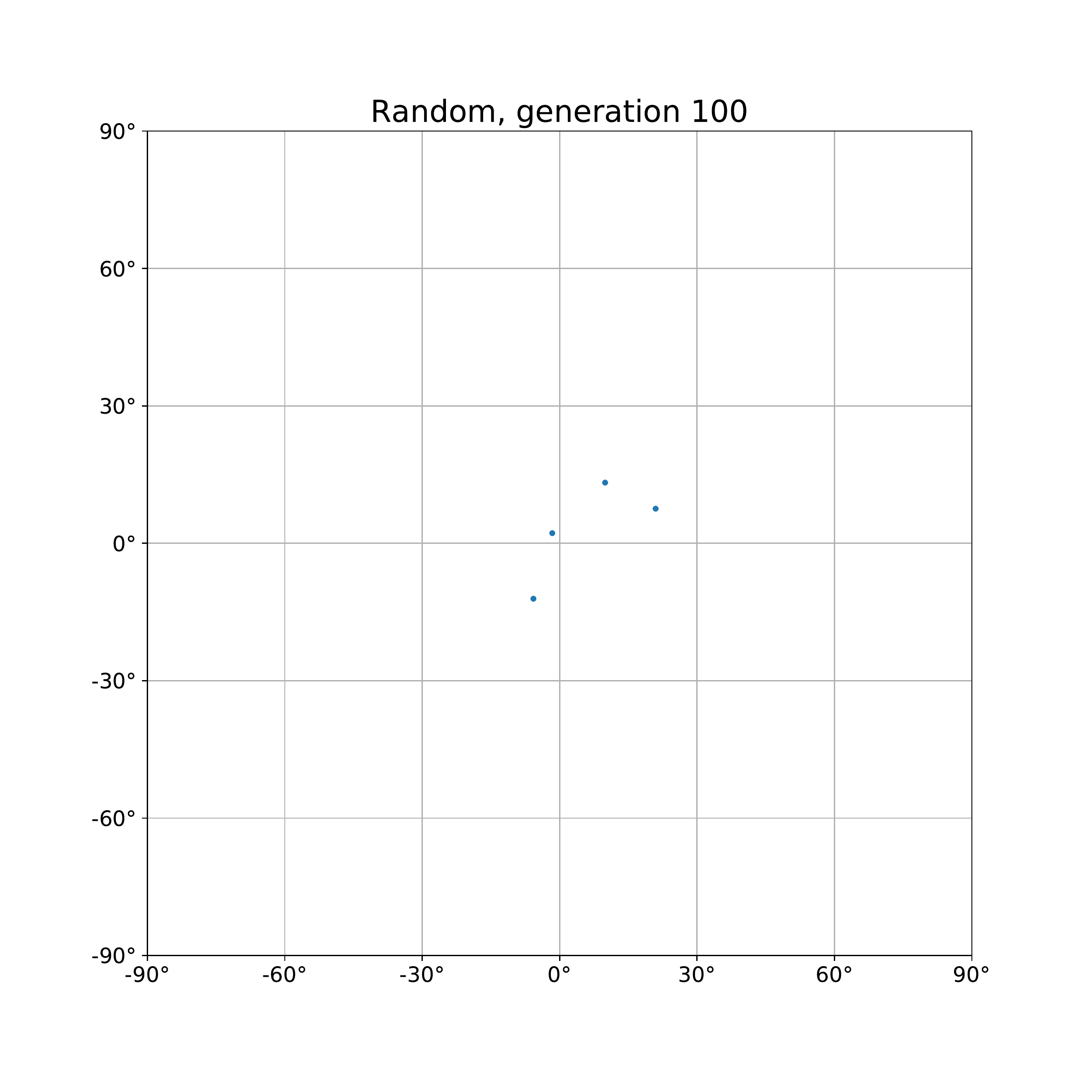}
\end{subfigure}
\\
\begin{subfigure}{.4\textwidth}
\includegraphics[width=5Cm]{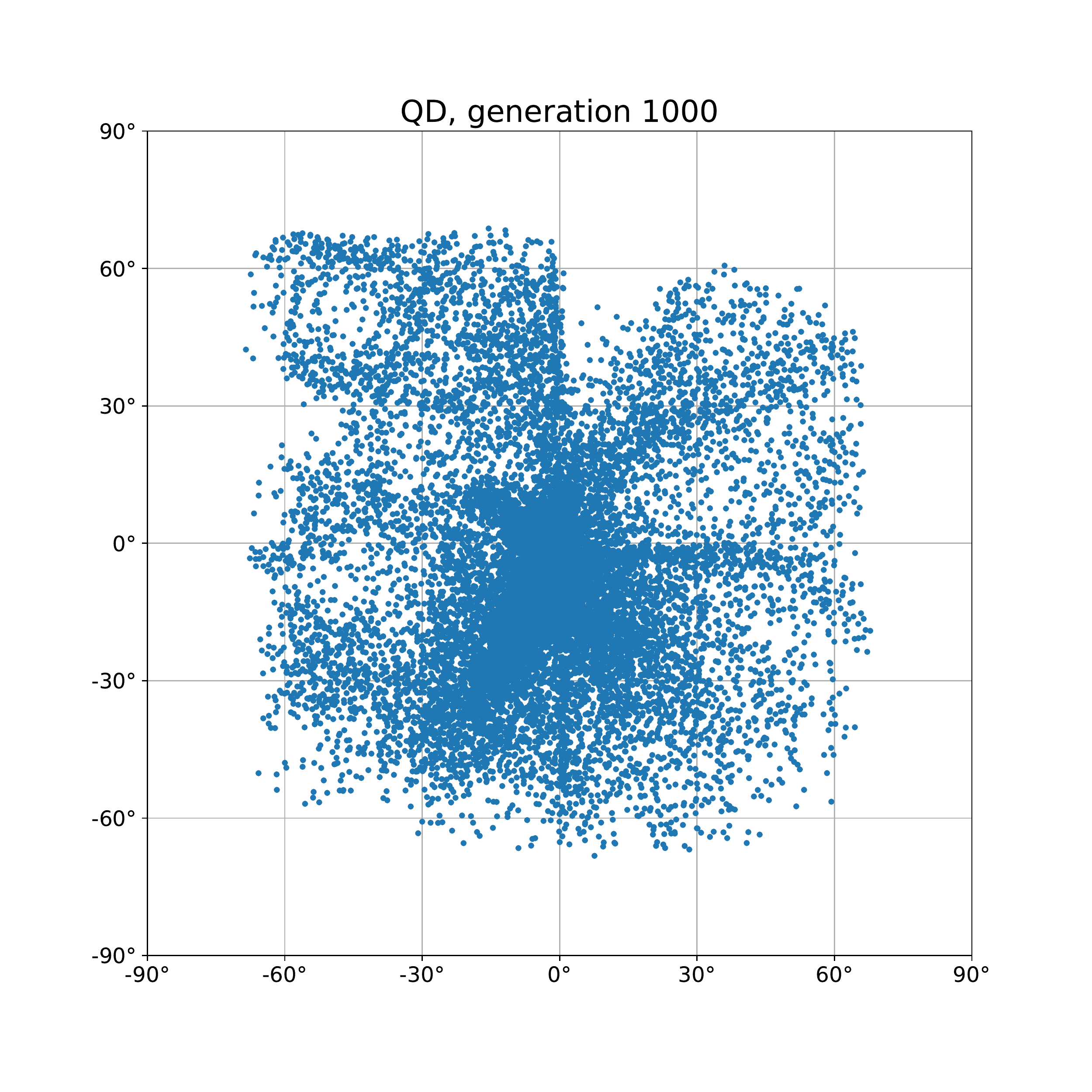}
\end{subfigure}
\hspace{.05\textwidth}
\begin{subfigure}{.4\textwidth}
\includegraphics[width=5Cm]{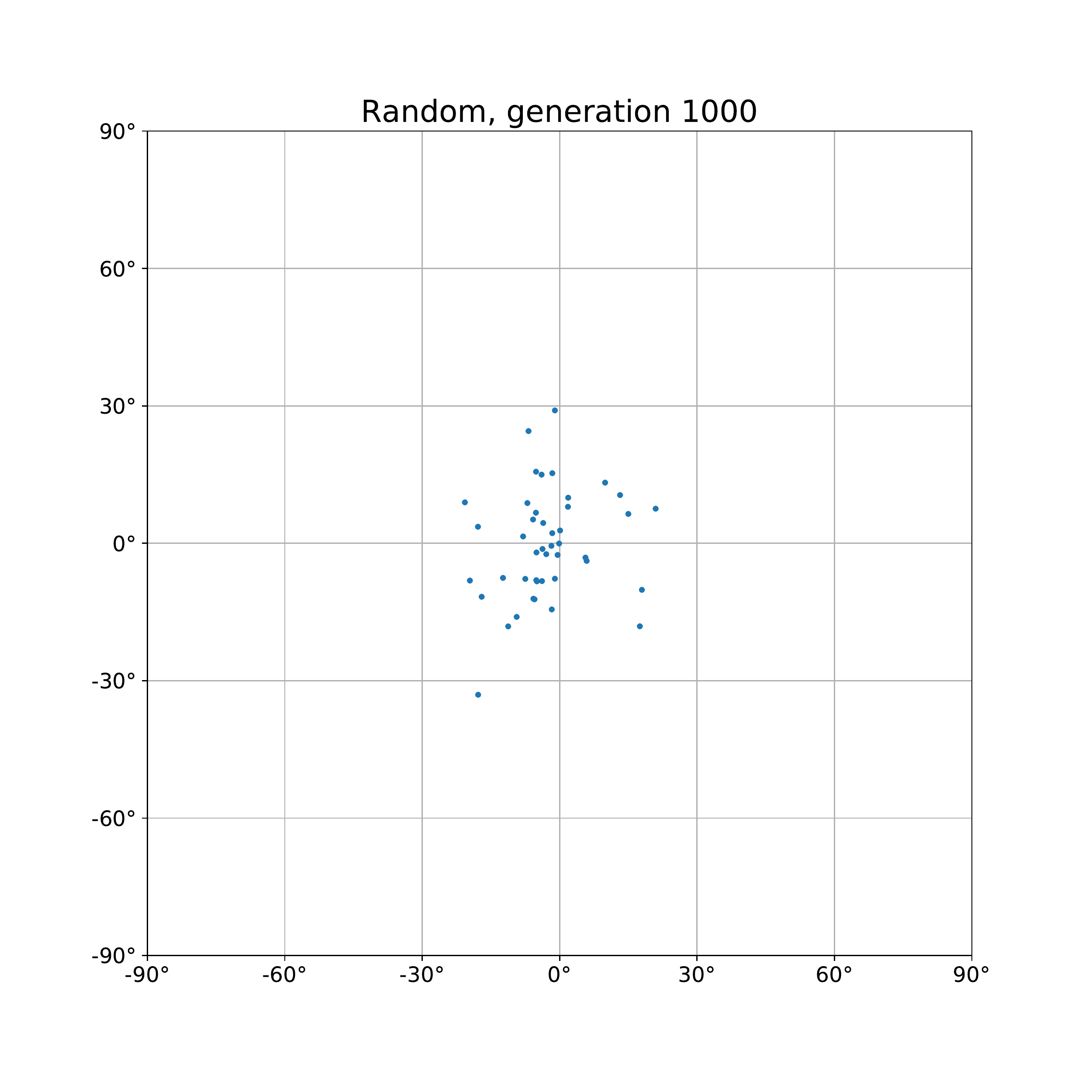}
\end{subfigure}
\caption{Solutions found by the QD search (left) and with random action parameters generation (right). The solutions are shown on the two-dimensional joystick angles space. Dense area contain diverse solutions for controlling the joystick with different robot motion.}
\label{fig:qd_2d_result_joystick}
\end{figure}

The evolution of a repertoire is shown in Figure \ref{fig:qd_2d_result_joystick}. As the space of relevant actions for this task is very small (only a small subset of possible actions interact with the joystick), the initial random population generation is repeated until it finds at least one solution (it usually takes 1 to 4 generations). As can be seen on the figure, QD search gradually discovers solutions covering the state space. 


\begin{figure}[h!]
\center
	\includegraphics[width=.18\textwidth]{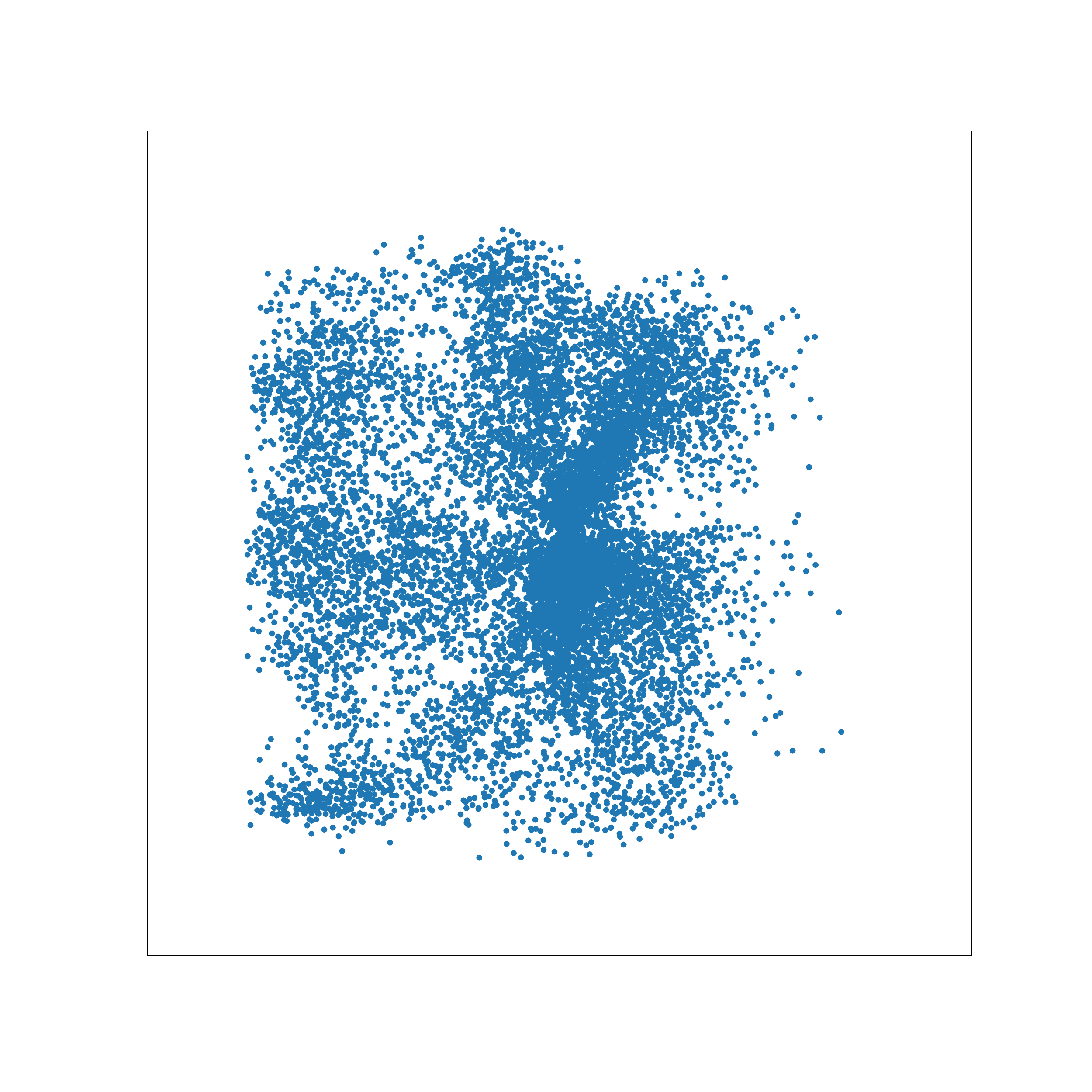}
	\hspace{.01\textwidth}
	\includegraphics[width=.18\textwidth]{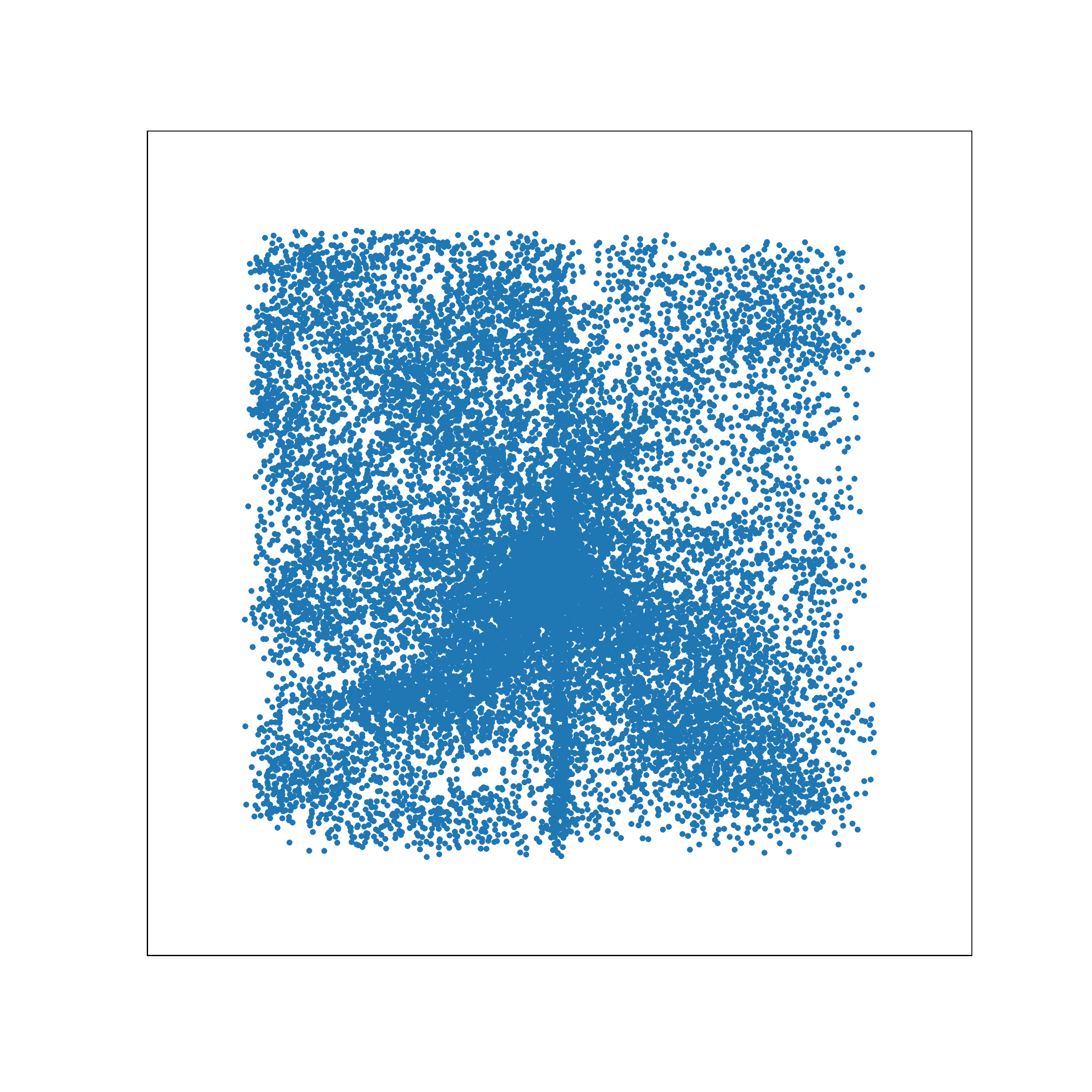}
	\hspace{.01\textwidth}
	\includegraphics[width=.18\textwidth]{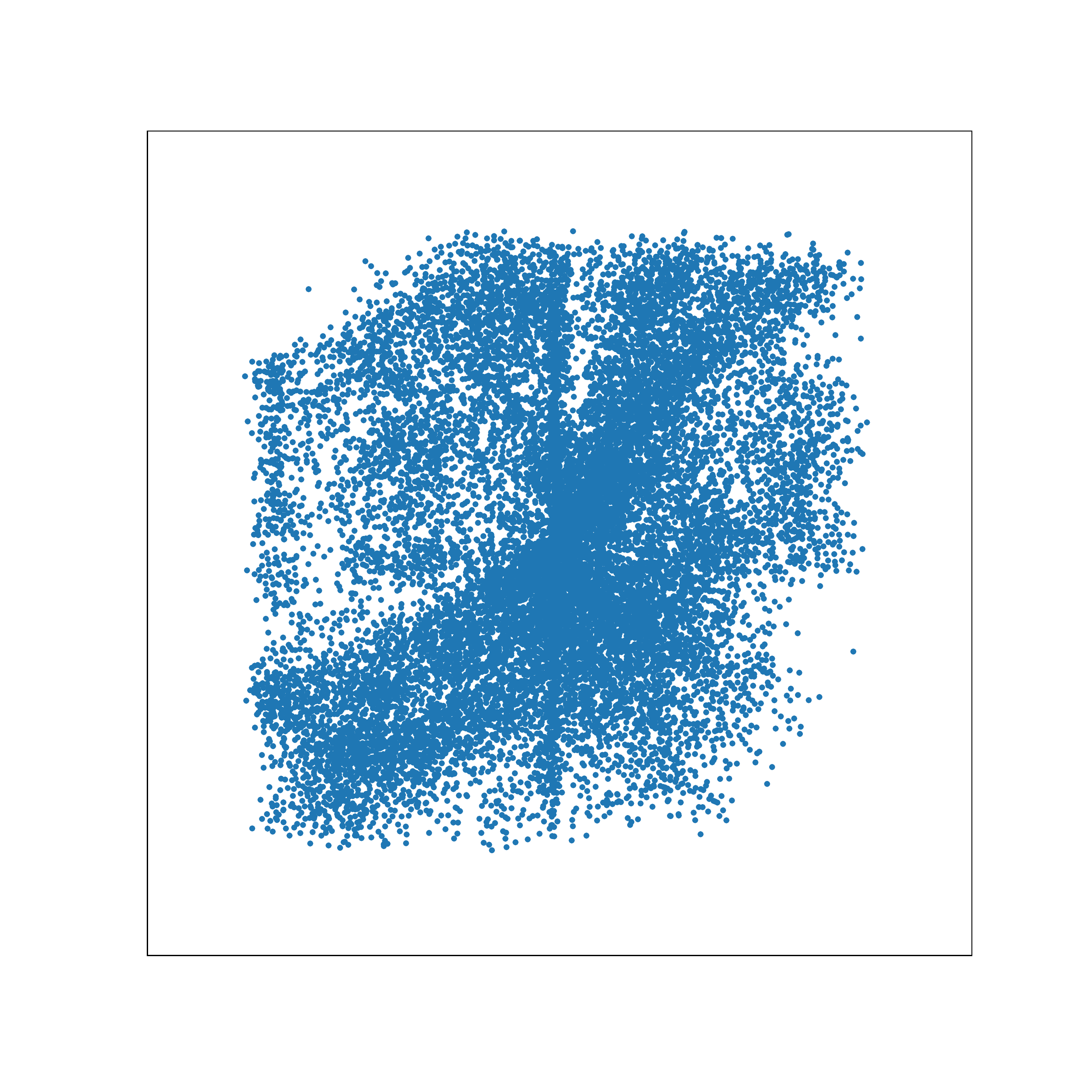}
	\hspace{.01\textwidth}
	\includegraphics[width=.18\textwidth]{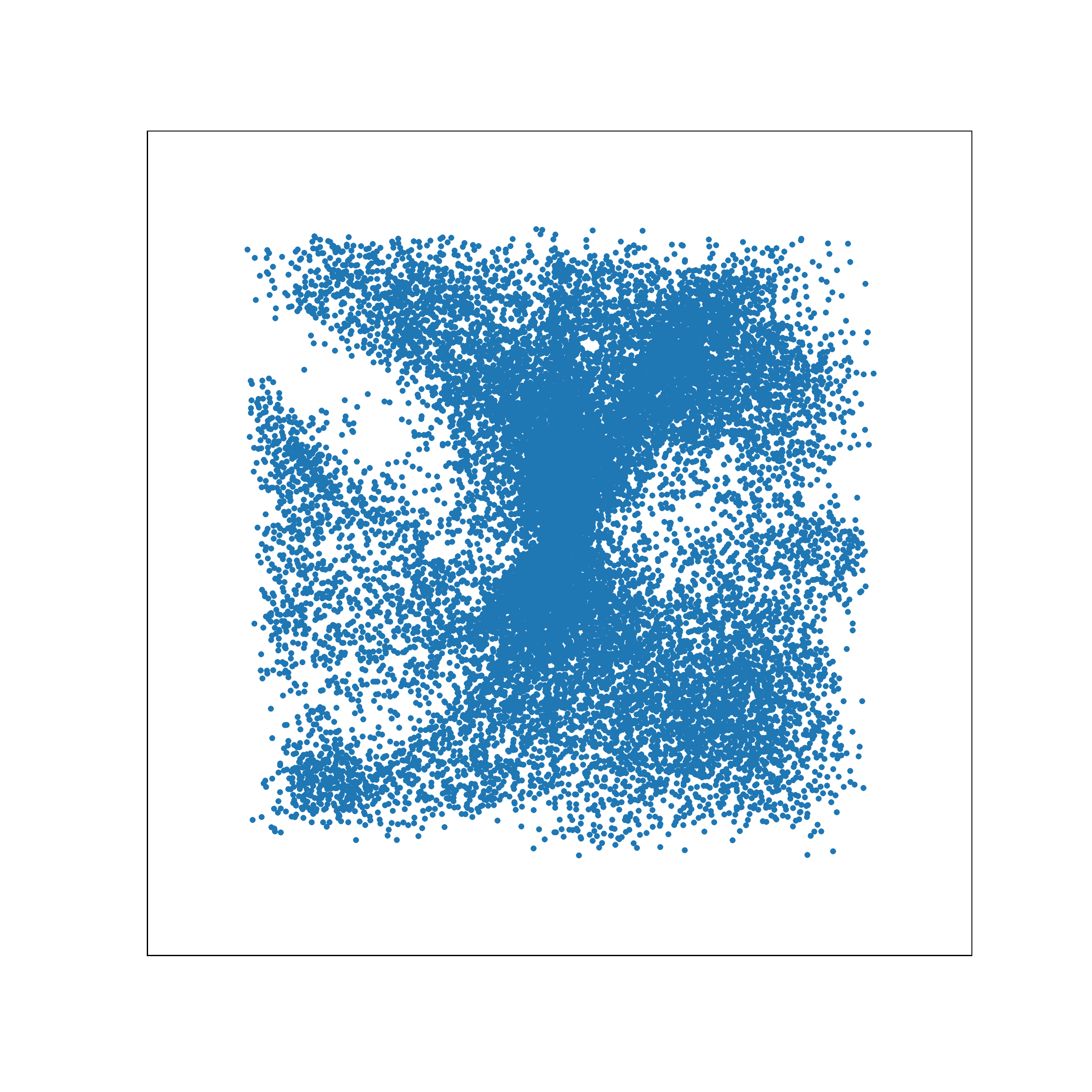}
	\hspace{.01\textwidth}
	\includegraphics[width=.18\textwidth]{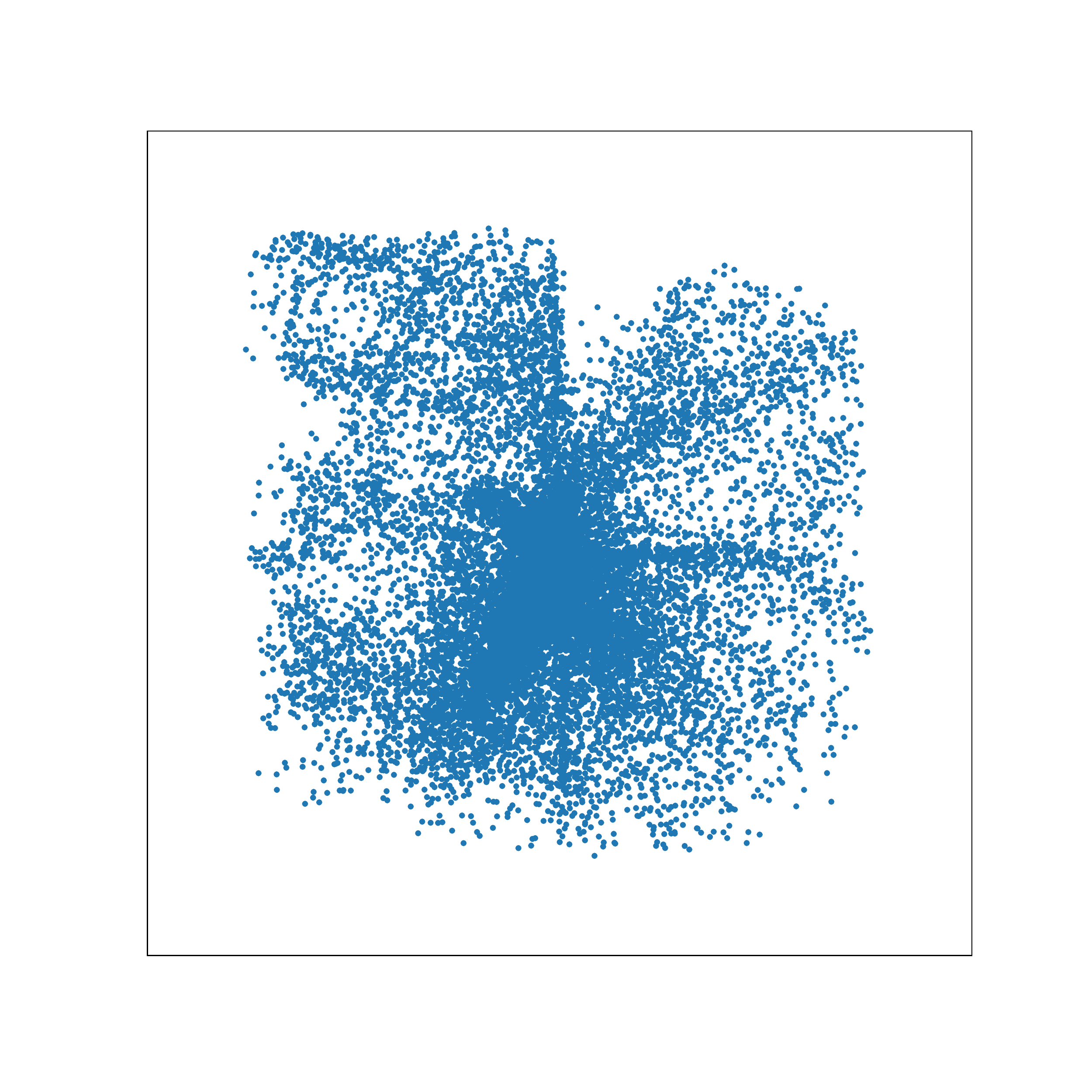}
\caption{Illustration of 5 of the 12 generated repertoires after 1000 generations of the QD algorithm. Although most generated repertoires miss some small part of the goal space, they all densely cover most of the reachable goal space.}
\label{fig:consistency-joy}
\end{figure}

As with the ball throwing scenario, we will conduct the further work on reality gap crossing on one of the generated repertoire, the one containing the median number of individuals among the runs. As depicted in Figure~\ref{fig:consistency-joy}, the final repertoires often miss a small part of the goal space and are not as thorough as in the ball throwing scenario, but all of them densely cover most of the goal space. Running the QD algorithm longer would likely allow the same consistency as for the ball throwing scenario, but the generated repertoires are sufficient to apply our method.

\subsubsection{Crossing the reality gap}

As before, to test the ability of the proposed approach to cross the reality gap, mis-configurations are introduced in the simulator: an offset of \SI{0.01}{\radian} (\SI{0.75}{\degree}) is added to the two joints of the robot shoulder. A smaller offset than for the previous ball throwing experiment is used as this setup has a higher non-linearity. The repertoire used was the one containing the median number of individual among the runs. We consider the reality gap crossed if the joystick is put within \SI{10}{\degree} of the target position.

\begin{figure}[h!]
\center
\includegraphics[height=5Cm]{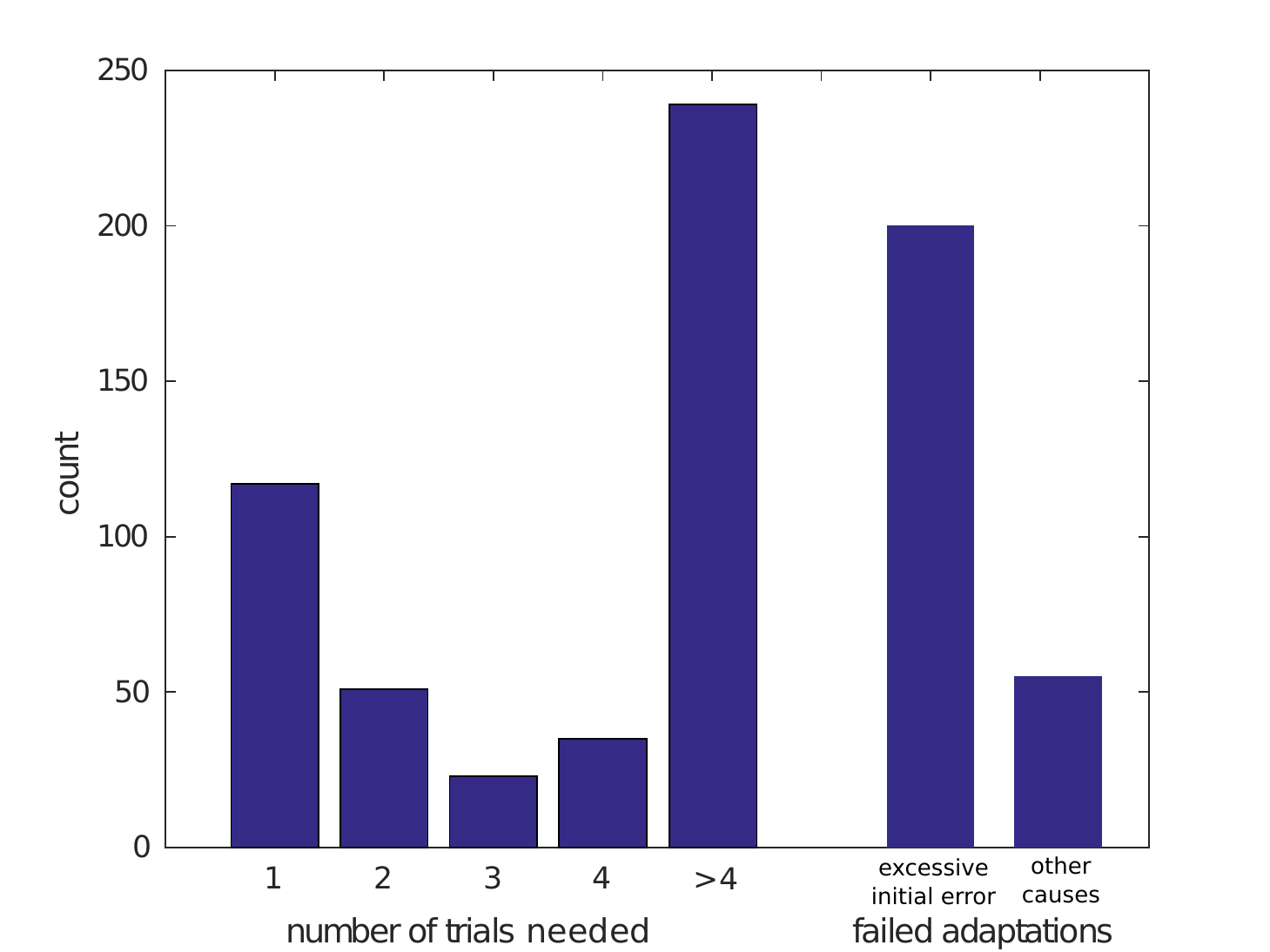}
\includegraphics[height=5Cm]{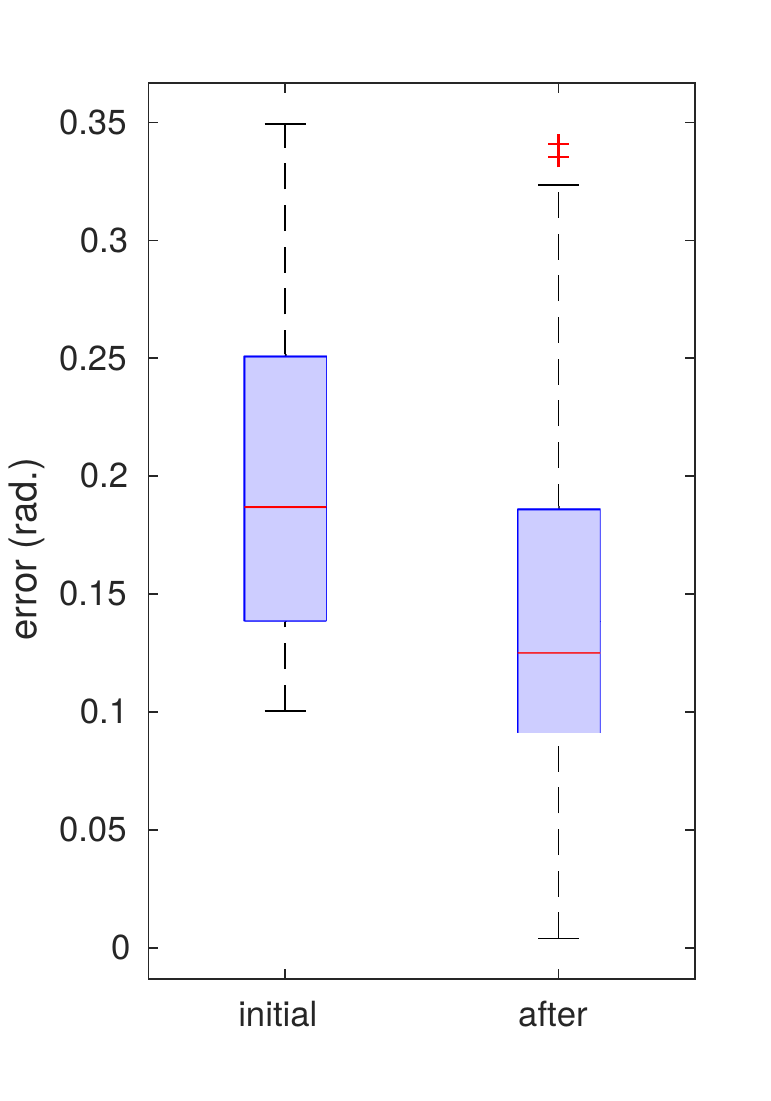}
\caption{Left: histogram of the number of required iteration to cross the reality-gap for the joystick control experiment; ``excessive initial error'' refers to cases where the initial reality gap was so large that the approach could not build a useful local linear model, ``other causes'' groups the other failure cases. Right: reality gap in the 1000 selected actions before and after applying the proposed adaptation approach.} 
\label{fig:crossing_reality_gap_no_update_joystick}
\end{figure}

\num{1000} actions were randomly selected in the repertoire and evaluated in the mis-configured simulator. Among them, \num{280} did not require adaptation (the behavioral error was already smaller than \SI{10}{\degree}) and \num{720} actions needed to be updated (error greater than \SI{10}{\degree}). The proposed adaptation method allowed \num{226} actions to adapt to the new configuration within four iterations or less (see Figure~\ref{fig:crossing_reality_gap_no_update_joystick}), and for \num{239} others the behavior did approach the desired joystick target angles, but did not reach them with the required accuracy within four iterations.

The approach failed for \num{255} actions. For \num{200} of them, the mis-configurations induced a very large initial behavioral error, that was too high for the proposed approach to cope with. The \num{55} others failed for various reasons specific to the experimental setup: for \num{33} actions, the robot arm collided with the table or the joystick base; for \num{14} others the method did not work because the local jacobian was not locally linear, and for \num{8} actions the robot motions for the new actions exceeded the joint position limits.


\begin{figure}[h!]
\center
\includegraphics[width=6Cm]{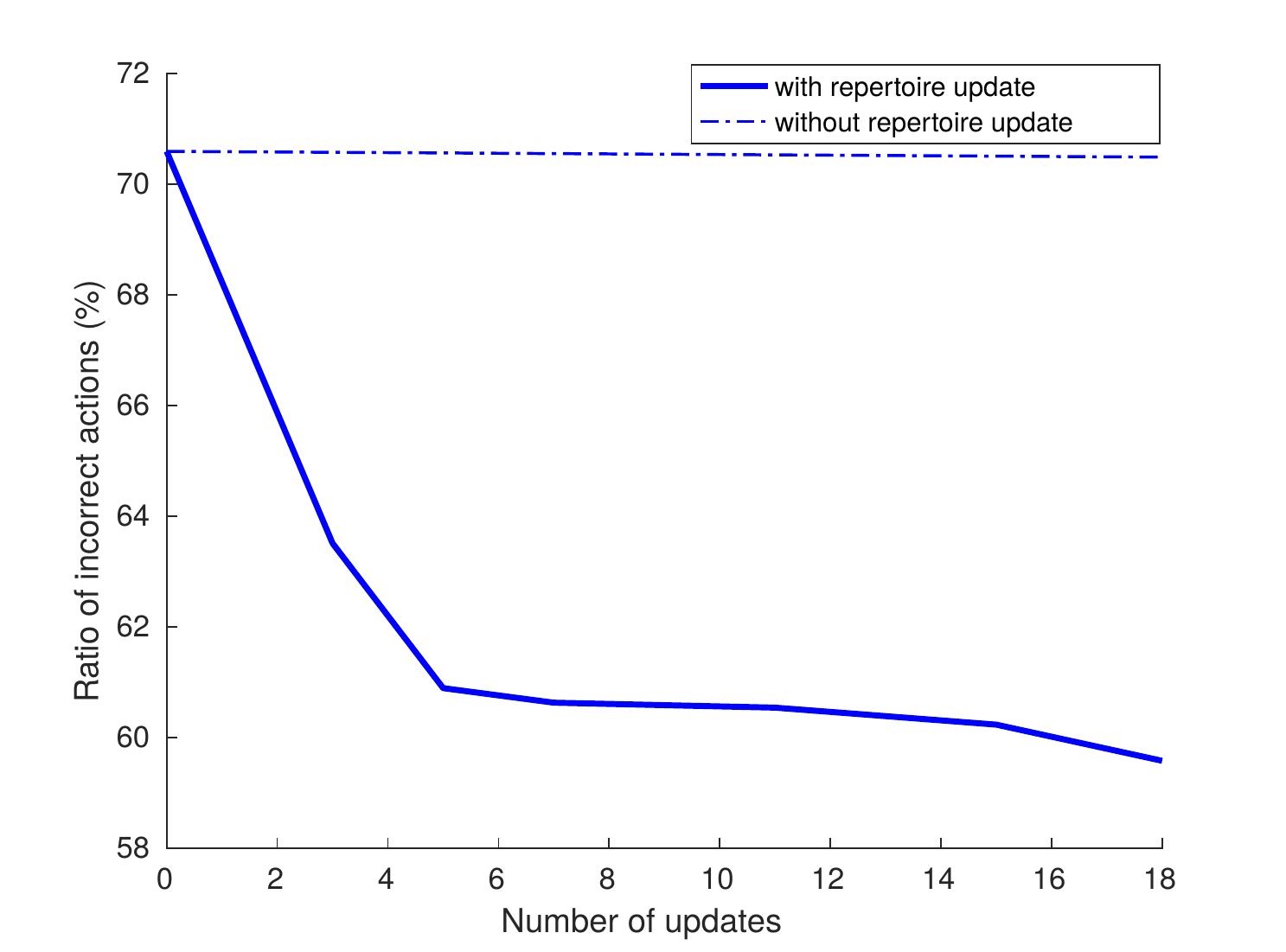}
\includegraphics[width=6Cm]{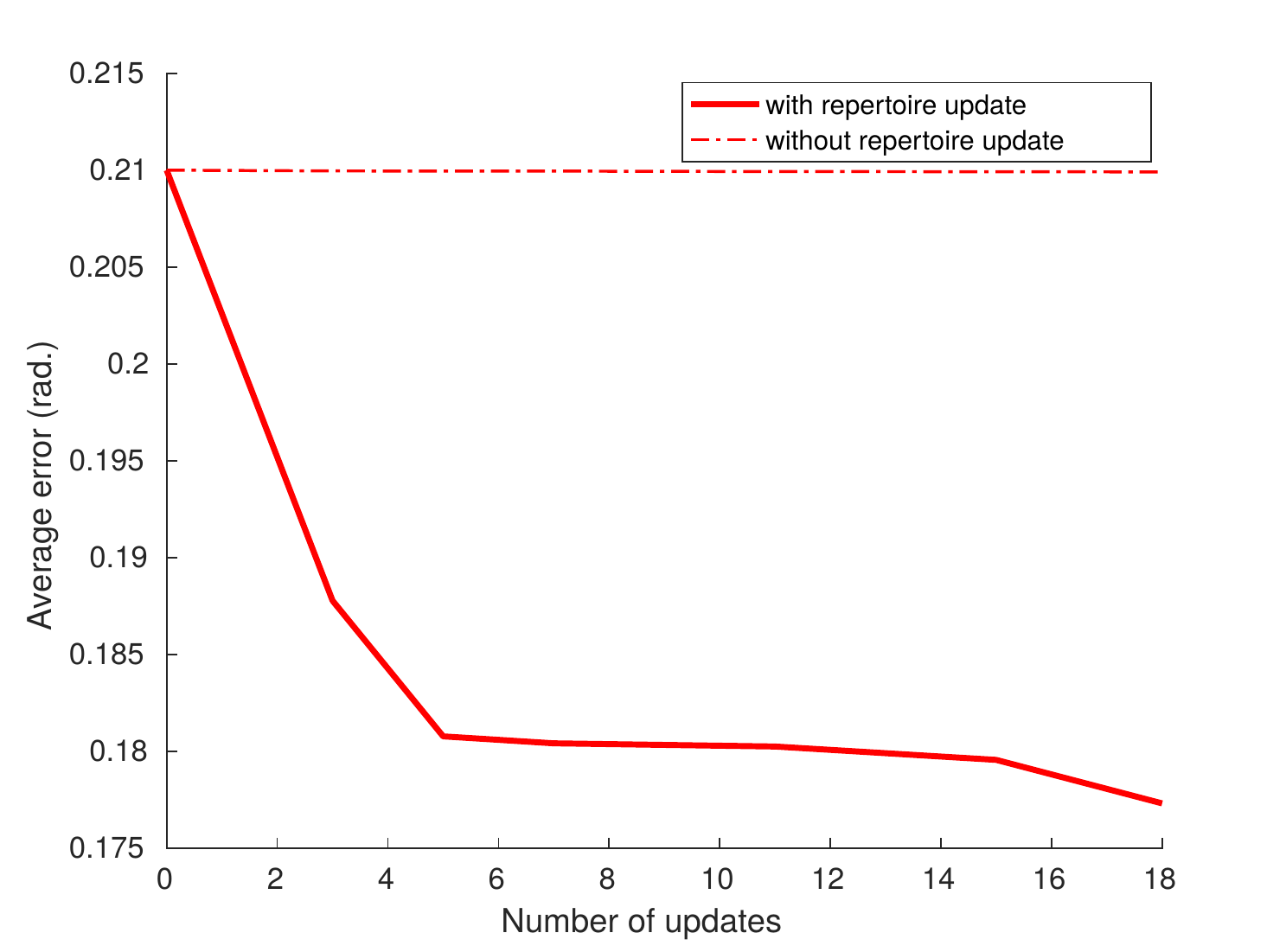}
\caption{
Evolution of the reality gap error in the whole repertoire with action adaptations: (left) ratio of actions for which the error is above the threshold (\SI{10}{\degree})  and (right) average behavioral error. The solid lines show that only running the reality gap crossing process for a few actions and using the computed jacobians to update all the neighboring actions in the action parameters space allows to quickly improve the accuracy of the whole repertoire. By contrast, the dashed lines show the evolution of the same metrics if only the actions on which the reality gap crossing process is run is updated.}
\label{fig:result_reality_gap_by_observations_joystick}
\end{figure}

Whenever an action is evaluated in reality, its behavior is observed and used to update both the action and the other neighboring solutions, as described in section~\ref{sec:method_update}. Initially, around \SI{70}{\percent} of the actions generated an error greater than \SI{10}{\degree} in the artificially mis-configured simulator. After some trials  with repertoire updates, the total number of failing actions is reduced below \SI{60}{\percent}, and the average behavioral error also decreases (Figure~\ref{fig:result_reality_gap_by_observations_joystick}).

\begin{figure}
\center
\includegraphics[width=6Cm]{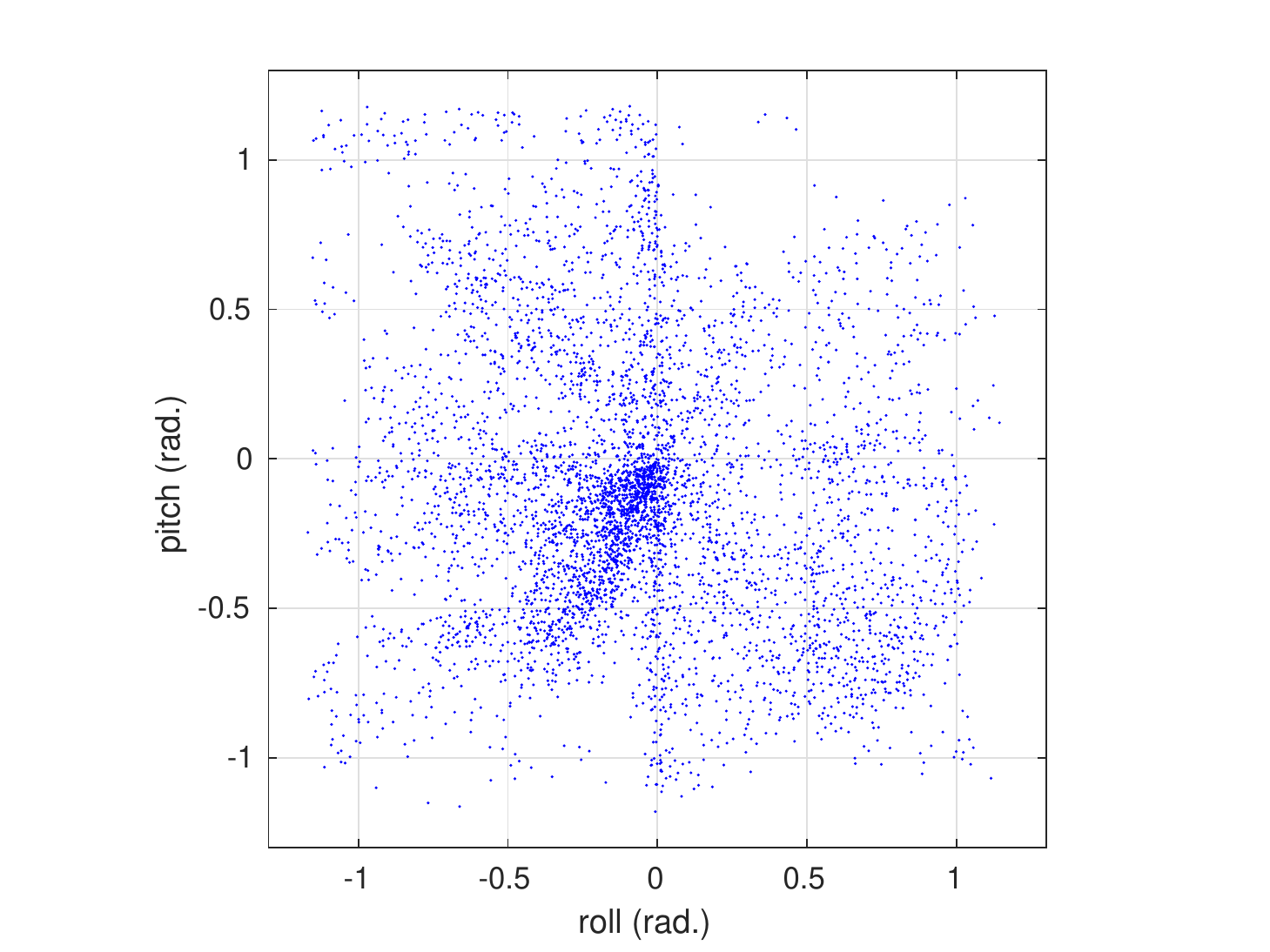}
\includegraphics[width=6Cm]{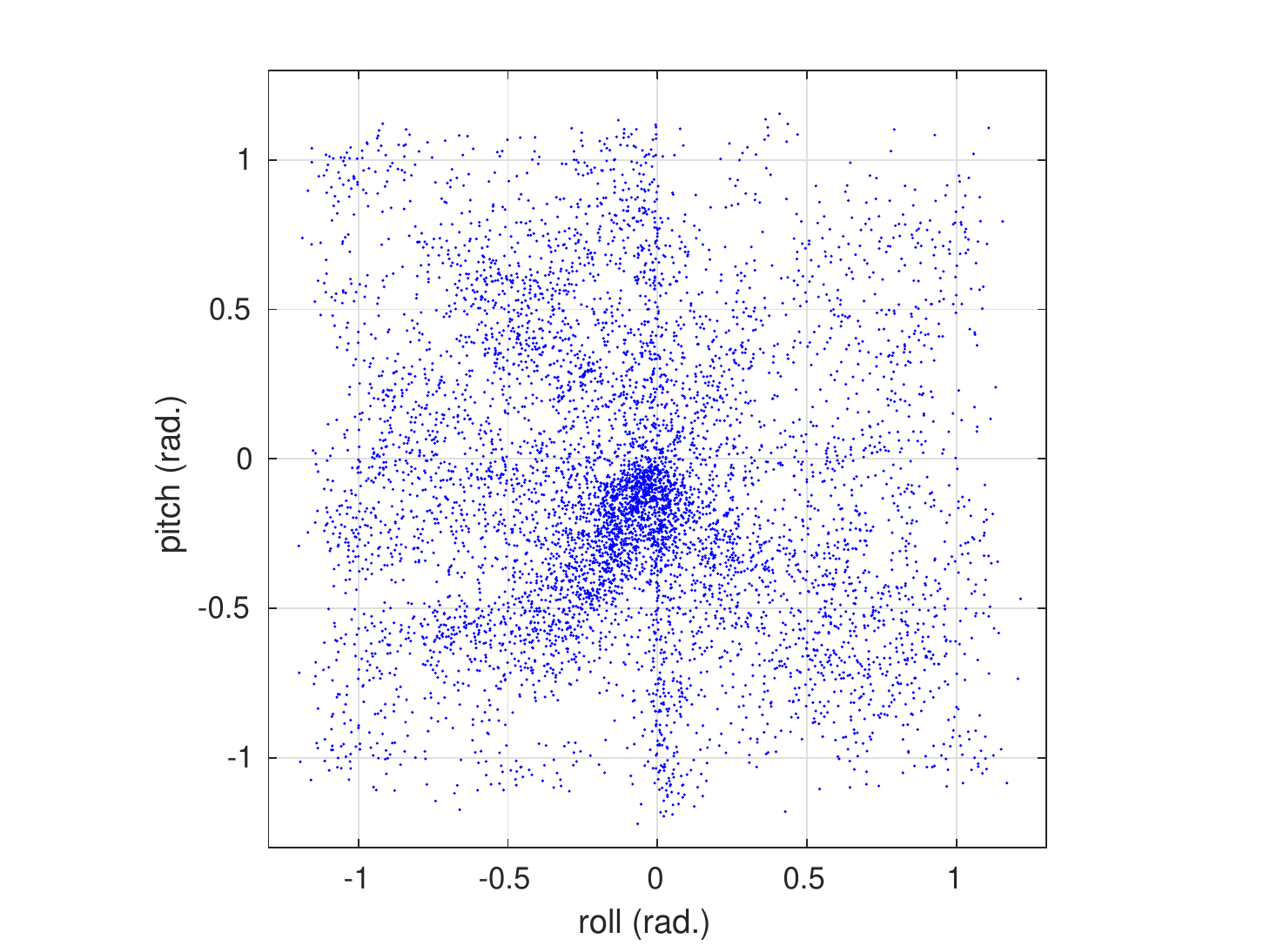}
\caption{Reachable points in the 2D behavior space with the mis-configured simulator, initial (left) and after \num{18} iterations of updates (right).}
\label{fig:2d_crossing_update}
\end{figure}

Figure~\ref{fig:2d_crossing_update} illustrates the feasible actions that behave similarly to the expected behaviors (less than \SI{10}{\degree}). The corresponding repertoire is less dense than the one generated in simulation, but most areas of the behavior space are within reach. Furthermore, the less dense areas (pitch between \SI{0.5}{\radian} and \SI{1}{\radian} and roll below \SI{-0.8}{\radian} or between \SI{0}{\radian} and \SI{0.5}{\radian}, for instance) contain more points after the updates. Although the proposed update method does not allow to update all points, it is enough to keep the control of the state space after the mis-configuration and reach any part of it.


\section{Discussion}
\label{sec:discussion}

Having a robot learn how to control a given goal space and solve tasks with no reversible model or initial demonstrations remains a challenging problem as the space of possible robot motions is large and continuous. Providing a limited set of primitive actions makes learning simpler, but also limits the capacity of the robot to adapt to new situations. Giving the robot the ability to build, on its own, a repertoire of actions according to its needs it thus a promising path towards more adaptive robots. Our results show that this can efficiently be addressed by the use of quality-diversity algorithms, which learn a repertoire of highly diverse actions able to cover the whole goal space. This approach is significantly more efficient than uniform random search in the action space, especially when only a very small proportion of the possible actions affect the goal space, as it is the case with the joystick control experiment. Separating the exploration process, which is slow but can be run in simulation in advance, and the exploitation of the repertoire, which is quick and requires little computation, makes sense in a life-long learning context where a robotic system both interacts with its environment to perform tasks and collect data and uses offline machine learning techniques to learn new skills and restructure its knowledge in an iterative process~\cite{Doncieux2018framework}.

Our approach does not build global task-specific inverse models. Instead, it relies on local inverse models built from a close action and its immediate neighborhood in the action space. Our results show that this approach efficiently reaches arbitrary points in the continuous goal space, generalizing beyond the discrete set of actions provided in the repertoire. Another strength of this approach, compared to using a single inverse model, is that, as shown in figures~\ref{fig:result_generalization_2d} and \ref{fig:qd_3d_result_joystick}, it can fully exploit the redundancy of the robotic system and generate multiple ways to reach a given point in the goal space. This can be useful to address further transfer problems, for example if the environment contains obstacles that make some actions impossible to apply. 
  

The actions in the repertoire are generated in a simulator and may thus generate different effects on the real robot. To deal with this issue, we have proposed to use the local linear inverse models defined by our method to gradually adapt the actions in the repertoire to the real environment. This approach has revealed to be efficient on the ball launching task, but had a more limited impact on the joystick manipulation, highlighting its non-linearity. As the local linearity of the mapping is measured by the $\lambda$ confidence value, those non-linearities could easily be detected and several methods could be proposed to deal with them. First, this value could be used to determine whether the system could be rapidly adapted by following the gradient suggested by the Jacobian, or if it needs to switch to a different approach that would be adapted to a non-linear mapping, for instance a stochastic hill-climbing or an evolutionary search method. Another approach would be to use the repertoire to train a reversible model of the mapping that would be used to deal with the non-linearities. 

The proposed approach relies on an evolutionary algorithm and thus on a blind variation and selection process. This property is interesting in an adaptation process, as it is very versatile and does not require strong assumptions on the actions to be generated. It is furthermore hypothesized to be at the roots of human creativity~\cite{dietrich_human_2014}. However, this flexibility has a price: the search process needs to evaluate a large number of samples. This issue has been tackled here with an exploration in simulation, that has no impact on real robot wear and tear and can easily be accelerated and parallelized. In a life-long learning perspective, the sample efficiency of the proposed approach could be enhanced by relying on previously built repertoires, for instance to seed a new repertoire with the actions generated for a different but related behavior space. This would raise the question of the identification of the relevance of a given repertoire to generate a new one, which is still a challenge in the transfer learning community.

A model of the mapping could actually be learned during the exploration and used to bias the exploration and make the search more sample efficient, as done in Generative Adversarial Networks~\cite{goodfellow_generative_2014,Jegorova2018gan}. The advantage of an unbiased search as proposed here is that it may better lead to serendipity and the discovery of non linearities that have not been captured yet in the model. In a life long learning framework in which computational time is not the main issue, the two approaches could thus be complementary: a model based approach could be used on the real robot to get a first glance on the robot possibilities, then a deeper exploration based on the approach proposed here could be performed to get a larger control over the state space before using the generated data to train a more accurate model.

\section{Conclusion}

In this paper, we have introduced a novel framework for a robot to efficiently discover skills to control a goal space by exploring in a simulated environment, with no demonstration, no prior policy, and without building an explicit inverse model of the system. The proposed method builds a large, diverse discrete action repertoire densely covering the goal space, and then builds local models to generalize to other points in the goal space or to adapt to slightly different tasks and environments, for exemple crossing a reality gap. Repertoire acquisition is done with a Quality Diversity algorithm whose goal is to build a set of actions to reach a diverse set of states in a space described by predefined behavior descriptors. The outcome of this process is a set of up to tens of thousands of actions. The generalization and adaptation relies on a local linearization of the mapping between the action parameters and the corresponding state. The proposed framework has been validated in two different applications which make a robot discover diverse robot arm motions for two different tasks: launching a ball into a basket and controlling a joystick. In both cases, the method discovers a large number of diverse actions, allows the robot to accurately control the goal space, and is used to cross the reality gap from the simulator to a real environment.


\section*{Acknowledgement}
This work is supported by the DREAM project \footnote{http://www.robotsthatdream.eu} through the European Union's Horizon 2020 research and innovation program under grant agreement 640891. 
The authors would like to thank Ghanim Mukhtar for his help on the real robot experiment. 

\bibliographystyle{elsarticle-num-names}
\bibliography{zotero,Repertoire_based,library}

\appendix 
\section{Parameters used for QD search}
\label{appendix:qd_pram}
\begin{center}
\begin{tabular}{ l l l } 
 \hline
Population size & & 240 \\ 
Number of generation & & 2000 \\  
Mutation rate & & 0.2 \\  
Cross over & & 0.1 \\  
 \hline
action distance threshold in repertoire (ball throwing experiment) & $l_{repertoire}$ & 0.01 \\ 
action distance threshold in repertoire (joystick experiment) & $l_{repertoire}$ & 0.05 \\ 
number of neighbors to compute a Jacobian & $K$ & 30 \\ 
Jacobian confidence threshold & $\eta_{threshold}$ & 0.3 \\ 
  \hline
\end{tabular}
\end{center}

\end{document}